\documentclass{article} 
\usepackage{iclr2021_conference,times}

\usepackage{amsmath,amsfonts,bm}

\def\eqref#1{equation~\ref{#1}}

\def\1{\bm{1}}

\DeclareMathAlphabet{\mathsfit}{\encodingdefault}{\sfdefault}{m}{sl}
\SetMathAlphabet{\mathsfit}{bold}{\encodingdefault}{\sfdefault}{bx}{n}

\DeclareMathOperator*{\argmax}{arg\,max}

\usepackage{hyperref}
\usepackage{url}
\usepackage{microtype}
\usepackage{graphicx}
\usepackage{subfigure}
\usepackage{booktabs} 

\usepackage{bm}
\usepackage{amsmath}
\usepackage{multirow}
\usepackage{amsfonts}

\usepackage{xr}
\title{Towards Robustness against Unsuspicious Adversarial Examples}

\author{
Liang Tong\textsuperscript{\rm 1},
Minzhe Guo\textsuperscript{\rm 1},
Atul Prakash\textsuperscript{\rm 2},
Yevgeniy Vorobeychik\textsuperscript{\rm 1}\\
\textsuperscript{\rm 1}Washington University in St. Louis,  
\textsuperscript{\rm 2}University of Michigan  \\
\texttt{\{liangtong, guominzhe, yvorobeychik\}@wustl.edu} \\
\texttt{aprakash@umich.edu}
}

\iclrfinalcopy 
\begin{document}

\maketitle

\begin{abstract}
Despite the remarkable success of deep neural networks, significant concerns have emerged about their robustness to
adversarial perturbations to inputs.
While most attacks aim to ensure that these are
imperceptible, \emph{physical} perturbation attacks typically aim for
being unsuspicious, even if perceptible.
However, there is no universal notion of what it means for adversarial
examples to be unsuspicious.
We propose an approach for modeling suspiciousness by leveraging cognitive salience.
Specifically, we split an image into foreground (salient region) and
background (the rest), and allow significantly larger adversarial perturbations in
the background, while ensuring that cognitive salience of background
remains low.
We describe how to compute the resulting non-salience-preserving
dual-perturbation attacks on classifiers.
We then experimentally demonstrate that our attacks indeed do not
significantly change perceptual salience of the background, but are
highly effective against classifiers robust to conventional attacks.
Furthermore, we show that adversarial training with dual-perturbation
attacks yields classifiers that are more robust to these than state-of-the-art robust
learning approaches, and comparable in terms of robustness to
conventional attacks.
\end{abstract}

\section{Introduction}
\label{sec:introduction}

An observation by \citet{szegedy14intriguing} that
state-of-the-art deep neural networks that exhibit exceptional
performance in image classification are fragile in the face of small
adversarial perturbations of inputs has received a great deal of attention.
A series of approaches for designing adversarial examples followed~\citep{szegedy14intriguing, goodfellow15, carlini2017towards},
along with methods for defending against them~\citep{papernot2016distillation, madry2018towards}, and then new attacks
that defeat prior defenses, and so on.
Attacks can be roughly classified along three dimensions: 1)
introducing small $l_p$-norm-bounded perturbations, with the goal of
these being imperceptible to humans~\citep{madry2018towards}, 2) using non-$l_p$-based
constraints that capture perceptibility (often called
\emph{semantic perturbations})~\citep{bhattad2020unrestricted}, and 3) modifying physical objects, such
as stop signs~\citep{Eykholt2018RobustPA}, in a way that does not arouse suspicion.
One of the most common motivations for the study of adversarial
examples is safety and security, such as the potential for attackers to
compromise the safety of autonomous vehicles that rely on computer
vision~\citep{Eykholt2018RobustPA}.
However, while imperceptibility is certainly sufficient for
perturbations to be unsuspicious, it is far from necessary, as
physical attacks demonstrate.
On the other hand, while there are numerous formal definitions that
capture whether noise is perceptible~\citep{moosavi2016deepfool, carlini2017towards}, what makes adversarial examples
suspicious has been largely informal and subjective.

\begin{figure}
\centering
 \includegraphics[width=0.6\textwidth]{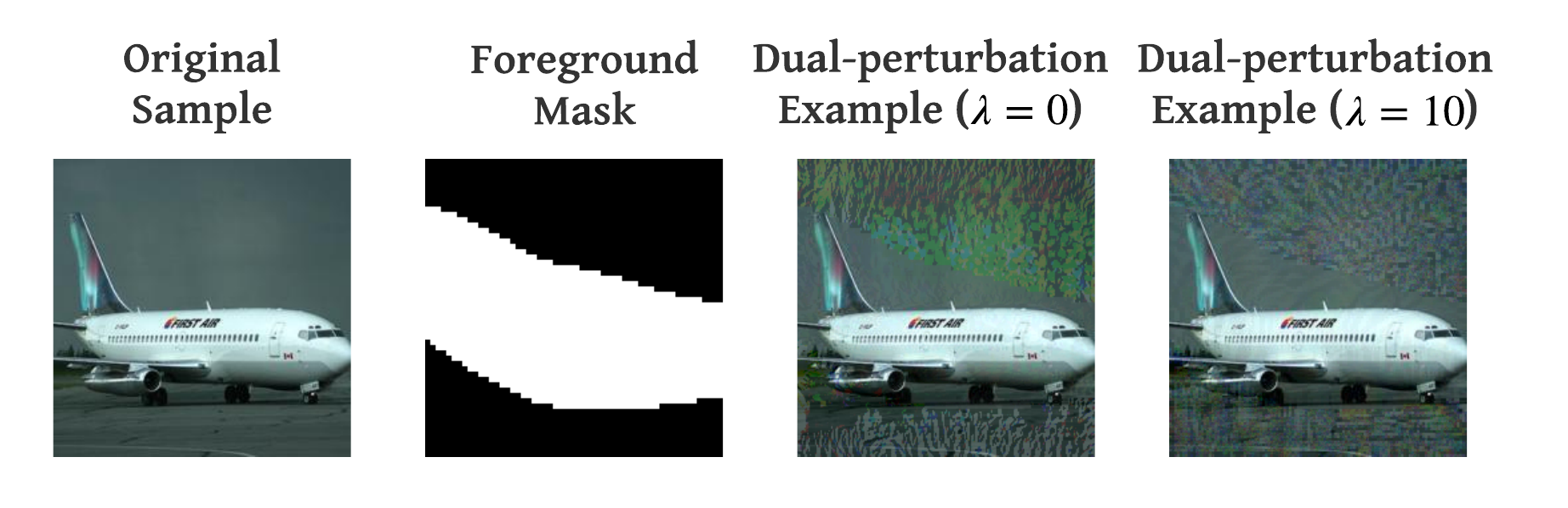} 
\caption{
An illustration of dual-perturbation attacks.
Adversarial examples are with large $\ell_\infty$ perturbations on the background ($\epsilon_B=20/255$) and small $\ell_\infty$ perturbations on the foreground ($\epsilon_F=4/255$).
A parameter $\lambda$ is used to control background salience explicitly.
A larger $\lambda$ results in less salient background under the same magnititude of perturbation. 
}
\label{fig:dual_pgd_example}
\end{figure}

We propose a simple formalization of an important aspect of what makes
adversarial perturbations unsuspicious.
Specifically, we make a distinction between image foreground and
background, allowing significantly more noise in the background than
the foreground.
This idea stems from the notion of cognitive salience~\citep{borji2015salient, kummerer2017understanding, he2018salient}, whereby an
image can be partitioned into the two respective regions to reflect
how much attention a human viewer pays to the different parts of the
captured scene.
In effect, we posit that perturbations in the foreground, when
visible, will arouse significantly more suspicion (by being
cognitively more salient) than perturbations made in the background.

Our first contribution is a formal model of such
\emph{dual-perturbation attacks}, which is a generalization of the
$l_p$-norm-bounded attack models (see, e.g.,
Figure~\ref{fig:dual_pgd_example}), but explicitly aims to ensure that
adversarial perturbation does not make the background highly salient.
Second, we propose an algorithm for finding adversarial examples using
this model, which is an adaptation of the PGD
attack~\citep{madry2018towards}.
Third, we present a method for defending against dual-perturbation
attacks based on the adversarial training framework~\citep{madry2018towards}.
Finally, we present an extensive experimental study that demonstrates that
(a) the proposed attacks are significantly stronger than PGD,
successfully defeating all state-of-the-art defenses, (b) proposed
defenses using our attack model significantly outperform
state-of-the-art alternatives, \emph{with relatively small performance
degradation on non-adversarial instances}, and (c) proposed defenses are
comparable to, or better than alternatives \emph{even against
  traditional attacks}, such as PGD.

\noindent{\bf Related Work: }
Recent studies have shown that neural networks are vulnerable to adversarial examples.
A variety of approaches have been proposed to produce adversarial examples~\citep{szegedy14intriguing, goodfellow15, papernot2016limitations, moosavi2016deepfool, carlini2017towards}.
These approaches commonly generate adversarial perturbations within a bounded $\ell_p$ norm so that the perturbations are imperceptible.  
A related thread has considered the problem of generating adversarial examples that are semantically imperceptible without being small in norm~\citep{brown2018unrestricted,bhattad2020unrestricted}, for example, through small perturbations to the color scheme.
However, none of these account for the perceptual distinction between the foreground and background of images.

Numerous approaches have been proposed for defending neural networks against adversarial examples~\citep{papernot2016distillation, carlini2017towards, madry2018towards, cohen2019certified,madry2018towards,Raghunathan18}.
Predominantly, these use $\ell_p$-bounded perturbations as the threat model, and while some account for semantic perturbations (e.g.~\cite{Mohapatra20}), none consider perceptually important difference in suspiciousness between foreground and background.

Two recent approaches by \citet{vaishnavi2019attention} and \citet{brama2020heat} have the strongest conceptual connection to our work.
Both are defense-focused by either eliminating~\citep{vaishnavi2019attention} or blurring~\citep{brama2020heat} the background region for robustness.
However, they \emph{assume} that we can reliably segment an image \emph{at prediction time}, leaving the approach vulnerable to attacks on image segmentation~\citep{arnab2018robustness}. 
\citet{xiao2020noise} propose to disentangle foreground and background signals on images but unsuspiciousness of their attacks is not ensured.

\section{Background}
\label{sec:background}


\subsection{Adversarial Examples and Attacks}

The problem of generating adversarial examples is commonly modeled as follows.
We are given a a learned model $h_{\bm{\theta}}(\cdot)$ parameterized by $\bm{\theta}$ which maps an input $\bm{x}$ to a $k$-dimensional prediction, where $k$ is the number of classes being predicted.
The final predicted class $y_p$ is obtained by $y_p = \argmax_i h_{\bm{\theta}}(\bm{x})_i$, where $h_{\bm{\theta}}(\bm{x})_i$ is the $i$th element of $h_{\bm{\theta}}(\bm{x})$.
Now, consider an input $\bm{x}$ along with a correct label $y$.
The problem of identifying an adversarial example for $\bm{x}$ can be captured by the following optimization problem:
\begin{equation}
\max_{\bm{\delta} \in \Delta(\epsilon)}\mathcal{L}\left(h_{\bm{\theta}}(\bm{x}+\bm{\delta}), y\right),
\label{eq:adv_example}
\end{equation}
where $\mathcal{L}(\cdot)$ is the adversary's utility function (for example, the loss function used to train the classifier $h_{\bm{\theta}}$).
$\Delta(\epsilon)$ is the feasible perturbation space which is commonly represented as a $\ell_p$ ball:
$\Delta(\epsilon)=\left\{\bm{\delta} :\|\bm{\delta}\|_{p} \leq \epsilon\right\}$.

A number of approaches have been proposed to solve the optimization problem shown in Eq.~(\ref{eq:adv_example}), among which two are viewed as state of the art:  \emph{CW attack} developed by \citet{carlini2017towards}, and \emph{Projected Gradient Descent (PGD) attack} proposed in \citet{madry2018towards}.
In this work, we focus on the PGD attack with $\ell_\infty$ and $\ell_2$ as the distance metrics. 

\subsection{Robust Learning}


An important defense approach that has proved empirically effective even against adaptive attacks is \emph{adversarial training}~\citep{szegedy14intriguing, cohen2019certified, goodfellow15, madry2018towards}.
The basic idea of adversarial training is to produce adversarial examples and incorporate these into the training process.
Formally, adversarial training aims to solve the following robust learning problem:
\begin{equation}
\underset{\bm{\theta}}{\min} \frac{1}{|D|} \sum_{\bm{x}, y \in D} \max _{\|\bm{\delta}\|_{p} \leq \epsilon} \mathcal{L}\left(h_{\bm{\theta}} (\bm{x}+\bm{\delta}), y\right),
\label{eq:adv_training}
\end{equation}
where $D$ is the training dataset. 
In practice, this problem is commonly solved by iteratively using the following two steps~\citep{madry2018towards}: 1) use a PGD (or other) attack to produce adversarial examples of the training data; 2) use any optimizer to minimize the loss of those adversarial examples. 
It has been shown that adversarial training can significantly boost the adversarial robustness of a classifier against $\ell_p$ attacks, and it can be scaled to neural networks with complex architectures.

%

\section{Dual-Perturbation Attacks}
\label{sec:threat_model}


\subsection{Motivation}

\begin{figure}
\centering
 \includegraphics[width=0.45\textwidth]{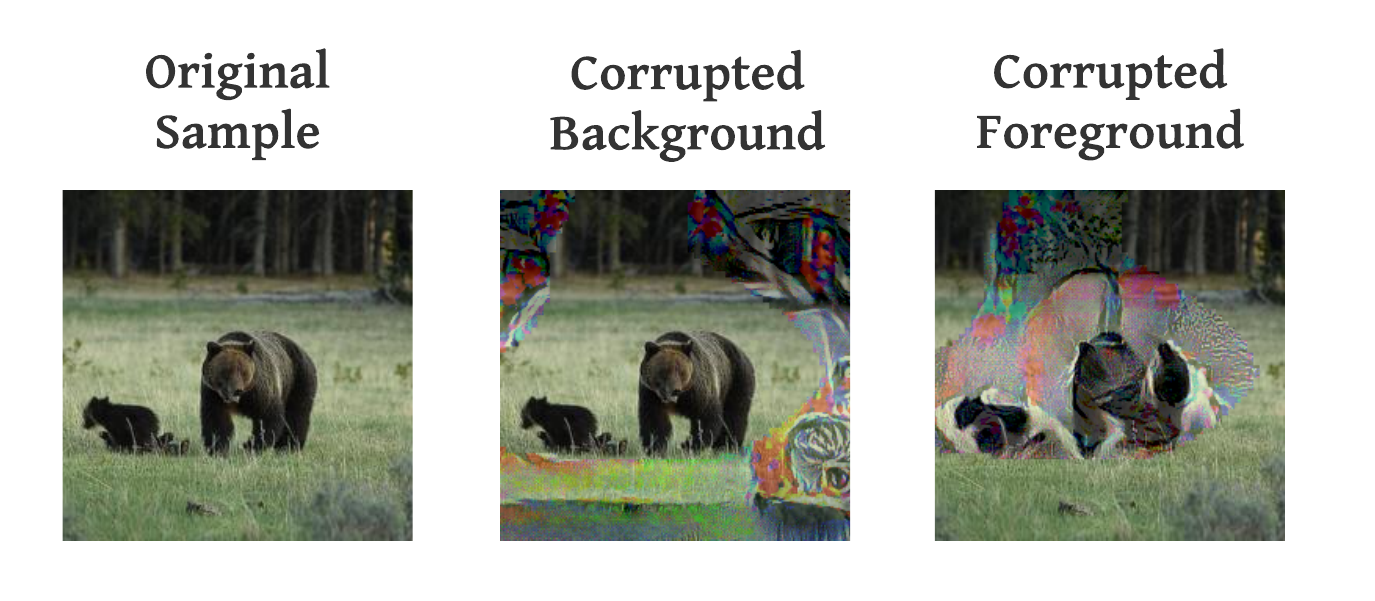} 
\caption{
Semantic distinction between foreground and background. Left: Original image of bears. Middle: Adversarial example with $\ell_\infty$ bounded perturbations ($\epsilon=40/255$) on the background, the sematic meaning (bear) is preserved. Right: Adversarial example with $\ell_\infty$ bounded perturbations ($\epsilon= 40/255$) on the foreground, with more ambiguous semantics.
}
\label{fig:semantics}
\end{figure}

Our threat model is motivated by the \emph{feature integration theory}~\citep{treisman1980feature} in cognitive science: regions that have features that are different from their surroundings are more likely to catch a viewer's gaze.
Such regions are called \emph{salient regions}, or \emph{foreground}, while the others are called \emph{background}.
Accordingly, for a given image, the semantics of the object of interest is more likely to be preserved in the foreground, as it catches more visual attention of a viewer compared to the background.
If the foreground of an image is corrupted, then the semantics of the object of interest is broken.
In contrast, the same extent of corruption in the background nevertheless preserves the overall semantic meaning of the scene captured (see, e.g., Figure~\ref{fig:semantics}).
Indeed, detection of salient regions, as well as the segmentation of foreground and background, have been extensively studied in computer vision~\citep{borji2015salient}.
These approaches either predict human fixations, which are sparse bubble-like salient regions sampled from a distribution ~\citep{kummerer2017understanding}, or salient objects that contain smooth connected areas in an image~\citep{he2018salient}.

Despite this important cognitive distinction between foreground and background, essentially all of the attacks on deep neural networks for image classification make no such distinction, even though a number of other semantic factors have been considered~\citep{bhattad2020unrestricted,Mohapatra20}.
Rather, much of the focus has been on adversarial perturbations that are \emph{not noticeable} to a human, but which are applied equally \emph{to the entire image}.
However, in security applications, the important issue is not merely that an attack cannot be noticed, but that whatever observed is \emph{not suspicious}.
This is, indeed, the frame of reference for many high-profile \emph{physical} attacks on image classification, which are clearly visible, but not suspicious because they hide in the ``human psyche'', that is, are easily ignored~\citep{Sharif16,Eykholt2018RobustPA}.
The main goal of the threat model we introduce next is therefore to capture more precisely the notion that an adversarial example is not suspicious by leveraging the cognitive distinction between foreground and background of an image.

\subsection{Dual-Perturbation Attacks}
\label{subsec:dual}

At the high level, our proposed threat model involves producing small (imperceptible) adversarial perturbations in the foreground of an image, and larger perturbations in the background.
This can be done by incorporating state-of-the-art attacks into our method:
we can use one attack with small $\epsilon$ in the foreground, and another with a large $\epsilon$ in the background.
Consequently, we term our approach \emph{dual-perturbation attacks}.
Note that these clearly generalize the standard small-norm (e.g., PGD) attacks, since we can set the $\epsilon$ to be identical in both the foreground and background.
However, the key consideration is that after we add the large amount of noise to the background, \emph{we must ensure that we do not thereby make it highly salient to the viewer}.
We capture this second objective by including in the optimization problem a \emph{salience} term that decreases with increasing salience of the background.

Formally, the \emph{dual-perturbation} attack solves the following optimization problem:
\begin{equation}
\max_{||\bm{\delta} \circ \mathcal{F}(\bm{x})||_p \leq \epsilon_F,\\ ||\bm{\delta}\circ\mathcal{B}(\bm{x})||_p \leq \epsilon_B} \mathcal{L}\left(h_{\bm{\theta}}(\bm{x}+\bm{\delta}), y\right) + \lambda \cdot \mathcal{S} \left( \bm{x}+\bm{\delta} \right),
\label{eq:dual_pgd}
\end{equation}
where $\mathcal{S}\left( \bm{x}+\bm{\delta} \right)$ measure the relative salience of the foreground compared to background after adversarial noise $\bm{\delta}$ has been added, with $\lambda$ a parameter that explicitly balances the two objectives: maximizing predicted loss on adversarial examples, and limiting background salience (compared to foreground) so that the adversarial example produced is unsuspicious.
Here $\mathcal{F}$ returns the mask matrix constraining the area of the perturbation in the foreground, and $\mathcal{B}$ returns the mask matrix restricting the area of the perturbation in the background, for an input image $\bm{x}$.
$\mathcal{F}(\bm{x})$ and $\mathcal{B}(\bm{x})$ have the same dimension as $\bm{x}$ and contain 1s in the area which can be perturbed and 0s elsewhere.
$\circ$ denotes element-wise multiplication for matrices.
Hence, we have $\bm{x} = \mathcal{F}(\bm{x})+\mathcal{B}(\bm{x})$ which indicates that any input image can be decomposed into two independent images: one containing just the foreground, and the other containing the background.

We model the suspiciousness $\mathcal{S}(\bm{x})$ of an input image $\bm{x}$ by leveraging a recent computational model of image salience, DeepGaze II~\citep{kummerer2017understanding}.
DeepGaze II outputs predicted pixel-level density of human fixations on an image with the total density over the entire image summing to 1.
Our measure of relative salience of the foreground to background is the \emph{foreground score}, which is defined as $\mathcal{S}(\bm{x}) = \sum_{i \in \{k|\mathcal{F}(\bm{x})_k \neq 0\}} s_i$, where $s_i$ is the saliency score produced by DeepGaze II for pixel $i$ of image $\bm{x}$.
Since foreground, as a fraction of the image, tends to be around 50-60\%, a score significantly higher than 0.5 indicates that predicted human fixation is relatively localized to the foreground.

A natural approach for solving the optimization problem shown in Equation~\ref{eq:dual_pgd} is to apply an iterative method, such as the PGD attack.
However, the use of this approach poses two challenges in our setting.
First, as in the PGD attack, the problem is non-convex, and PGD only converges to a local optimum. 
We can address this issue by using \emph{random starts}, i.e.,~by randomly initializing the starting point of the adversarial perturbations, as in~\citet{madry2018towards}.
Second, and unlike PGD, the optimization problem in Equation~\ref{eq:dual_pgd} involves \emph{two hard constraints} $||\bm{\delta}\circ\mathcal{F}(\bm{x})||_p \leq \epsilon_F$ and $||\bm{\delta}\circ\mathcal{B}(\bm{x})||_p \leq \epsilon_B$.
Thus, the feasible region of the adversarial perturbation $\bm{\delta}$ is not an $\ell_p$ ball, which makes computing the projection $\mathcal{P}_{\epsilon}$ computationally challenging in high-dimensional settings.
To address this challenge, we split the \emph{dual-perturbation} attack into two individual processes in each iteration, one for the adversarial perturbation in the foreground and the other for the background, and then merge these two perturbations when computing the gradients, like a standard PGD attack.
Full details of our algorithms for computing dual perturbation examples are provided in Appendix A.

Now, the question that remains is how to partition an input image $\bm{x}$ into foreground, $\mathcal{F}(\bm{x})$, and background, $\mathcal{B}(\bm{x})$.
We address this next.

\subsection{Identifying Foreground and Background}

Given an input $\bm{x}$, we aim to compute $\mathcal{F}(\bm{x})$, the foreground mask and $\mathcal{B}(\bm{x})$, the background mask.
We consider two approaches for this: fixation prediction and segmentation.


Our first method leverages the fixation prediction approach~\citep{kummerer2017understanding} to identify foreground and background.
This enables a general approach for foreground-background partition as fixation predictions are not limited to any specific collection of objects.
Specifically, we first use DeepGaze II~\citep{kummerer2017understanding} to output predicted pixel-level density of human fixations on an image.
We then divide the image into foreground and background by setting a threshold $t = 0.5\cdot(s_{min}(\bm{x})+s_{max}(\bm{x}))$ for each input image $\bm{x}$ where $(s_{min}, s_{max})$ are the minimum and maximum values of human fixation on pixels of $\bm{x}$.
Pixels with larger values than $t$ are grouped into the foreground, and the others are identified as background subsequently.

Our second approach is to make use of semantic segmentation to provide a partition of the foreground and background in pixel level.
This can be done in two steps:
First, we use state-of-the-art paradigms for semantic segmentation (e.g., \citet{long2015fully}) to identify pixels that belong to each corresponding object, as there might be multiple objects in an image.
Next, we identify the pixels that belong to the object of interest as the foreground pixels, and the others as background pixels.

We use both of the above approaches in dual-perturbation attacks when evaluating the robustness of classifiers, as well as designing robust models.
More details are available in Section~\ref{sec:experiments}.

\section{Defense against Dual-Perturbation Attacks}
\label{sec:defense_approach}

Once we are able to compute the dual-perturbation attack, we can incorporate it into conventional adversarial training paradigms for defense,
as it has been demonstrated that adversarial training is highly effective in designing classification models that are robust to a given attack.
Specifically, we replace the PGD attack in the adversarial training framework proposed by~\citet{madry2018towards}, with the proposed dual-perturbation attack.
We term this approach \emph{AT-Dual}, which aims to solve the following optimization problem:
\begin{equation}
\underset{\bm{\theta}}{\min} \frac{1}{|D|} \sum_{\bm{x}, y \in D} \max _{\substack{||\bm{\delta} \circ \mathcal{F}(\bm{x})||_p \leq \epsilon_F,\\ ||\bm{\delta}\circ\mathcal{B}(\bm{x})||_p \leq \epsilon_B
}} \mathcal{L}\left(h_{\bm{\theta}} (\bm{x}+\bm{\delta}), y\right)  + \lambda \cdot \mathcal{S} \left( \bm{x}+\bm{\delta} \right).
\label{eq:at_dual}
\end{equation}


Note that \emph{AT-Dual} needs to identify background and foreground for any input when solving the inner maximization problems in Equation~\ref{eq:at_dual} at training time.
At prediction time,  our approaches classify test samples like any standard classifiers, which is independent of the semantic partitions so as to close the backdoors to attacks on object detection approaches~\citep{xie2017adversarial}.
We evaluate the effectiveness of our approaches in Section~\ref{sec:experiments}.

\section{Experiments}
\label{sec:experiments}


\subsection{Experimental Setup}

{\bf Datasets}.
We conducted the experiments on the following three datasets (detailed in Appendix B):
The first is Segment-6~\citep{cong2019masked}, which are images with $32\times32$ pixels obtained by pre-processing the Microsoft COCO dataset~\citep{lin2014microsoft} to make it compatible with image classification tasks.
We directly used the semantic segmentation based foreground masks provided in this dataset.
Our second dataset is STL-10, a subset that contains images with $96\times96$ pixels.
Our third dataset is ImageNet-10, a 10-class subset of the ImageNet dataset~\citep{deng2019imagenet}.
We cropped all its images to be with $224\times224$ pixels.
For STL-10 and ImageNet-10, we used fixation prediction to identify foreground and background as described in Section~\ref{sec:threat_model}.

{\bf Baselines}.
We consider \emph{PGD} attack as a baseline adversarial model, and \emph{Adversarial Training with PGD Attacks} as a baseline robust classifier.
We also consider a classifier trained on non-adversarial data (henceforth, \emph{Clean}).
Additionally, we consider \emph{Randomized Smoothing}~\citep{cohen2019certified} and defer the corresponding results to Appendix J.


{\bf Evaluation Metrics}.
We use two standard evaluation metrics for both attacks and defenses:
1) accuracy of prediction on clean test data where no adversarial attacks were attempted. 
2) adversarial accuracy, which is accuracy when adversarial inputs are used in place of clean inputs.

Throughout our evaluation, we used both $\ell_2$ and $\ell_\infty$ norms to measure the magnitude of added adversarial perturbations.
Due to space limitations, we only present experimental results of the \emph{Clean} model and classification models that are trained to be robust to $\ell_2$ norm attacks using the ImageNet-10 dataset. 
The results for $\ell_\infty$ norm and other datasets are similar and deferred to Appendix.

In the following experiments, all classifiers were trained with 20 epochs on a ResNet34 model~\citep{he2016deep} pre-trained on ImageNet and with a customized final fully connected layer. 
Specifically, we trained AT-PGD by using 50 steps of $\ell_2$ PGD attack with $\epsilon=2.0$, and AT-Dual by using 50 steps of $\ell_2$ dual-perturbation attack with $\{\epsilon_F, \epsilon_B,\lambda\} = \{2.0, 20.0, 0.0\}$ at each training epoch.
At test time, we used both $\ell_2$ PGD and dual-perturbation attacks with 100 steps to evaluate robustness.

\subsection{Saliency Analysis of Dual-Perturbation Adversarial Examples}

We begin by considering a natural question: is our particular distinction between foreground and background actually consistent with cognitive salience?
In fact, this gives rise to two distinct considerations: 1) whether foreground as we identify it is in fact significantly more salient than the background, and 2) if so, whether background becomes significantly more salient \emph{as a result of our dual-perturbation attacks}.
We answer both of these questions by appealing to DeepGaze II~\citep{kummerer2017understanding} to compute the \emph{foreground score (FS)} of dual-perturbation examples as described in Section~\ref{sec:threat_model}, and using the accuracy of different classifiers on dual-perturbation examples with different background salience.   


\begin{figure}[h]
\centering
\begin{tabular}{cc}
  \includegraphics[width=0.48\textwidth]{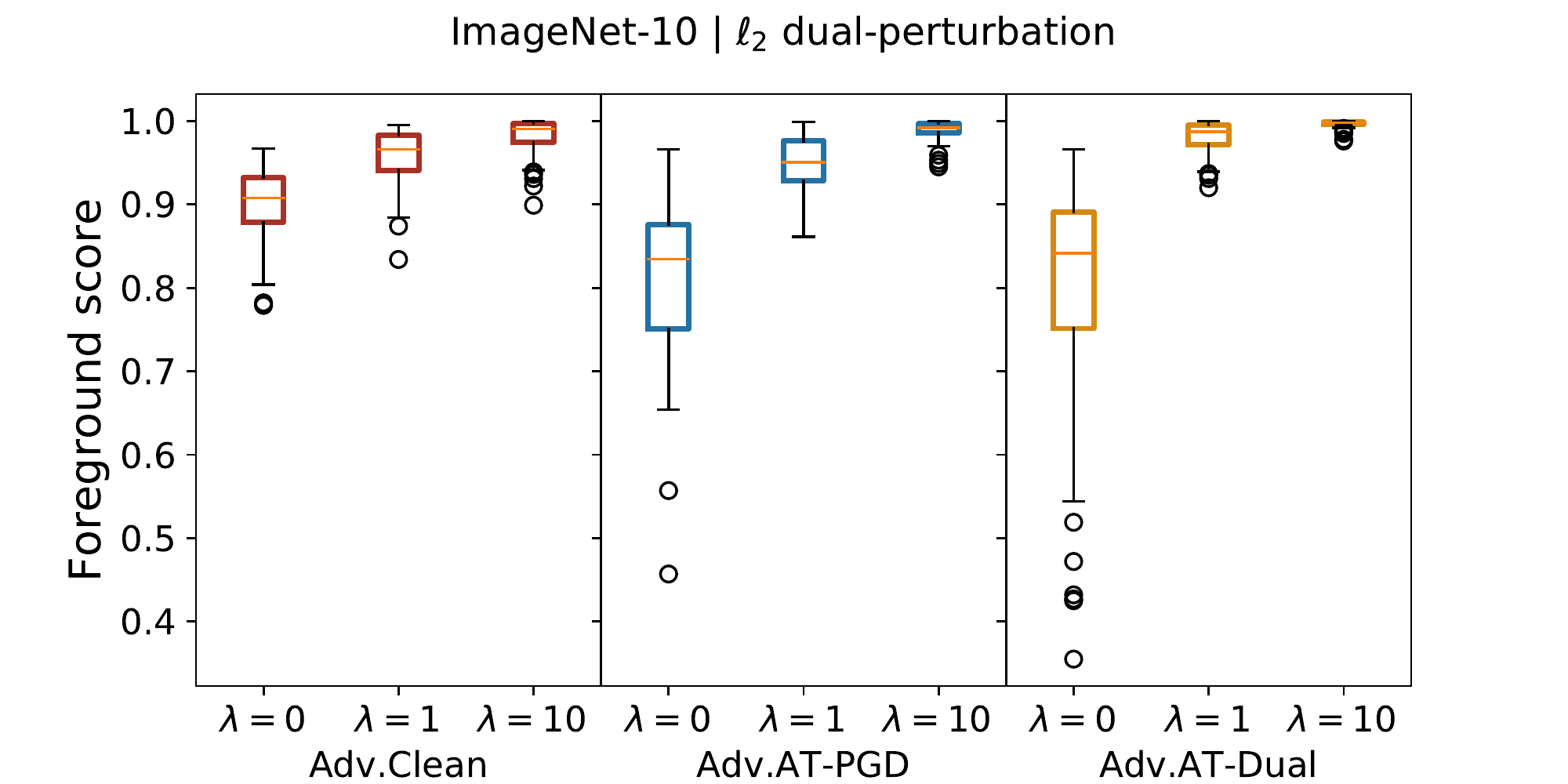} &
 \includegraphics[width=0.24\textwidth]{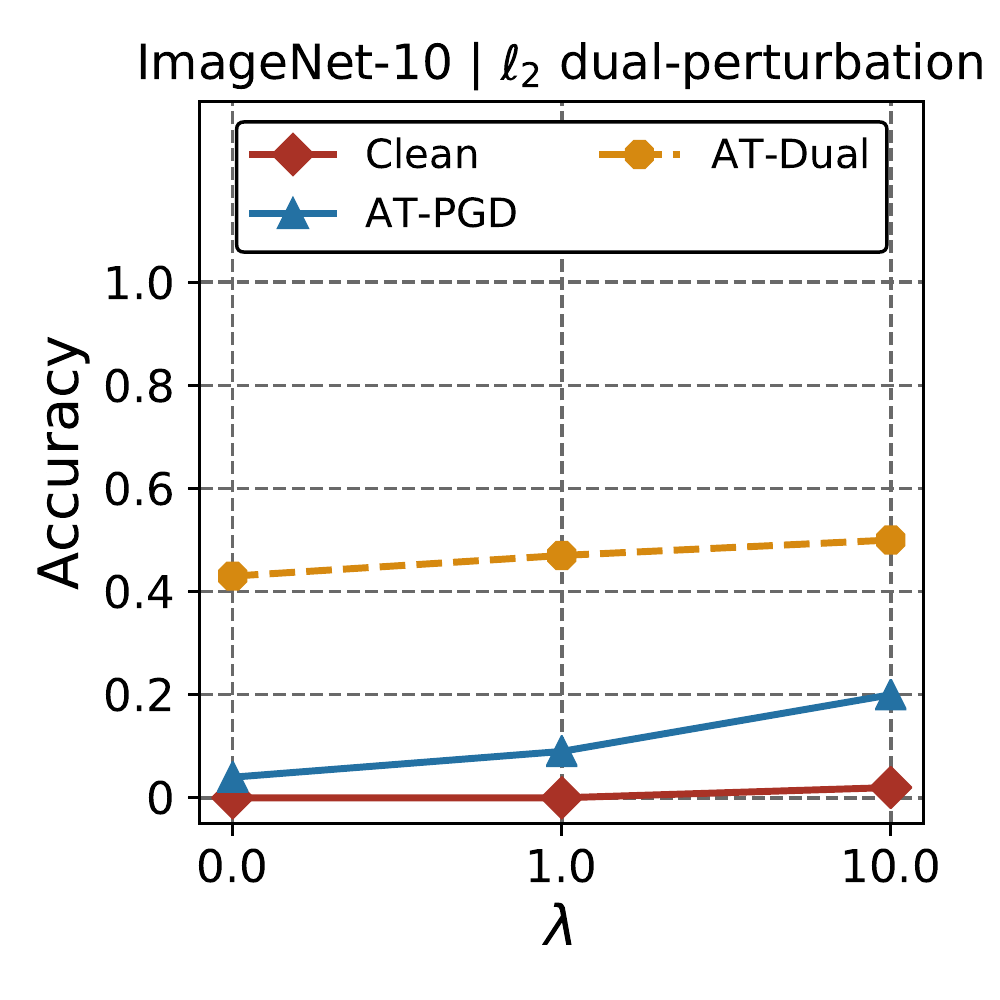}\\
\end{tabular}
	\caption{
	Saliency analysis. 
	Dual-perturbation attacks are performed by using $\{\epsilon_F, \epsilon_B\}=\{2.0, 20.0\}$ and a variety of $\lambda$ displayed in the figure. 
	Left: foreground scores of dual-perturbation examples in response to different classifiers.
	Right: accuracy of classifiers on dual-perturbation examples with salience control.  	
	}
	\label{fig:saliency_analysis}
\end{figure}


Figure~\ref{fig:saliency_analysis} presents the answer to both of the questions above.
First, observe that in Figure~\ref{fig:saliency_analysis}, \emph{FS} (vertical axis) is typically well above 0.5, and in most cases above 0.9, for all attacks.
Second, this is true whether we attack the \emph{Clean} model, or either \emph{AT-PGD} or \emph{AT-Dual} robust models.
Particularly noteworthy, however, is the impact that the parameter $\lambda$ has on the \emph{FS}, especially when robust classifiers are employed.
Recall that $\lambda$ reflects the relative importance of salience in generating adversarial examples, with larger values forcing our approach to pay more attention to preserving unsuspiciousness of background relative to foreground.
As we increase $\lambda$, we note significantly higher \emph{FS}, i.e., lower background salience (again, Figure~\ref{fig:saliency_analysis}, left).
Figure~\ref{fig:dual_pgd_example} offers a visual illustration of this effect.

As significantly, Figure~\ref{fig:saliency_analysis} (right) shows that moderately increasing $\lambda$ does not significantly reduce the effectiveness of the attack, on either the \emph{Clean} or the robust classifiers.

\subsection{Dual-perturbation Attacks on Robust Classifiers}

Next, we evaluate the effectiveness of dual-perturbation attacks against state-of-the-art robust learning methods, as well as the effectiveness of adversarial training that uses dual-perturbation attacks for generating adversarial examples.
We begin by considering white-box attacks, and subsequently evaluate transferability.
Due to space limitations, we defer the results of transferability to Appendix D.

\begin{figure}[t]
\centering
\begin{tabular}{ccc}
  \includegraphics[width=0.26\textwidth]{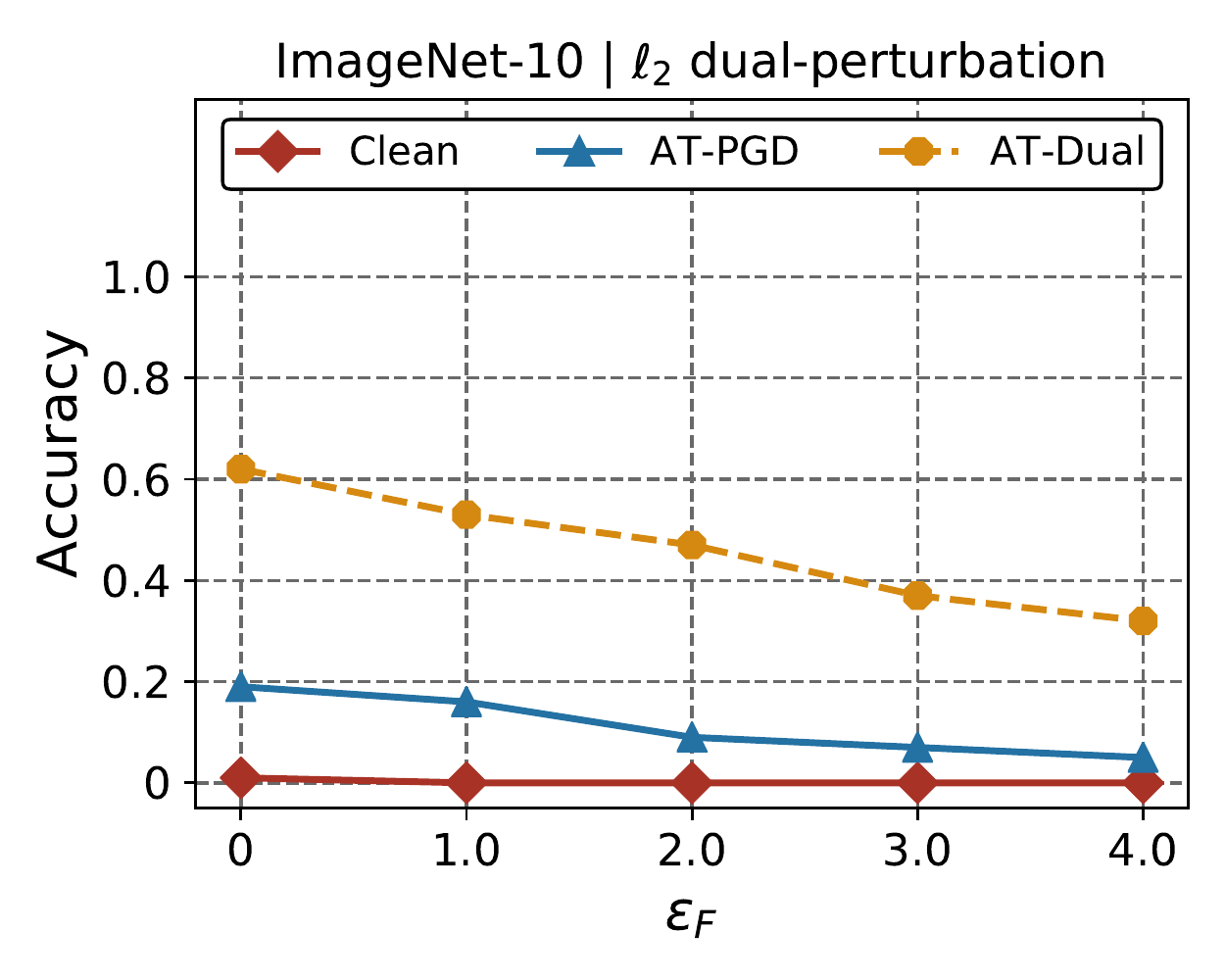} &
  \includegraphics[width=0.26\textwidth]{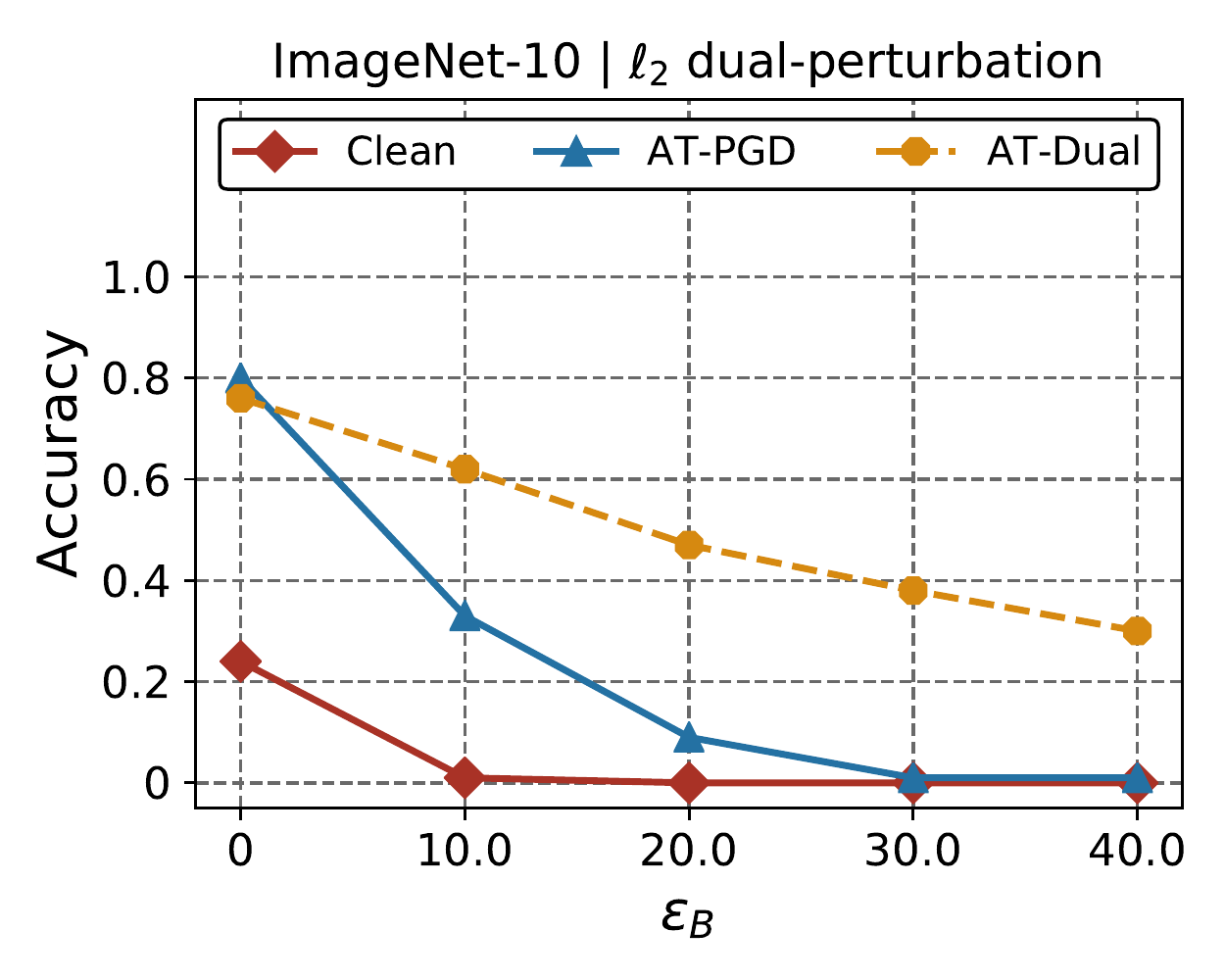} &
  \includegraphics[width=0.26\textwidth]{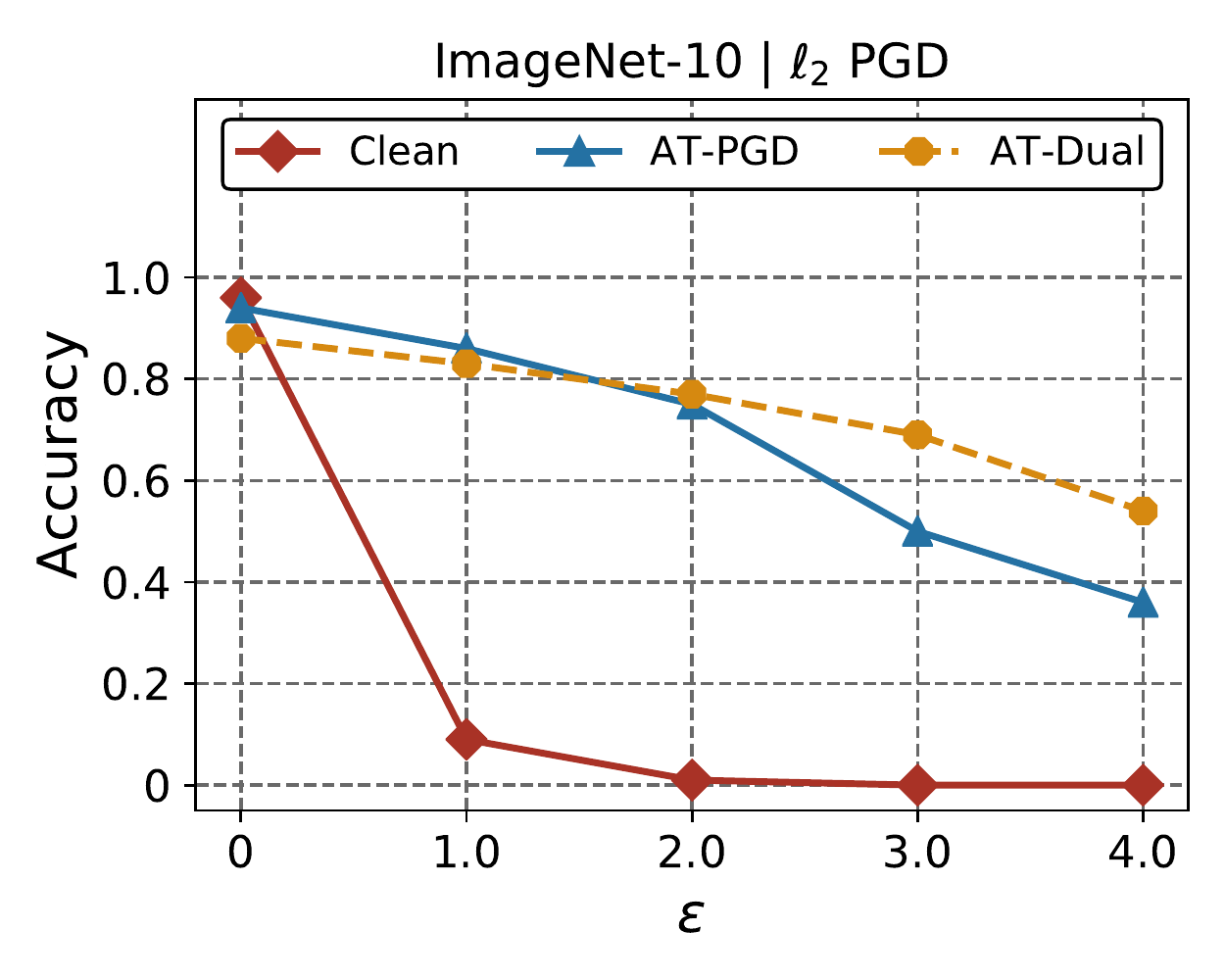}\\
\end{tabular}
\caption{
Robustness to white-box $\ell_2$ attacks on ImageNet-10.
Left: dual-perturbation attacks with different foreground distortions. $\epsilon_B$ is fixed to be 20.0 and $\lambda=1.0$.
Middle:dual-perturbation attacks with different background distortions. $\epsilon_F$ is fixed to be 2.0 and $\lambda=1.0$.
Right: PGD attacks. 
}
\label{F:l2attacks}
\end{figure}

The results for white-box attacks are presented in Figure~\ref{F:l2attacks}.
First, consider the dual-perturbation attacks (left and middle plots).
Note that in all cases these attacks are highly successful against the baseline robust classifier (AT-PGD); indeed, even relatively small levels of foreground noise yield near-zero accuracy when accompanied by sufficiently large background perturbations.
For example, when the perturbation to the foreground is $\epsilon_F=2.0$ and background perturbation is $\epsilon_B=20.0$, \emph{AT-PGD} achieves robust accuracy below $10\%$.
In contrast, AT-Dual remains significantly more robust, with an improvement of up to $40\%$ compared to the baseline.
Second, consider the standard PGD attacks (right plot).
It can be observed that all of the robust models are successful against the $\ell_2$ PGD attacks.
However, our defense exhibit moderately higher robustness than the baselines under large distortions of PGD attacks, without sacrificing much in accuracy on clean data.
For example, when the perturbation of the $\ell_2$ PGD attack is above $\epsilon=3.0$, \emph{AT-Dual} can achieve 20\% more accuracy.

\subsection{Generalizability of Defense}

It has been observed that models robust against $l_p$-norm-bounded attacks for one value of $p$ can be fragile when facing attacks with a different norm $l_{p'}$~\citep{sharma2018attacking}.
\begin{figure}[t]
\centering
\begin{tabular}{cccc}
  \includegraphics[width=0.22\textwidth]{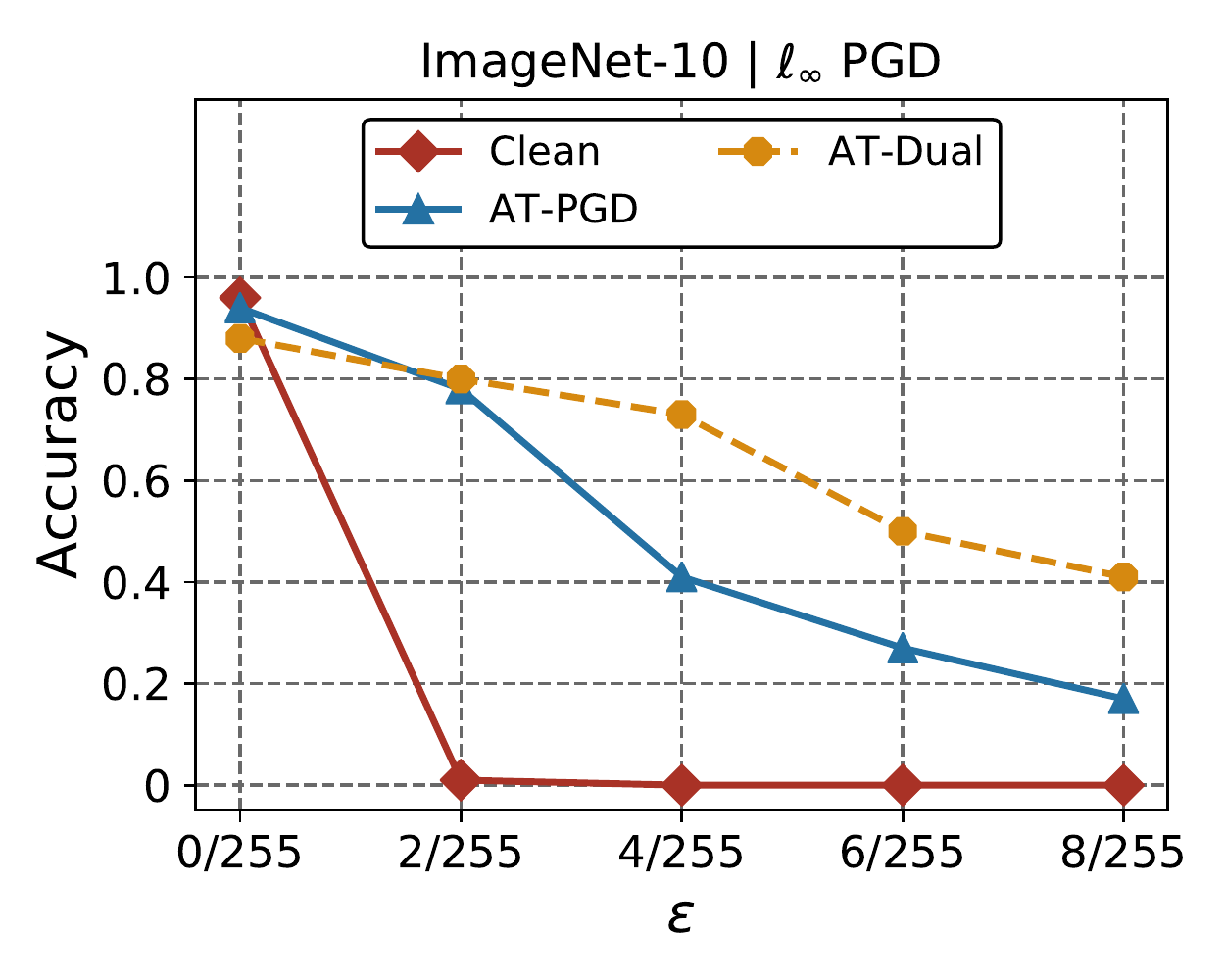} &
 \includegraphics[width=0.22\textwidth]{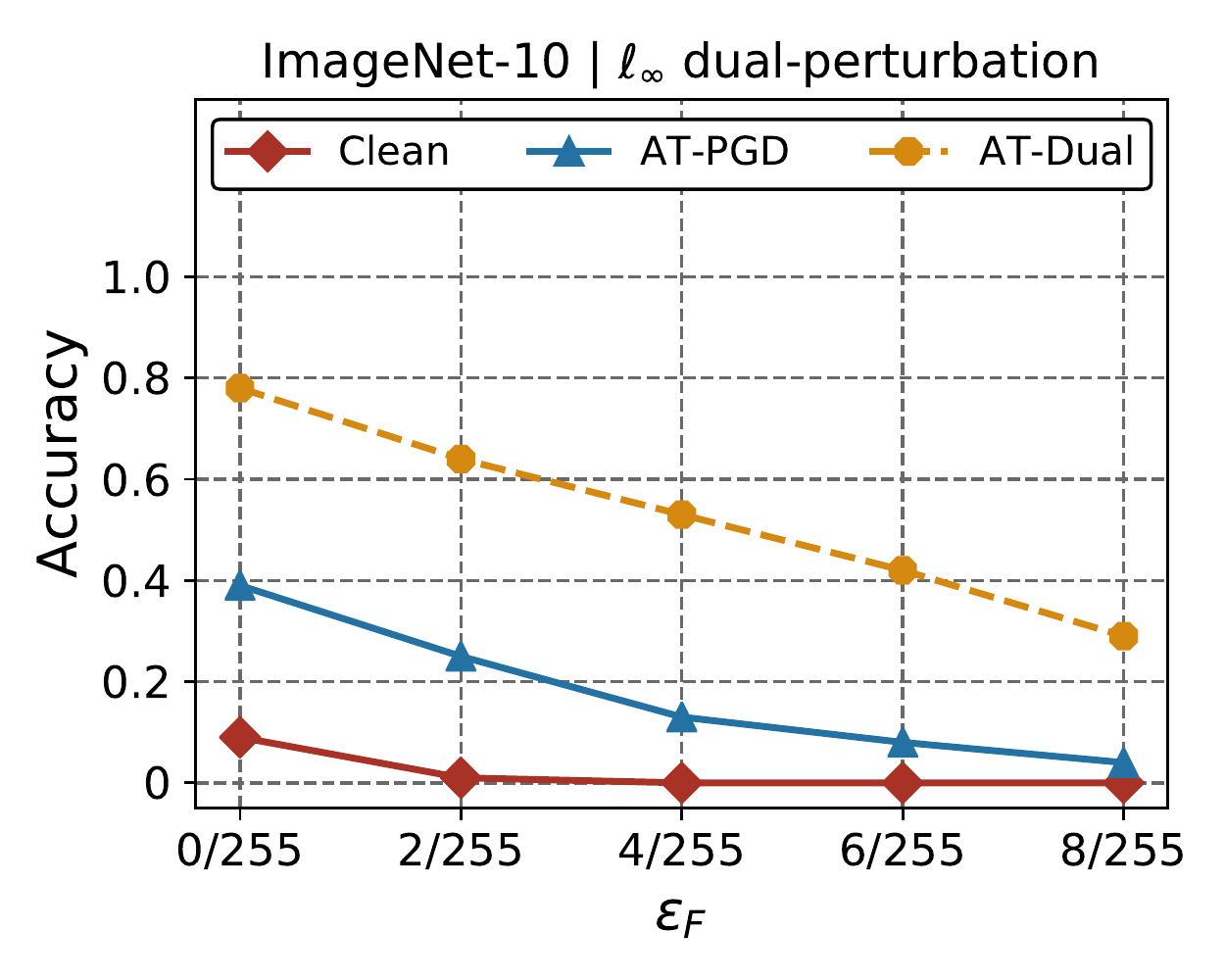} &
  \includegraphics[width=0.22\textwidth]{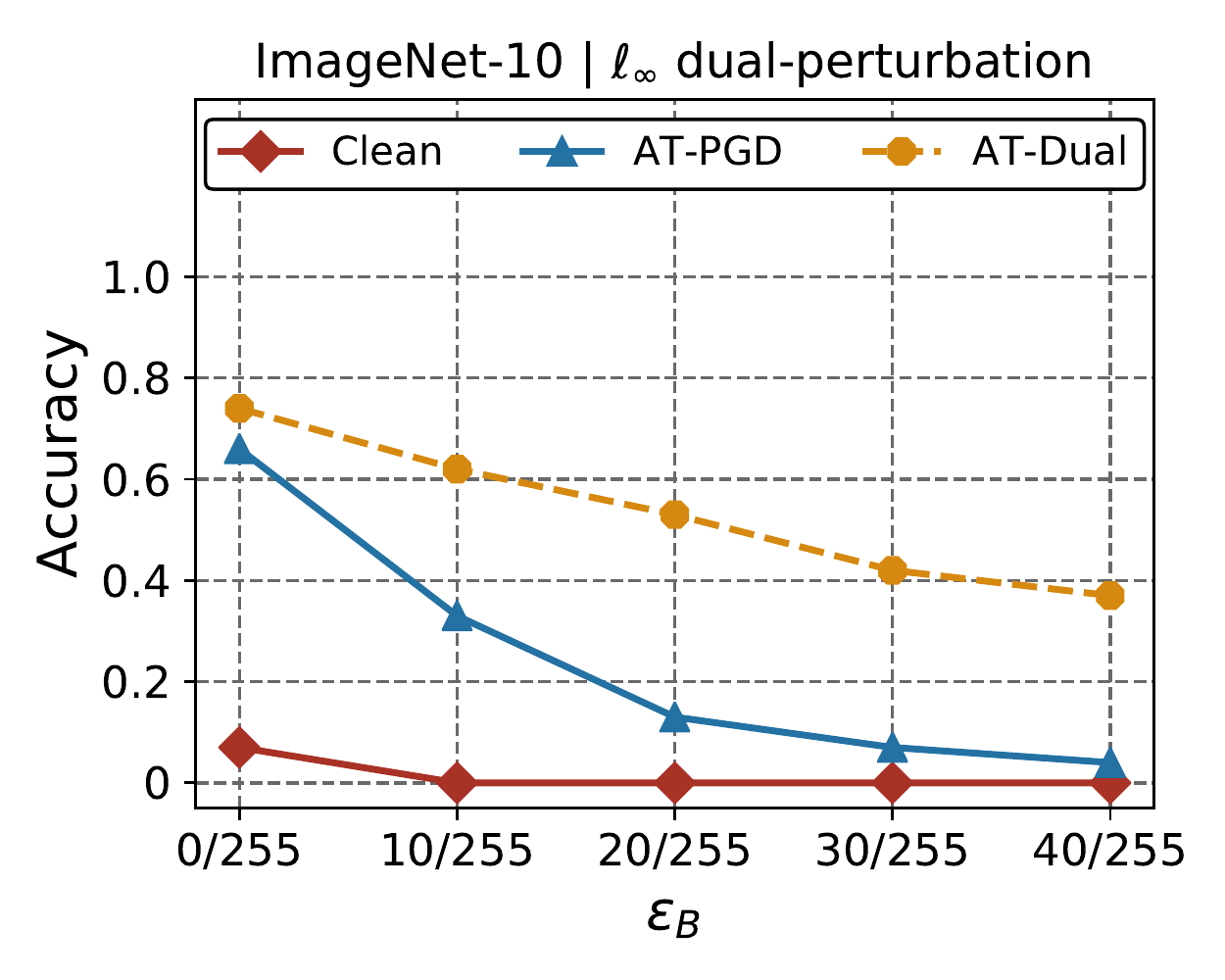} &
 \includegraphics[width=0.22\textwidth]{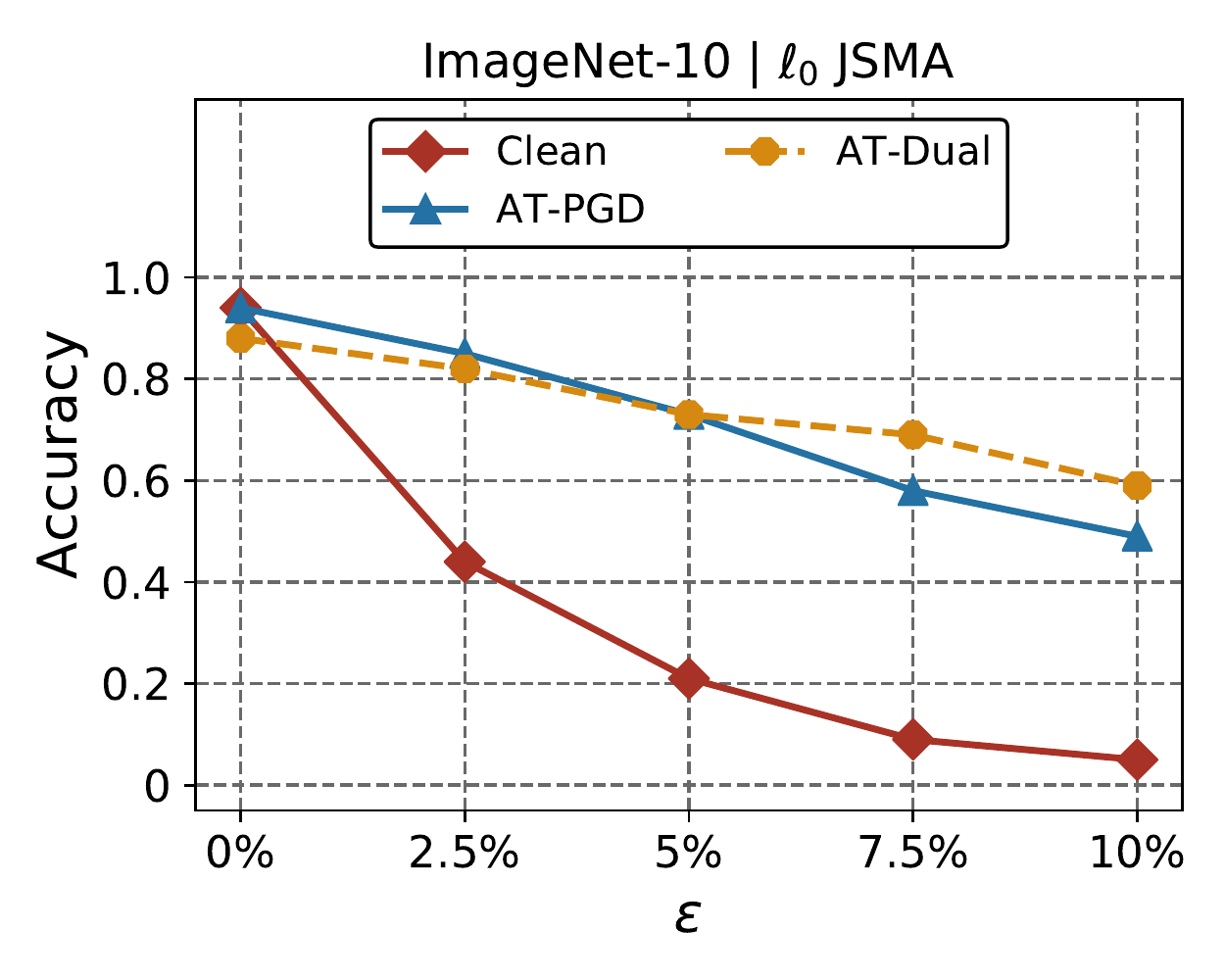} \\
\end{tabular}
	\caption{Robustness to additional white-box attacks on ImageNet-10. 
	Left: 20 steps of $\ell_\infty$ PGD attacks. 
	Middle left: 20 steps of $\ell_\infty$ dual-perturbation attacks with different foreground distortions. $\epsilon_B$ is fixed to be 20/255 and $\lambda=1.0$.
	Middle right: 20 steps of $\ell_\infty$ dual-perturbation attacks with different background distortions. $\epsilon_F$ is fixed to be 4/255 and $\lambda=1.0$.
	Right: $\ell_0$ JSMA attacks.
}
	\label{F:generalizability}
\end{figure}
Here, our final goal is to present evidence that the approaches for defense based on dual-perturbation attacks remain relatively robust even when faced with attacks generated using different norms.
Here, we show this when our models are trained using the $l_2$-bounded attacks, and evaluated against other attacks using other norms.
The results are presented in Figure~\ref{F:generalizability}.
We consider three alternative attacks: 1) PGD using the $l_\infty$-bounded perturbations, as in \citet{madry2018towards} (left in Figure~\ref{F:generalizability}) 2) dual-perturbation attacks with $l_\infty$-norm bounds (middle left and middle rigt in Figure~\ref{F:generalizability}), and 3) JSMA, a $l_0$-bounded attack~\citep{papernot2016limitations} (right in Figure~\ref{F:generalizability}).
We additionally considered $l_2$ attacks, per Carlini and Wagner~\citep{carlini2017towards}, but find that all of the robust models, whether based on PGD or dual-perturbation attacks, are successful against these.

Our first observation is that \emph{AT-Dual} is significantly more robust to $l_\infty$-bounded PGD attacks than the adversarial training approach in which adversarial examples are generated using $l_2$-bounded PGD attacks (Figure~\ref{F:generalizability} (left)).
Consequently, training with dual-perturbation attacks already exhibits better ability to generalize to other attacks compared to conventional adversarial training.

The gap between dual-perturbation-based adversarial training and standard adversarial training is even more significant when we consider $l_\infty$ dual-perturbation attacks (middle left and middle right figures of Figure~\ref{F:generalizability}).
Here, we see that robustness of PGD-based adversarially trained model is only marginally better than that of a clean model under large distortions (e.g., when $\epsilon_B\geq 20/255$ in the middle right plot of Figure~\ref{F:generalizability}), whereas \emph{AT-Dual} remains relatively robust.


Finally, considering JSMA attacks (see Figure~\ref{F:generalizability} (right)), we can observe that both \emph{AT-Dual} and \emph{AT-PGD} remain relatively robust.
However, a deeper look at Figure~\ref{F:generalizability} (right) reveals that compared to \emph{AT-PGD}, \emph{AT-Dual} exhibit moderately higher robustness than the baselines under large distortions of JSMA attacks.
Overall, in all of the cases, the model made robust using dual-perturbation attacks remains quite robust even as we evaluate against a different attack, using a different norm.

\subsection{Analysis of Defense}

\begin{figure}[t]
\centering
 \includegraphics[width=0.50\textwidth]{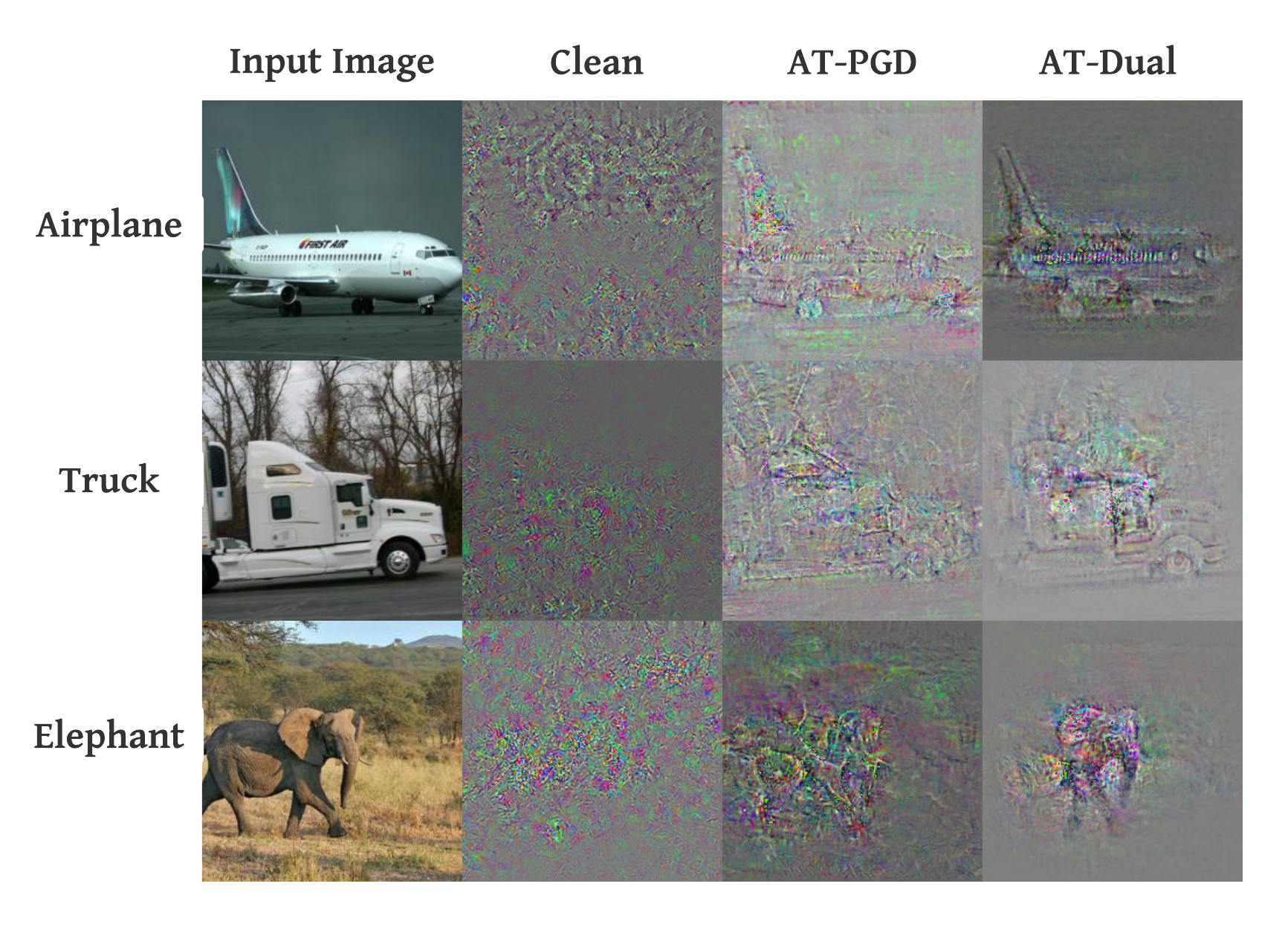} 
\caption{
Visualization of loss gradient of different classifiers with respect to pixels of \emph{non-adversarial} inputs.
}
\label{F:visualization}
\end{figure}

Finally, we conduct an exploratory experiment to study adversarial robustness by investigating which pixel-level features are important for different classifiers at prediction time .
To do this, we visualize the loss gradient of different classifiers with respect to pixels of the same \emph{non-adversarial} inputs (as introduced in ~\citet{tsipras2019robustness}), shown in Figure~\ref{F:visualization}.
Our first observation is that the gradients in response to adversarially robust classifiers (AT-PGD and AT-Dual) align well with human perception, while a standard training model (Clean) results in a noisy gradient for the input images. 
Second, compared to adversarial training with the conventional PGD attack (AT-PGD), the loss gradient of AT-Dual provides significantly better alignment with sharper foreground edges and less noisy background.
This indicates that adversarial training with the dual-pertubation attack which models unsuspiciousness can extract more perceptual semantics from an input image and are less dependant on the background at prediction time.    
In other words, our defense approach can extract highly robust and semantically meaningful features, which contribute to its robustness to a variety of attacks.

\section{Conclusion}
\label{conclusion}

In this paper, we proposed the dual-perturbation attack, a novel threat model that produces \emph{unsuspicious adversarial examples} by leveraging the cognitive distinction between image foreground and background.
As we have shown, our attack can defeat all state-of-the-art defenses. 
By contrast, the proposed defense approaches using our attack model can significantly improve robustness against unsuspicious adversarial examples, with relatively small performance degradation on non-adversarial data.
In addition, our defense approaches can achieve comparable to, or better robustness than the alternatives in the face of traditional attacks.

Our threat model and defense motivate several new research questions.
The first is whether there are more effective methods to identify foreground of images.
Second, can we further improve robustness to dual-perturbation attacks?
Finally, while we provide the first principled approach for quantifying suspiciousness, there may be effective alternative approaches for doing so.

%

\bibliography{dualpgd}
\bibliographystyle{iclr2021_conference}

\newpage
\appendix

\section{Detailed Descriptions of The Algorithm for Computing Dual-perturbation Examples}
\label{sec:solution}
We use the following steps to solve the optimization problem of dual-perturbation attacks:
\begin{enumerate}
\item \emph{Initialization}. 
Start with a random initial starting point $\bm{\delta}^{(0)}$. 
To do this, randomly sample a data point $\bm{\delta}_F^{(0)}$ in $\ell_p$ ball $\Delta(\epsilon_F)$ and $\bm{\delta}_B^{(0)}$ in $\Delta(\epsilon_B)$.
Then, $\bm{\delta}^{(0)}$ can be obtained by using $\bm{\delta}^{(0)} = \bm{\delta}_F^{(0)} \circ \mathcal{F}(\bm{x}) + \bm{\delta}_B^{(0)} \circ \mathcal{B}(\bm{x})$. 
This ensures that the initial perturbation is feasible in both foreground and background.

\item \emph{Split}.
At the $k$-th iteration, split the perturbation $\bm{\delta}^{(k)}$ into $\bm{\delta}_F^{(k)}$ for foreground and $\bm{\delta}_B^{(k)}$ for background:
\begin{equation}
\begin{cases}
\bm{\delta}_F^{(k)} = \bm{\delta}^{(k)} \circ \mathcal{F}(\bm{x}) \\
\bm{\delta}_B^{(k)} = \bm{\delta}^{(k)} \circ \mathcal{B}(\bm{x})
\end{cases}.
\end{equation}
Then update the foreground and background perturbations seperately using the following rules:
\begin{equation}
\begin{cases}
\bm{\delta}_F^{(k+1)}=\mathcal{P}_\epsilon(\bm{\delta}_F^{(k)}+\alpha_F \cdot g_F) \\
\bm{\delta}_B^{(k+1)}=\mathcal{P}_\epsilon(\bm{\delta}_B^{(k)}+\alpha_B \cdot g_B)
\end{cases}
\end{equation}
where $g_F$ is the update that corresponds to the \emph{normalized steepest descent} constrained in the foreground, and $g_B$ for the background.
Specifically, 
\begin{equation}
\begin{cases}
g_F = \mathcal{G}(\mathcal{F}(\bm{x}) \circ \nabla_{\bm{\delta}^{(k)}} \{ \mathcal{L}(h_{\bm{\theta}}(\bm{x}+\bm{\delta}^{(k)}), y)) + \lambda \cdot \mathcal{S} \left( \bm{x}+\bm{\delta}^{(k)} \right) \} \\
g_B = \mathcal{G}(\mathcal{B}(\bm{x}) \circ \nabla_{\bm{\delta}^{(k)}} \{ \mathcal{L}(h_{\bm{\theta}}(\bm{x}+\bm{\delta}^{(k)}), y)) + \lambda \cdot \mathcal{S} \left( \bm{x}+\bm{\delta}^{(k)} \right)\}
\end{cases}
\label{eq:gf_gb}
\end{equation}
where $\alpha_F$ is the stepsize for foreground, and $\alpha_B$ is the stepsize for background.

\item \emph{Merge}.
At the end of the $k$-th iteration, merge the perturbations obtained in the last step by using
\begin{equation}
\bm{\delta}^{(k+1)} = \bm{\delta}_F^{(k+1)} + \bm{\delta}_B^{(k+1)}.
\end{equation}
$\bm{\delta}^{(k+1)}$ is further used to derive the update for the normalized steepest descent at the next iteration.

\item Return to step 2 or terminate after either a fixed number of iterations.

\end{enumerate}

\section{Descriptions of Datasets}

\subsection{Segment-6}
The statistics of the Segment-6 dataset are displayed in Table~\ref{tab:segment-6}.

\begin{table*}[h]
\centering
\begin{tabular}{|l|l|l|}
\hline
\multirow{2}{*}{\textbf{Class}} & \multicolumn{2}{l|}{\textbf{Number of samples}} \\ \cline{2-3} 
                                & \textbf{Training}        & \textbf{Test}        \\ \hline \hline
Train                           & 3,000                    & 200                  \\ \hline
Bird                            & 3,000                    & 200                  \\ \hline
Cat                             & 3,000                    & 200                  \\ \hline
Dog                             & 3,000                    & 200                  \\ \hline
Toilet                          & 3,000                    & 200                  \\ \hline
Clock                           & 3,000                    & 200                  \\ \hline \hline
Total                           & 18,000                   & 1,200                \\ \hline
\end{tabular}
\caption{Number of samples in each class of the Segment-6 dataset.}
\label{tab:segment-6}
\end{table*}

\subsection{STL-10}
The statistics of the STL-10 dataset are displayed in Table~\ref{tab:stl-10}.

\begin{table*}[h]
\centering
\begin{tabular}{|l|l|l|}
\hline
\multirow{2}{*}{\textbf{Class}} & \multicolumn{2}{l|}{\textbf{Number of samples}} \\ \cline{2-3} 
                                & \textbf{Training}        & \textbf{Test}        \\ \hline \hline
Airplane                        & 500                     & 10                  \\ \hline
Bird                            & 500                     & 10                  \\ \hline
Car                             & 500                     & 10                  \\ \hline
Cat                             & 500                      & 10                  \\ \hline
Deer 						     & 500                      & 10                  \\ \hline		
Dog                             & 500                     & 10                  \\ \hline
Horse                           & 500                     & 10                  \\ \hline 
Monkey 							 & 500                     & 10                  \\ \hline 	
Ship							 & 500                     & 10                  \\ \hline 	
Truck                           & 500                     & 10                  \\ \hline \hline
Total                           & 5,000                   & 100                 \\ \hline
\end{tabular}
\caption{Number of samples in each class of the STL-10 dataset.}
\label{tab:stl-10}
\end{table*}

\subsection{ImageNet-10}
The labels and number of images per class in the ImageNet-10 dataset are listed in Table~\ref{tab:imagenet-10}.

\begin{table*}[h]
\centering
\begin{tabular}{|l|l|l|}
\hline
\multirow{2}{*}{\textbf{Class}} & \multicolumn{2}{l|}{\textbf{Number of samples}} \\ \cline{2-3} 
                                & \textbf{Training}        & \textbf{Test}        \\ \hline \hline
Airplane                        & 500                     & 10                  \\ \hline
Car                            & 500                     & 10                  \\ \hline
Cat                             & 500                     & 10                  \\ \hline
Dog                             & 500                      & 10                  \\ \hline
Truck                             & 500                     & 10                  \\ \hline
Elephant                           & 500                     & 10                  \\ \hline 
Zebra                           & 500                     & 10                  \\ \hline
Bus                           & 500                   & 10                 \\ \hline
Bear                           & 500                   & 10                 \\ \hline
Bicycle                           & 500                   & 10               \\ \hline \hline
Total                           & 5,000                   & 100                 \\ \hline
\end{tabular}
\caption{Number of samples in each class of the ImageNet-10 dataset.}
\label{tab:imagenet-10}
\end{table*}

\section{Implementations}
We implemented all the attack model, as well as the defense approaches in PyTorch\footnote{Available at \url{https://pytorch.org/}.}, an open-source library for neural network learning.
We used the ResNet34 model~\citep{he2016deep} and standard transfer learning, as the datasets employed in our experiments do not have a sufficient amount of data to achieve high accuracy.
Specifically, we initialized the network with the model pre-trained on ImageNet, reset the final fully connected layer, and added a \emph{normalization layer} in front of the ResNet34 model, which performs a channel-wise transformation of an input by subtracting $(0.485, 0.456, 0.406)$ (the mean of ImageNet) and then being divided by $(0.229, 0.224, 0.225)$ (the standard deviation of ImageNet);~\footnote{To fit the Segment-6 dataset which contains much smaller images compared to ImageNet, we also reset the first convolutional layer of the pre-trained ResNet34 model by reducing the kernel size from $7 \times 7$ to $3 \times 3$, stride from 2 to 1, and pad from 3 to 1.}
then, we train the neural networks as usual.

Unless otherwise specified, we used 60 epochs with training batch size 128 for Segment-6.
For STL-10 and ImageNet-10. we trained the classifiers for 20 epochs by using a batch size of 64.
We used Adam Optimizer~\citep{kingma2014adam} with initial learning rate of $10^{-4}$ for \emph{Clean}, and $10^{-3}$ for \emph{AT-PGD} and \emph{AT-Dual}, respectively.
We dropped the learning rate by 0.1 every 20 epochs on Segment-6, and similarly at the 8th and 15th epochs on STL-10 and ImageNet-10.

As mentioned above, we implemented \emph{PGD} and \emph{dual-perturbation} attacks, bounded by both $\ell_\infty$ and $\ell_2$ norms, to evaluate robustness of a classification model, as well as to build robust classifiers. 
For $\ell_\infty$ attacks, when they were used for evaluation, they are performed with 20 steps; for training robust classifiers, these attacks were performed with 10 steps at each epoch of adversarial training.
Similarly, for $\ell_2$ attacks, they were performed with 100 steps for evaluation, and 50 steps for adversarial training. 
We used the semantic segmentation masks on the Segment-6 dataset and used fixation prediction to identify foreground and backround on STL-10 and ImageNet-10. 

\section{Adversarial Training Using $\ell_2$ Norm Attacks on ImageNet-10}
\begin{figure}[h]
\centering
\begin{tabular}{cc}
  \includegraphics[width=0.35\textwidth]{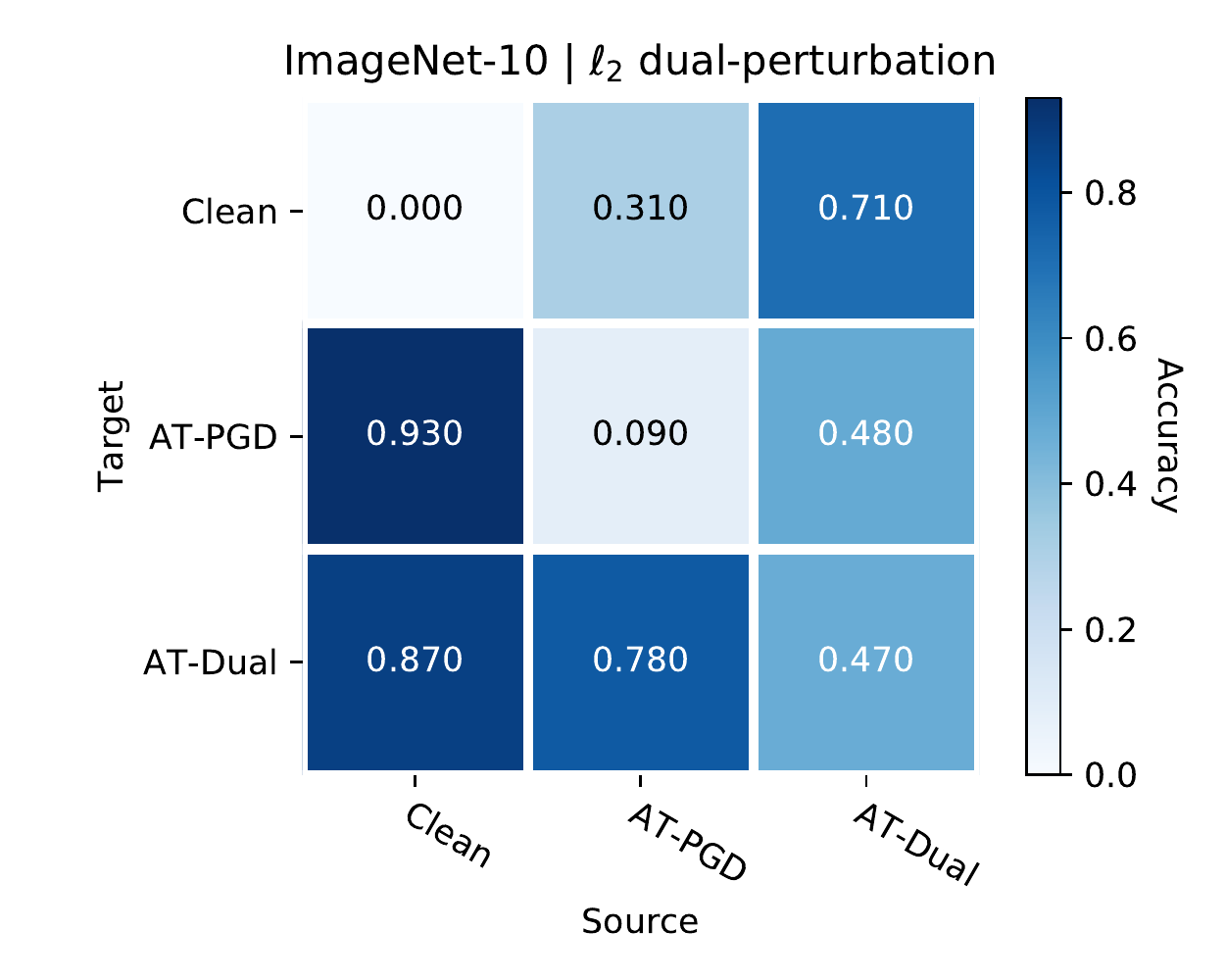} &
  \includegraphics[width=0.35\textwidth]{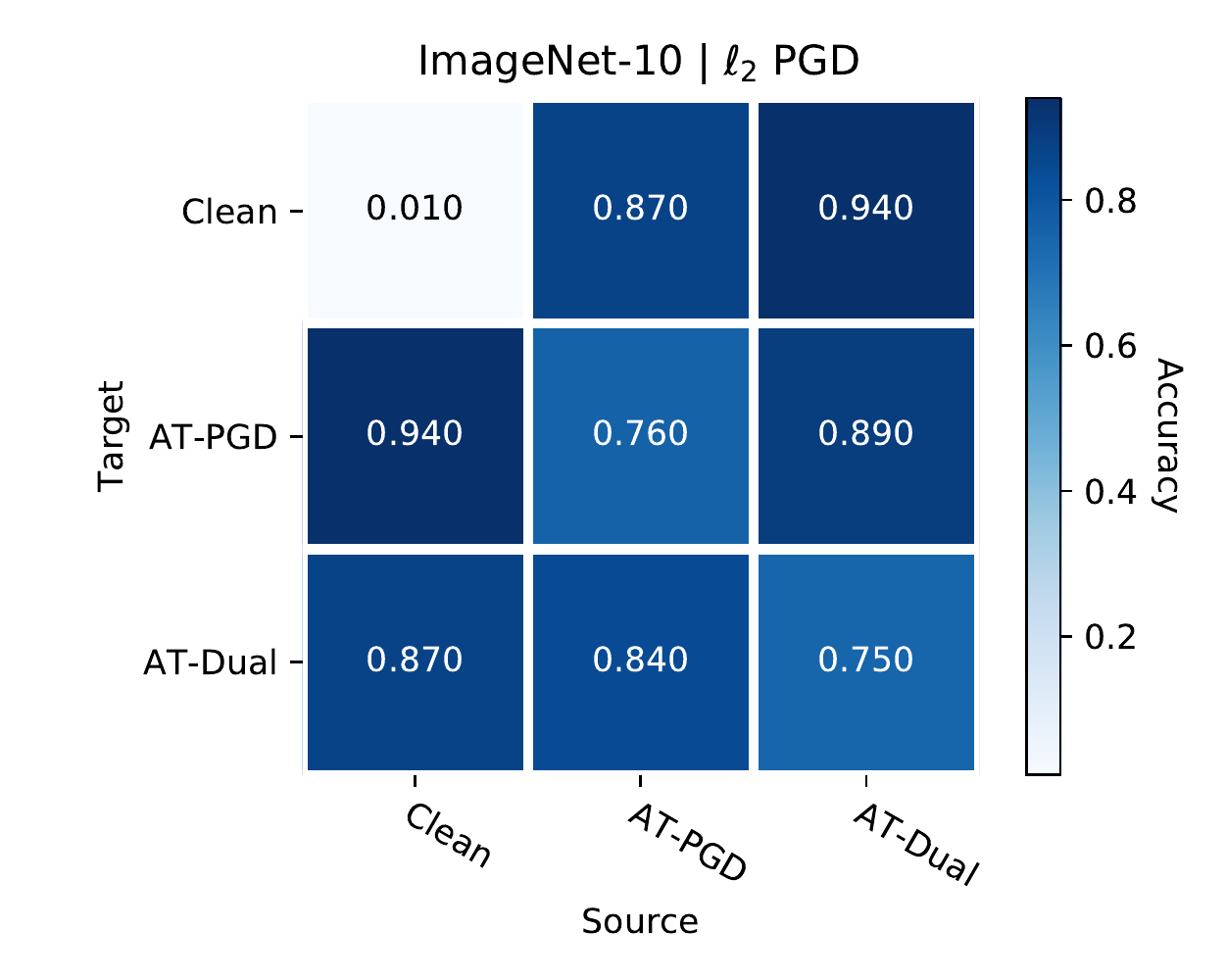}\\
\end{tabular}
\caption{Robustness against adversarial examples transferred from other models on ImageNet-10.  
Left: $\ell_2$ dual-perturbation attacks performed by using $\{\epsilon_F, \epsilon_B, \lambda\}=\{2.0, 20.0, 1.0\}$ on different source models.
Right: $\ell_2$ PGD attacks with $\epsilon=2.0$ on different source models.
}
\label{F:transfer_l2_imagenet}
\end{figure}

{\bf Transferability of Adversarial Examples.}
Here, we measure the \emph{transferability} of adversarial examples among different classification models.
To do this, we first produced adversarial examples by using $\ell_2$ PGD attack or dual-perturbation attack on a source model.
Then, we used these examples to evaluate the performance of an independent target model, where a higher prediction accuracy means weaker transferability.
The results are presented in Figure~\ref{F:transfer_l2_imagenet}.
The first observation is that dual-perturbation attacks exhibit significantly better transferability than the conventional PGD attacks (transferability is up to 40\% better for dual-perturbation attacks).
Second, we can observe that when \emph{AT-Dual} is used as the target (i.e., defending by adversarial training with dual-perturbation examples), these are typically resistant to adversarial examples generated against either the clean model, or against \emph{AT-PGD}.
This observation obtains even when we use PGD to generate adversarial examples.

\newpage
\section{Adversarila Training Using $\ell_2$ Norm Attacks on STL-10}

\begin{figure}[h]
\centering
\begin{tabular}{cc}
  \includegraphics[width=0.48\textwidth]{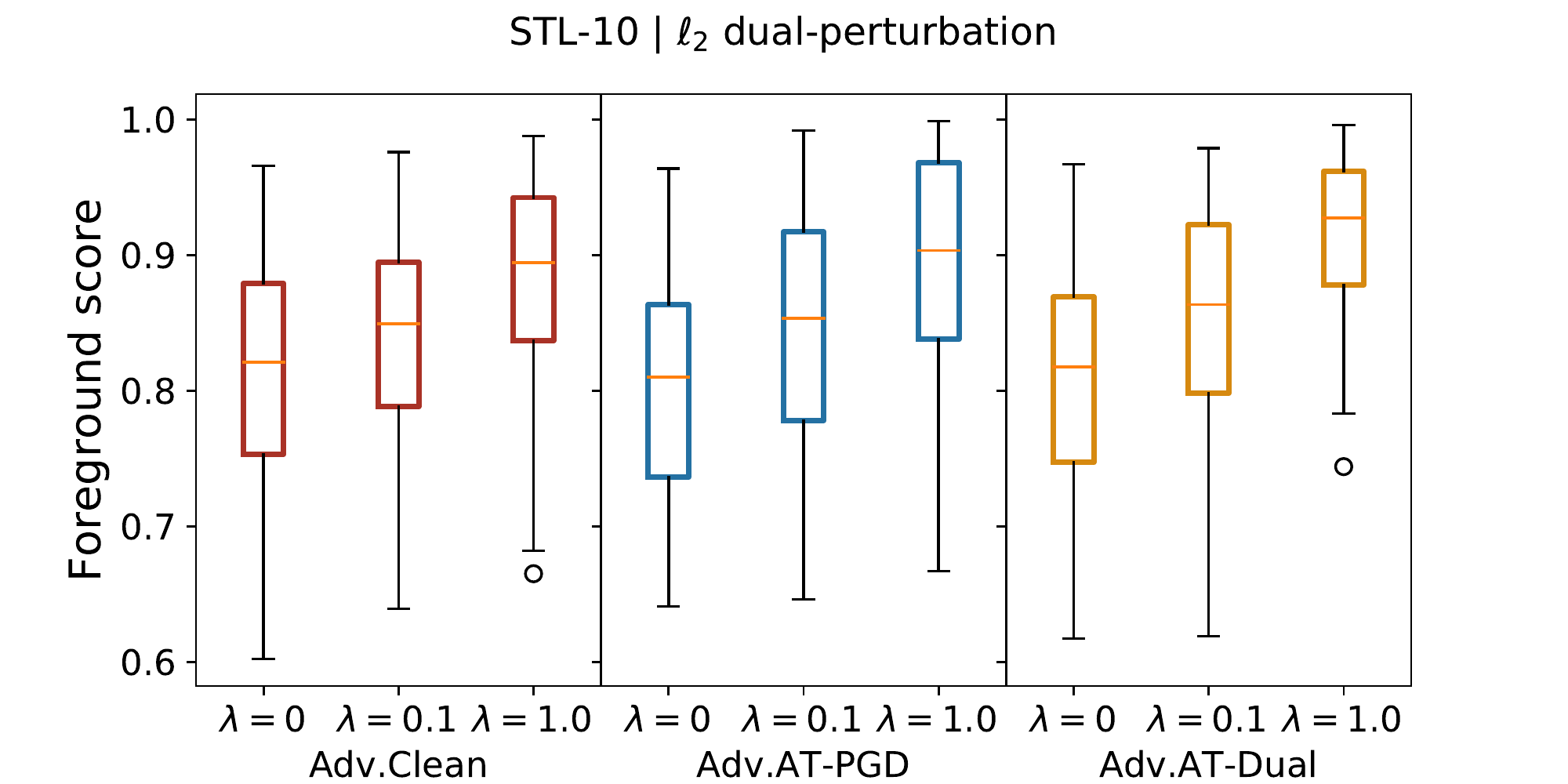} &
 \includegraphics[width=0.24\textwidth]{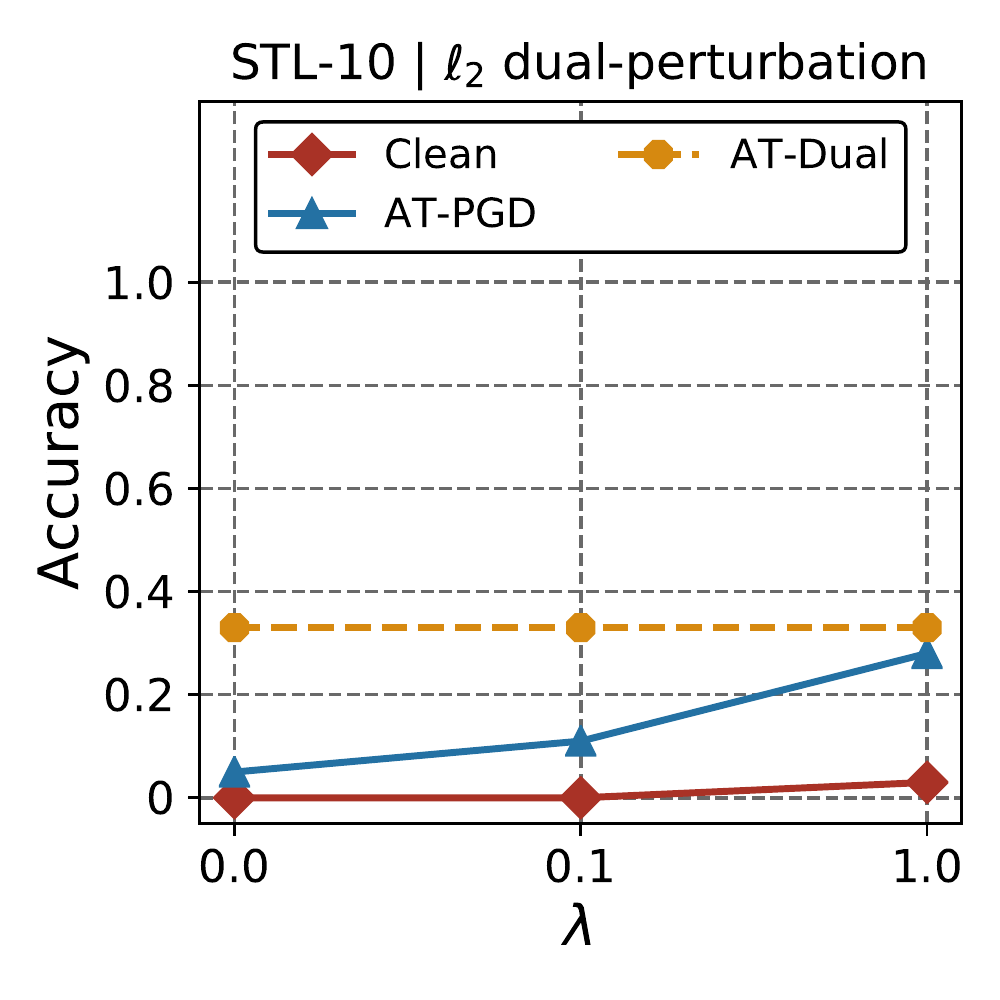}\\
\end{tabular}
	\caption{
	Saliency analysis. 
	The $\ell_2$ dual-perturbation attacks are performed by using $\{\epsilon_F, \epsilon_B\}=\{1.0, 5.0\}$, and a variety of $\lambda$ displayed in the figure. 
	Left: foreground scores of dual-perturbation examples in response to different classifiers.
	Right: accuracy of classifiers on dual-perturbation examples with salience control.  	
	}
	\label{fig:saliency_analysis_stl_l2}
\end{figure}

\begin{figure}[h]
\centering
\begin{tabular}{ccc}
  \includegraphics[width=0.26\textwidth]{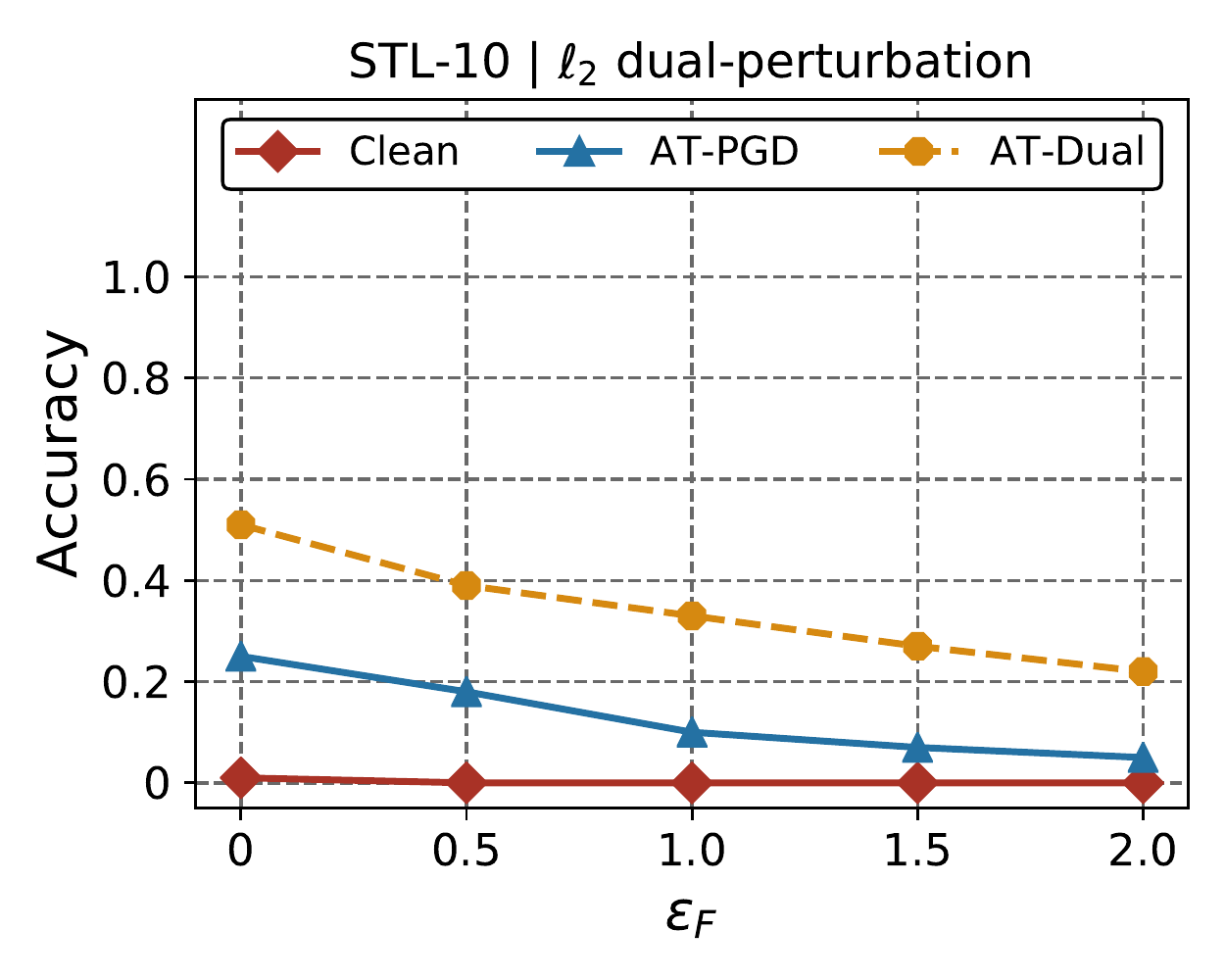} &
  \includegraphics[width=0.26\textwidth]{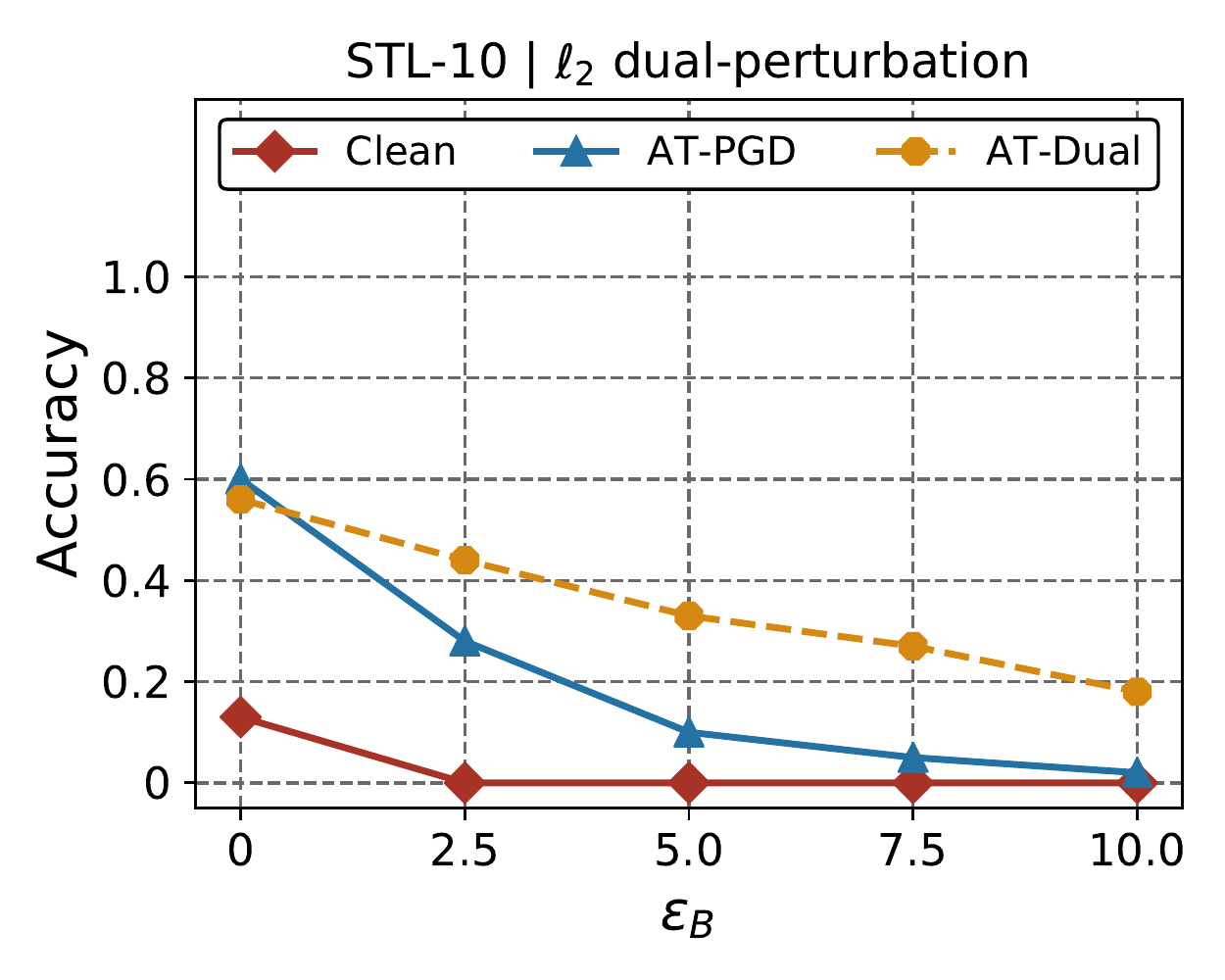} &
  \includegraphics[width=0.26\textwidth]{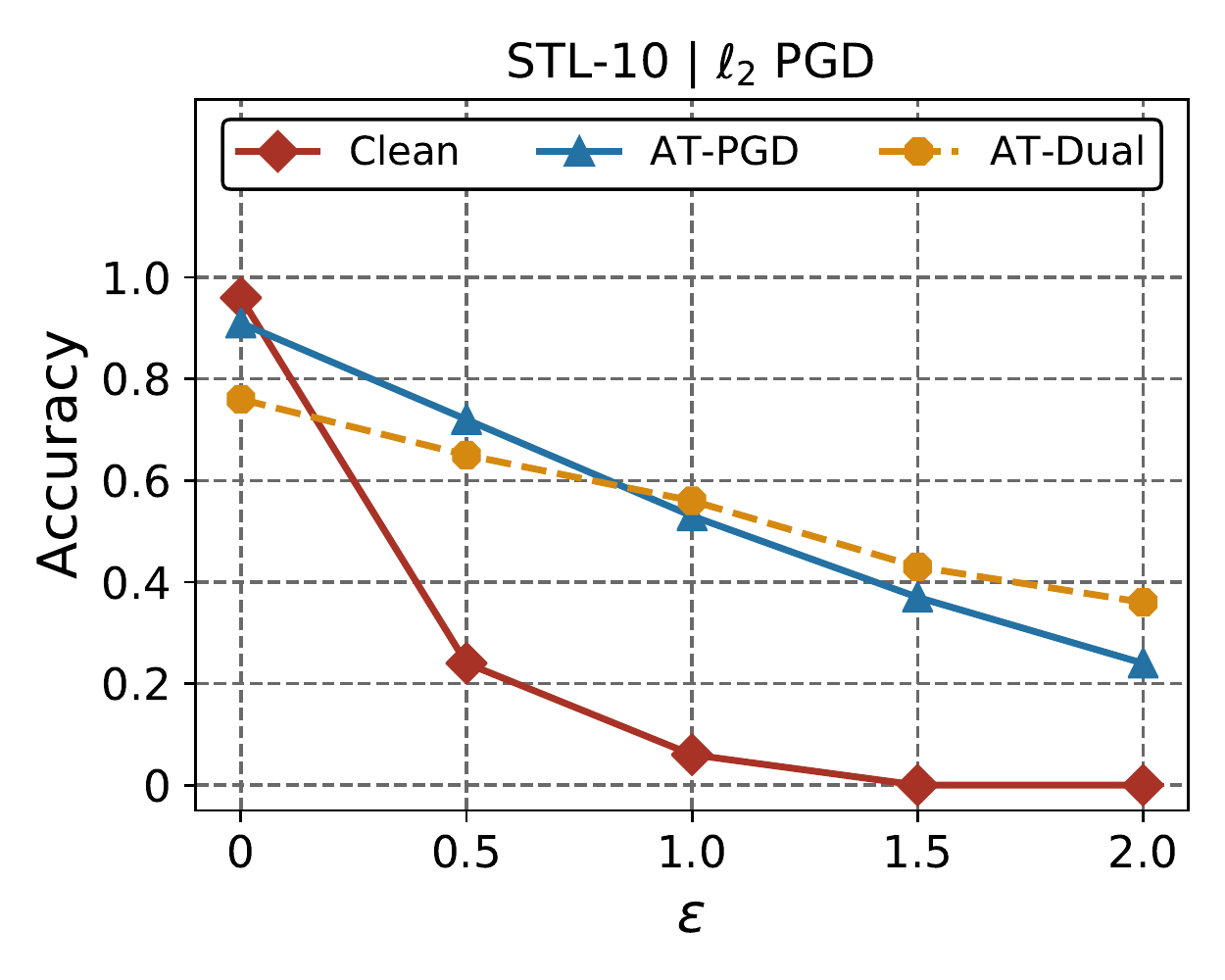}\\
\end{tabular}
\caption{
Robustness to white-box $\ell_2$ attacks on STL-10.
Left: $\ell_2$ dual-perturbation attacks with different foreground distortions. $\epsilon_B$ is fixed to be 5.0 and $\lambda=0.1$.
Middle: $\ell_2$ dual-perturbation attacks with different background distortions. $\epsilon_F$ is fixed to be 1.0 and $\lambda=0.1$.
Right: $\ell_2$ PGD attacks. 
}
\label{fig:white_stl_l2}
\end{figure}

\begin{figure}[h]
\centering
\begin{tabular}{cc}
  \includegraphics[width=0.35\textwidth]{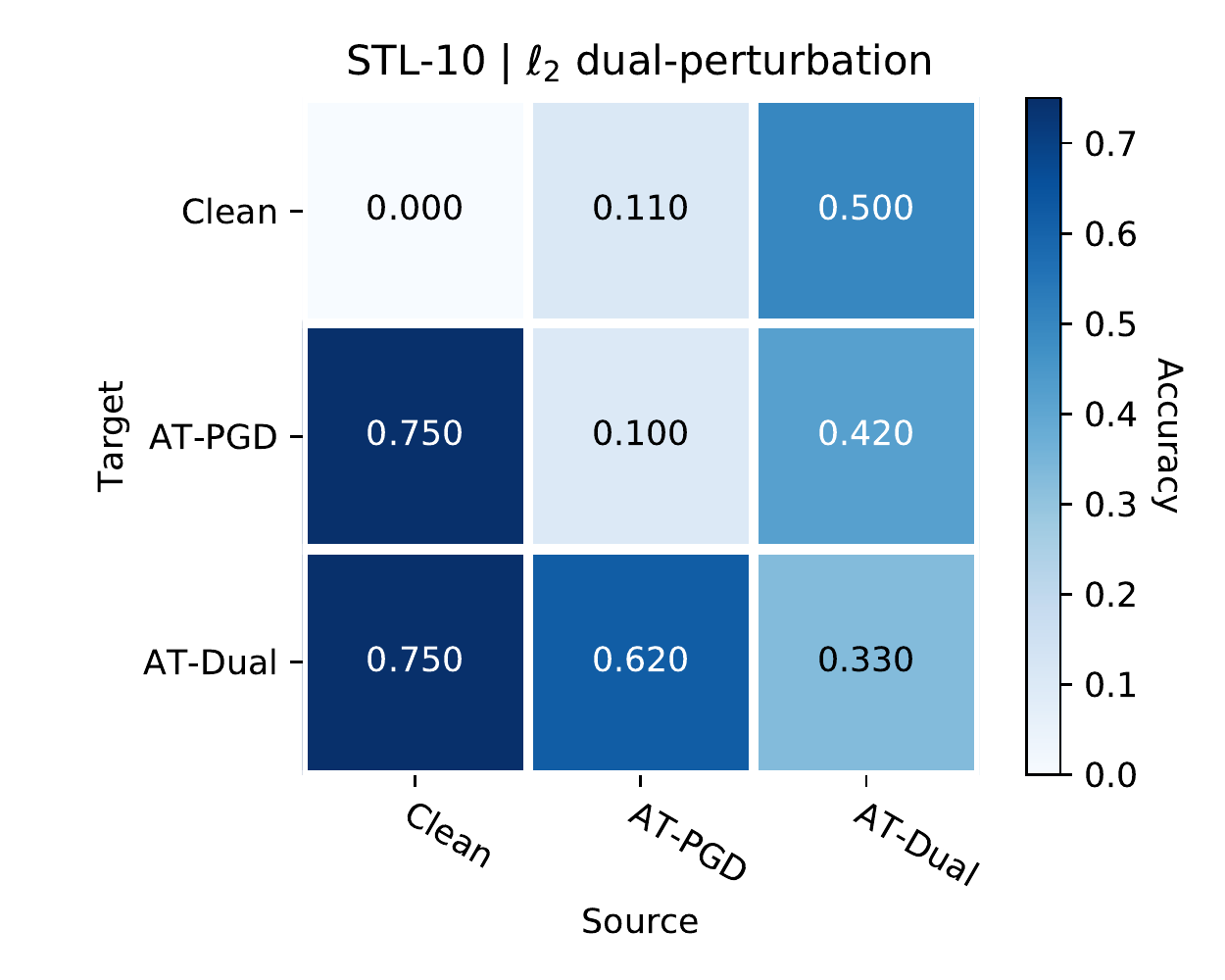} &
  \includegraphics[width=0.35\textwidth]{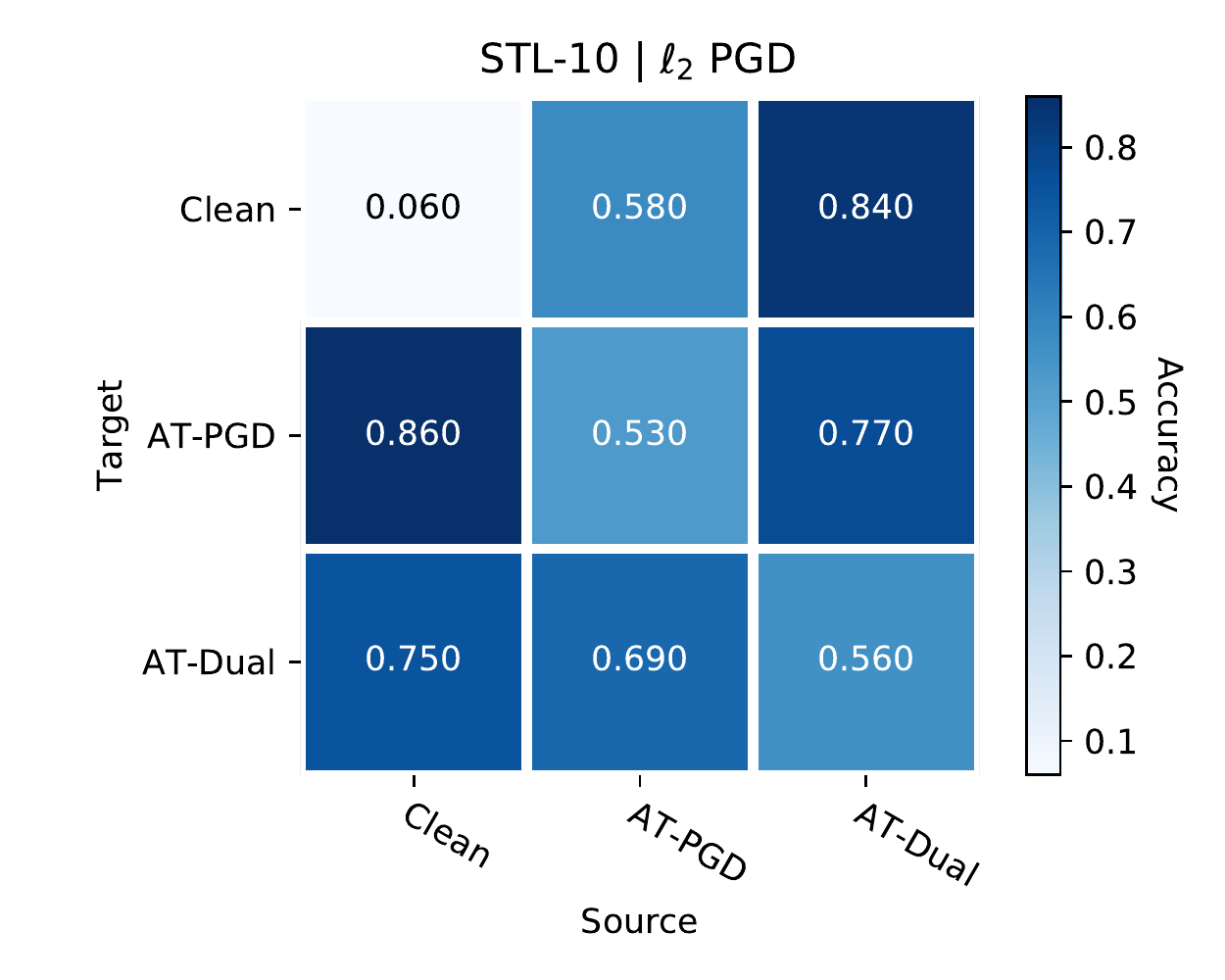}\\
\end{tabular}
\caption{
Robustness against adversarial examples transferred from other models on STL-10.  
Left: $\ell_2$ dual-perturbation attacks performed by using $\{\epsilon_F, \epsilon_B, \lambda\}=\{1.0, 5.0, 0.1\}$ on different source models.
Right: $\ell_2$ PGD attacks with $\epsilon=1.0$ on different source models.
}
\label{fig:black_stl_l2}
\end{figure}

\begin{figure}[h]
\centering
\begin{tabular}{cccc}
  \includegraphics[width=0.22\textwidth]{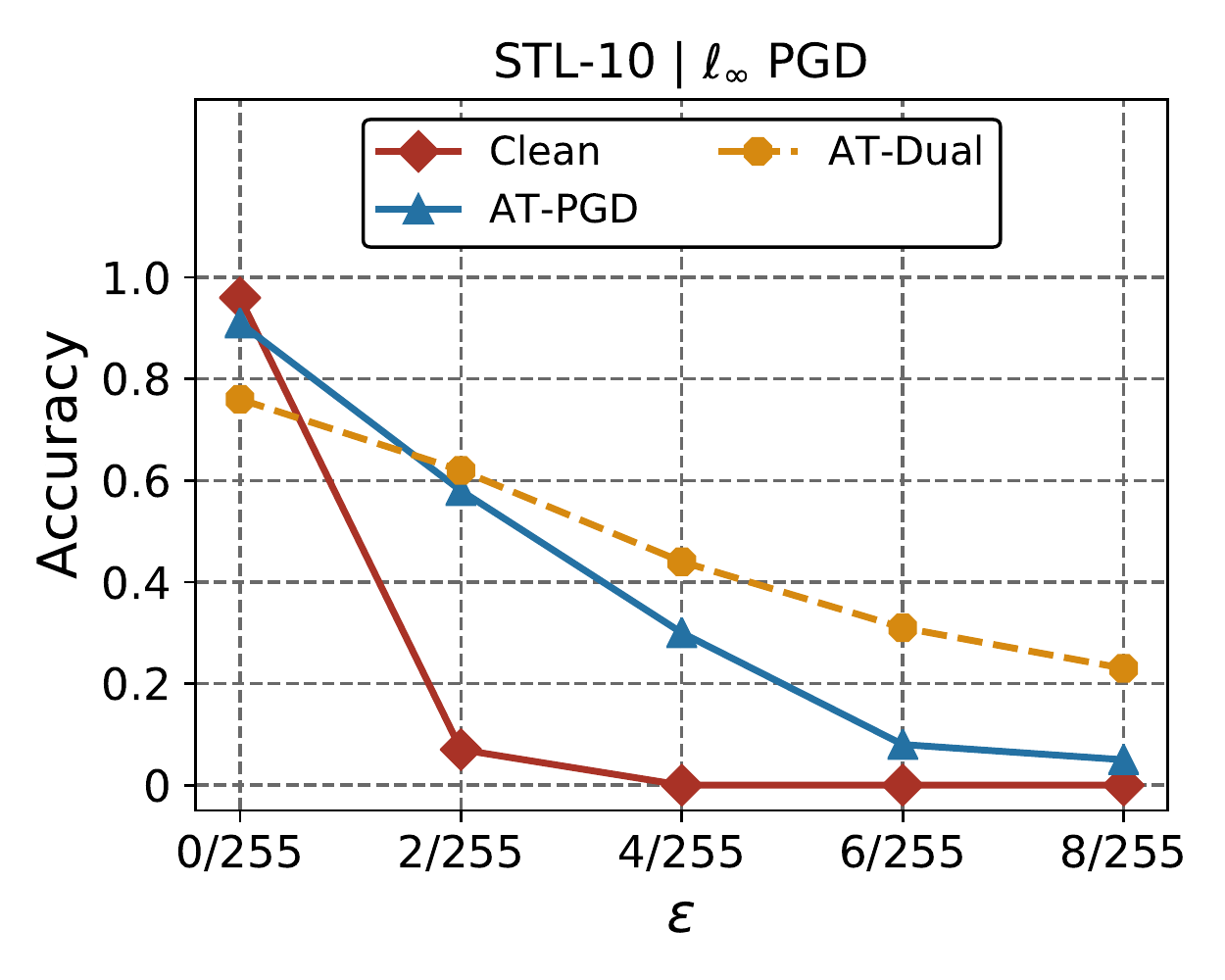} &
 \includegraphics[width=0.22\textwidth]{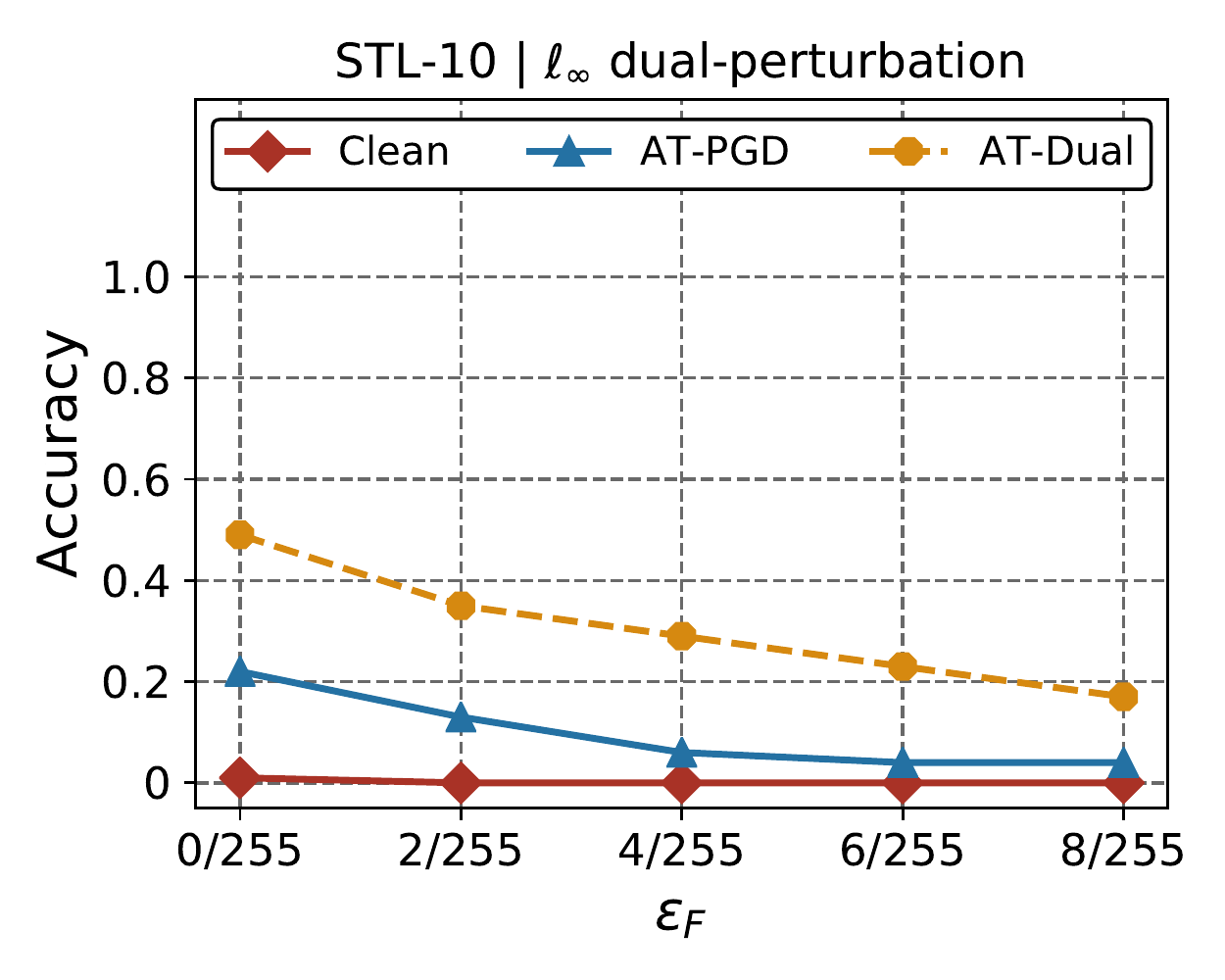} &
  \includegraphics[width=0.22\textwidth]{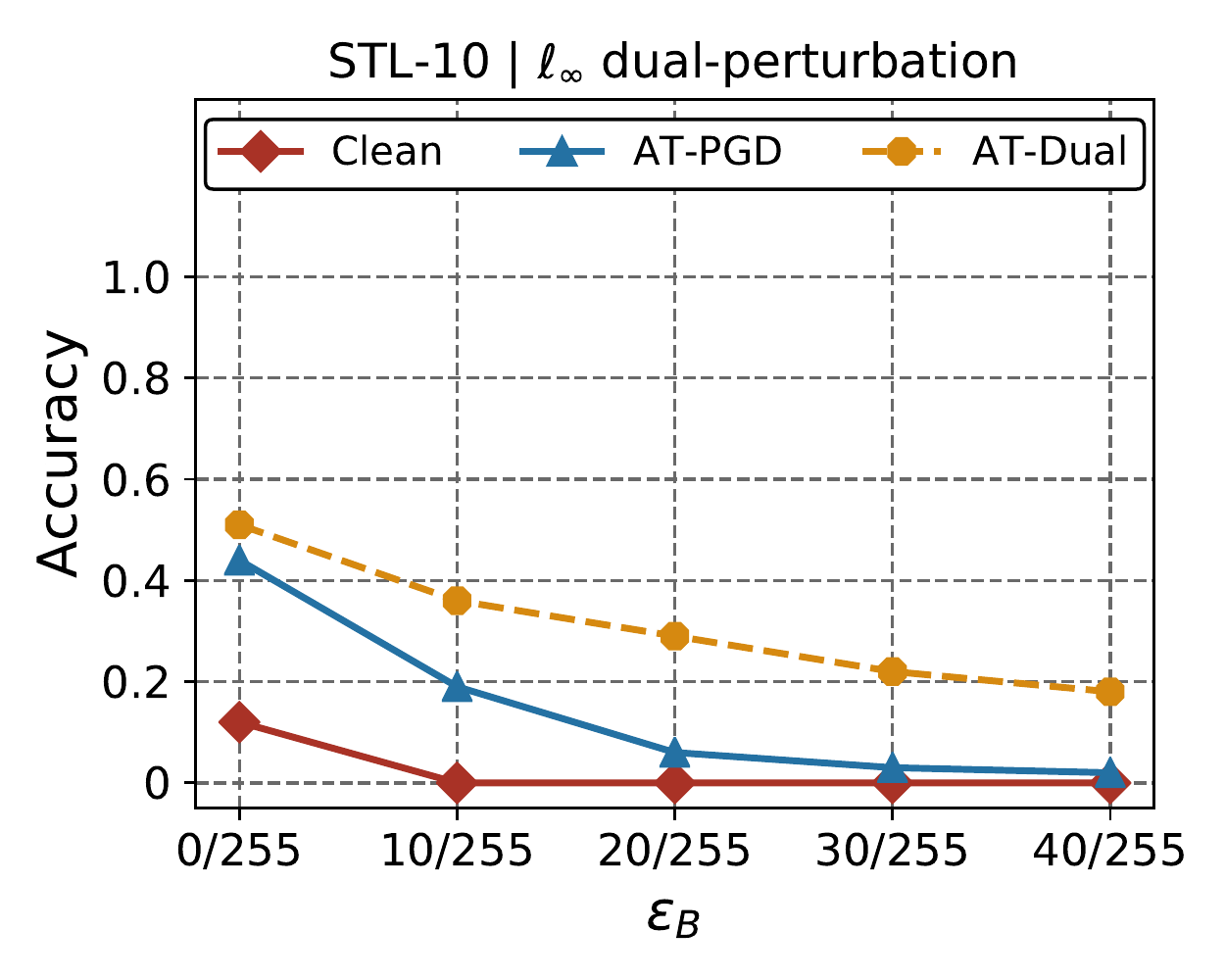} &
 \includegraphics[width=0.22\textwidth]{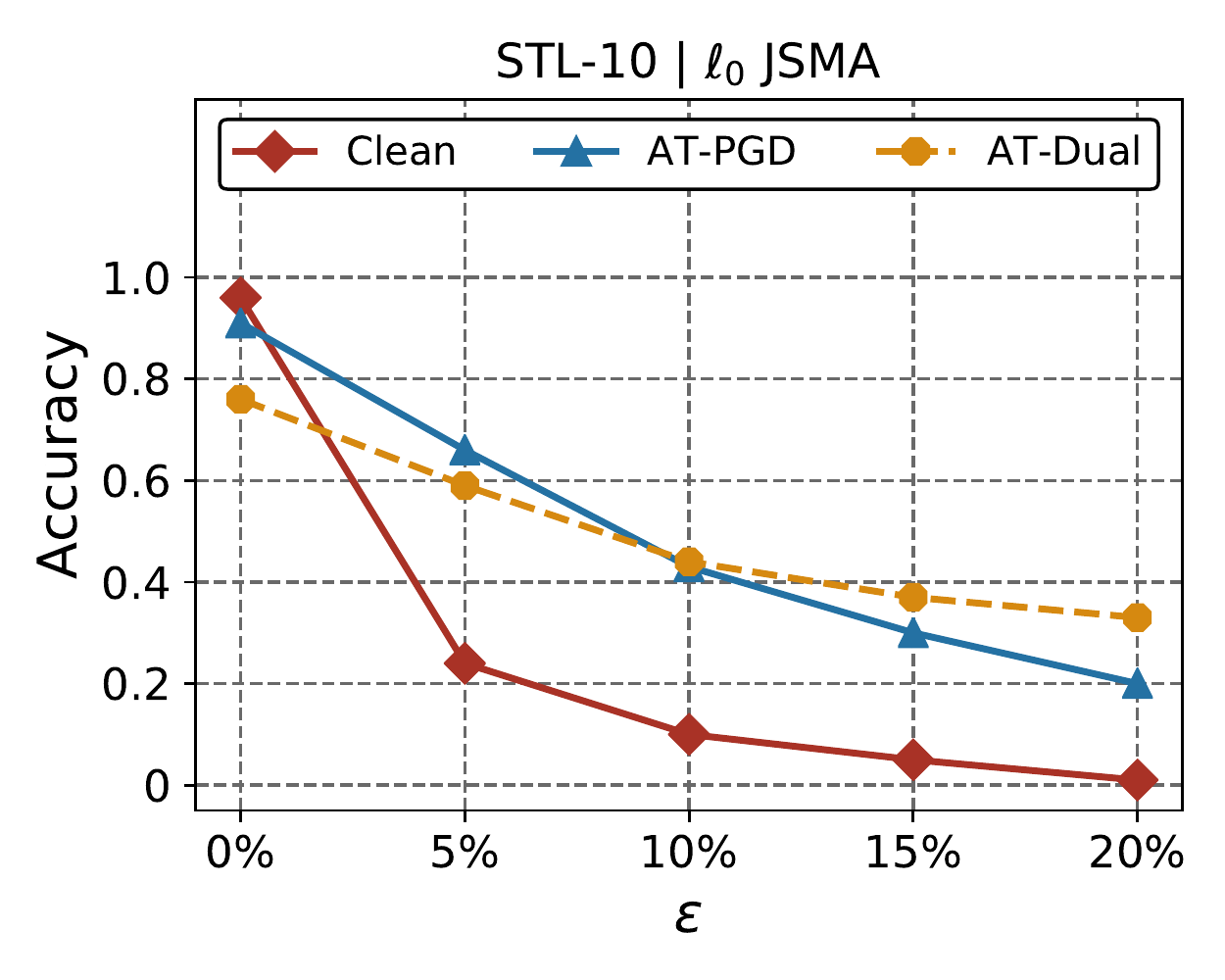} \\
\end{tabular}
\caption{
Robustness to additional white-box attacks on STL-10. 
Left: 20 steps of $\ell_\infty$ PGD attacks. 
Middle left: 20 steps of $\ell_\infty$ dual-perturbation attacks with different foreground distortions. $\epsilon_B$ is fixed to be 20/255 and $\lambda=0.1$.
Middle right: 20 steps of $\ell_\infty$ dual-perturbation attacks with different background distortions. $\epsilon_F$ is fixed to be 4/255 and $\lambda=0.1$.
Right: $\ell_0$ JSMA attacks.
}
\label{fig:general_stl_l2}
\end{figure}

Here, we present experimental results of the robustness of classifiers that use adversarial training with $\ell_2$ norm attacks on STL-10.
Specifically, we trained AT-PGD using $\ell_2$ PGD attack with $\epsilon=1.0$, and AT-Dual by using $\ell_2$ dual-perturbation attack with $\{\epsilon_F, \epsilon_B, \lambda\}=\{1.0, 5.0, 0.0\}$.
The results are shown in Figure~\ref{fig:saliency_analysis_stl_l2}, ~\ref{fig:white_stl_l2}, ~\ref{fig:black_stl_l2}, and ~\ref{fig:general_stl_l2}. 

\newpage
\section{Adversarial Training Using $\ell_2$ Norm Attacks on Segment-6}

\begin{figure}[h]
\centering
\begin{tabular}{cc}
  \includegraphics[width=0.35\textwidth]{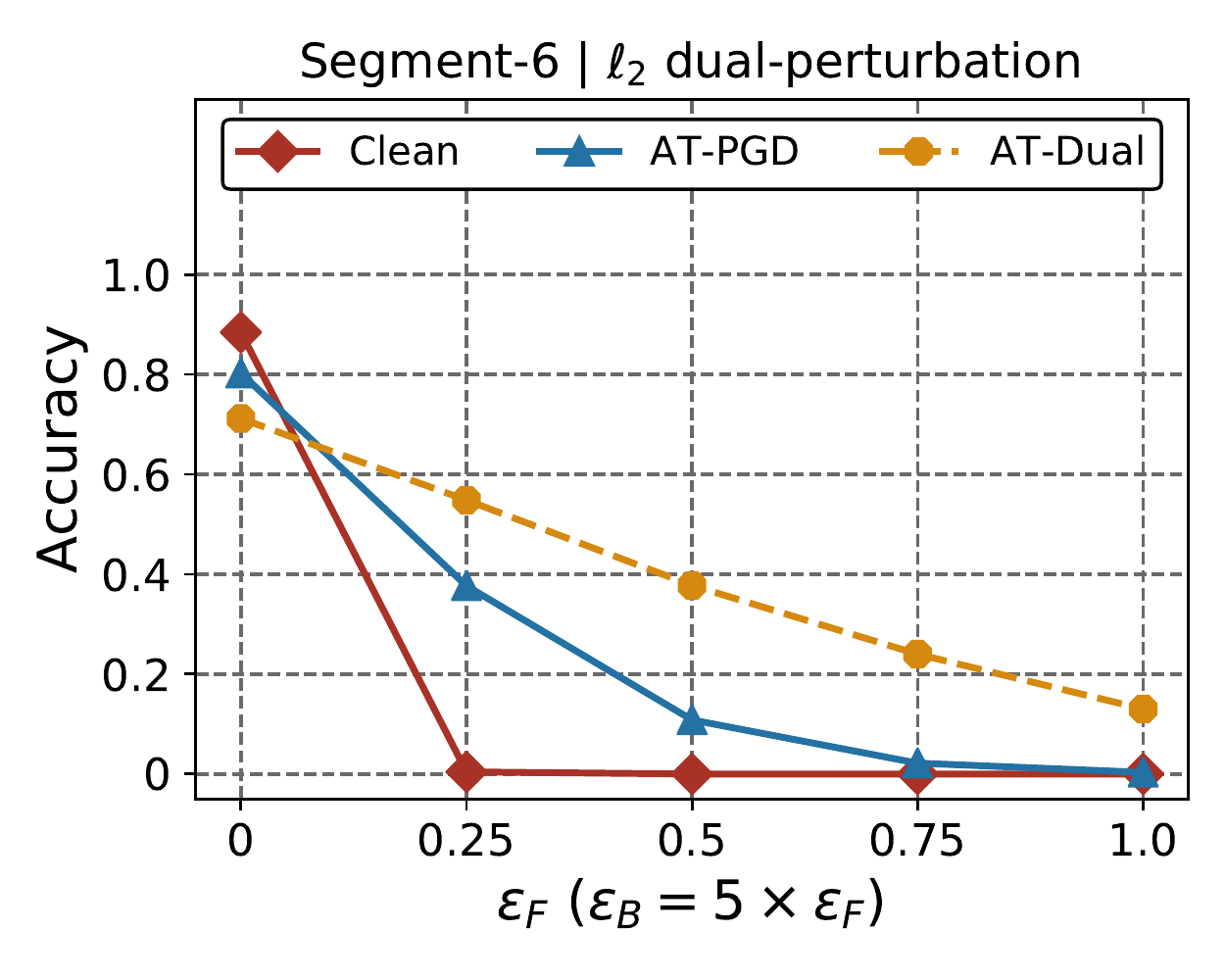} &
  \includegraphics[width=0.35\textwidth]{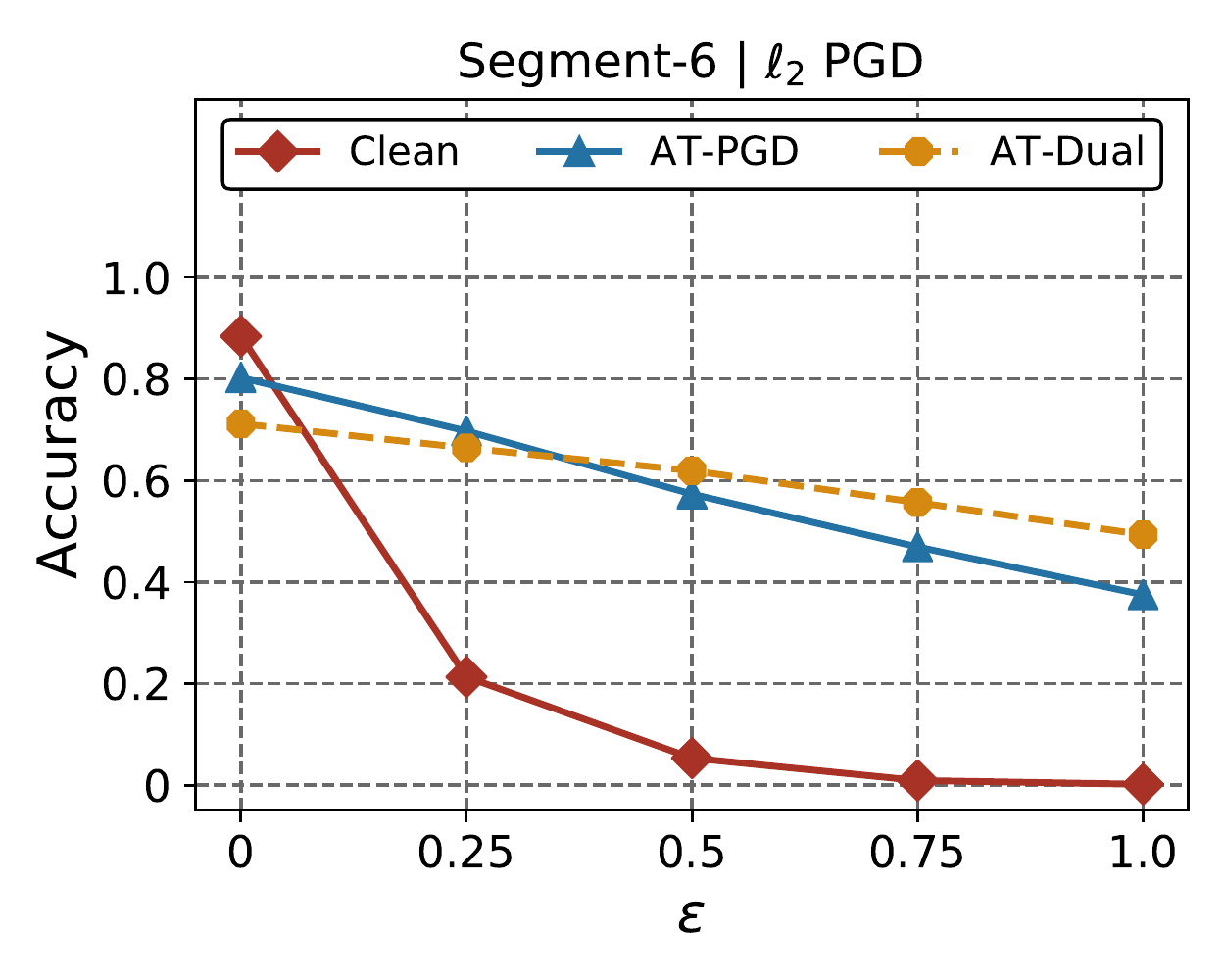}\\
\end{tabular}
\caption{
Robustness to white-box $\ell_2$ attacks on Segment-6.
Left: $\ell_2$ dual-perturbation attacks with different foreground and background distortions.
Right: $\ell_2$ PGD attacks. 
}
\label{fig:white_segment_l2}
\end{figure}

\begin{figure}[h]
\centering
\begin{tabular}{cc}
  \includegraphics[width=0.35\textwidth]{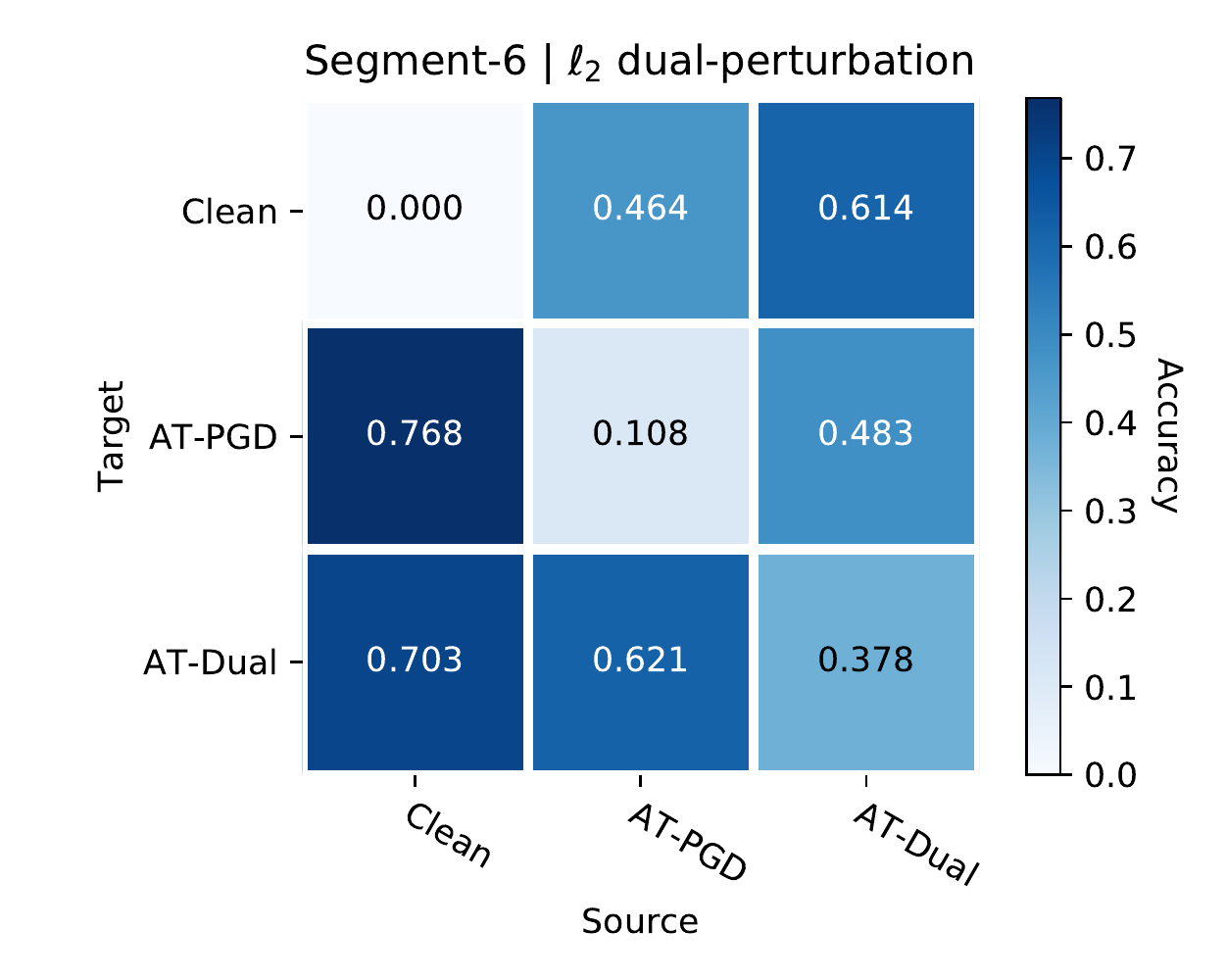} &
  \includegraphics[width=0.35\textwidth]{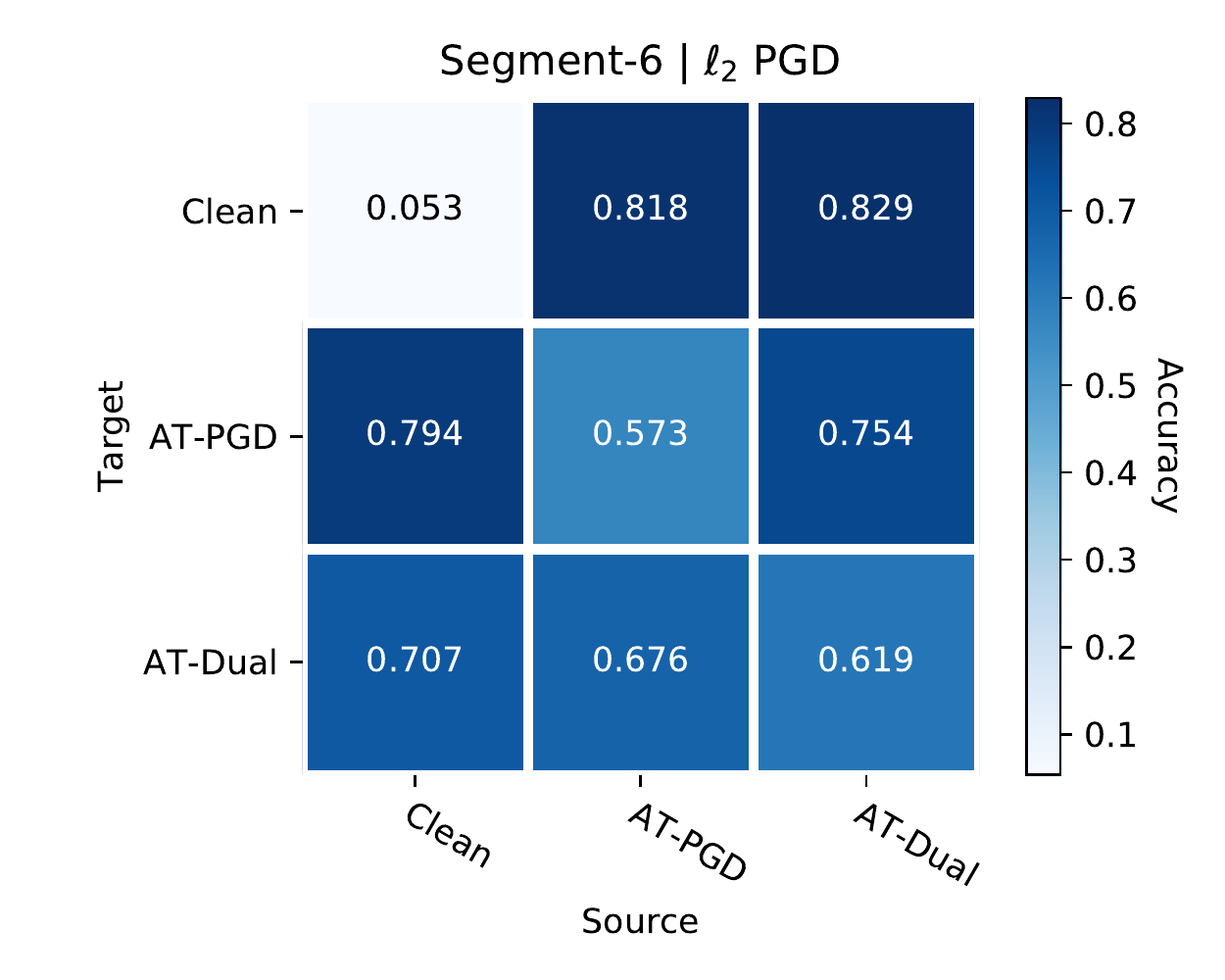}\\
\end{tabular}
\caption{
Robustness against adversarial examples transferred from other models on Segment-6.  
Left: $\ell_2$ dual-perturbation attacks performed by using $\{\epsilon_F, \epsilon_B\}=\{0.5, 2.5\}$ on different source models.
Right: $\ell_2$ PGD attacks with $\epsilon=0.5$ on different source models.
}
\label{fig:black_segment6_l2}
\end{figure}

\begin{figure}[h!]
\centering
\begin{tabular}{ccc}
  \includegraphics[width=0.26\textwidth]{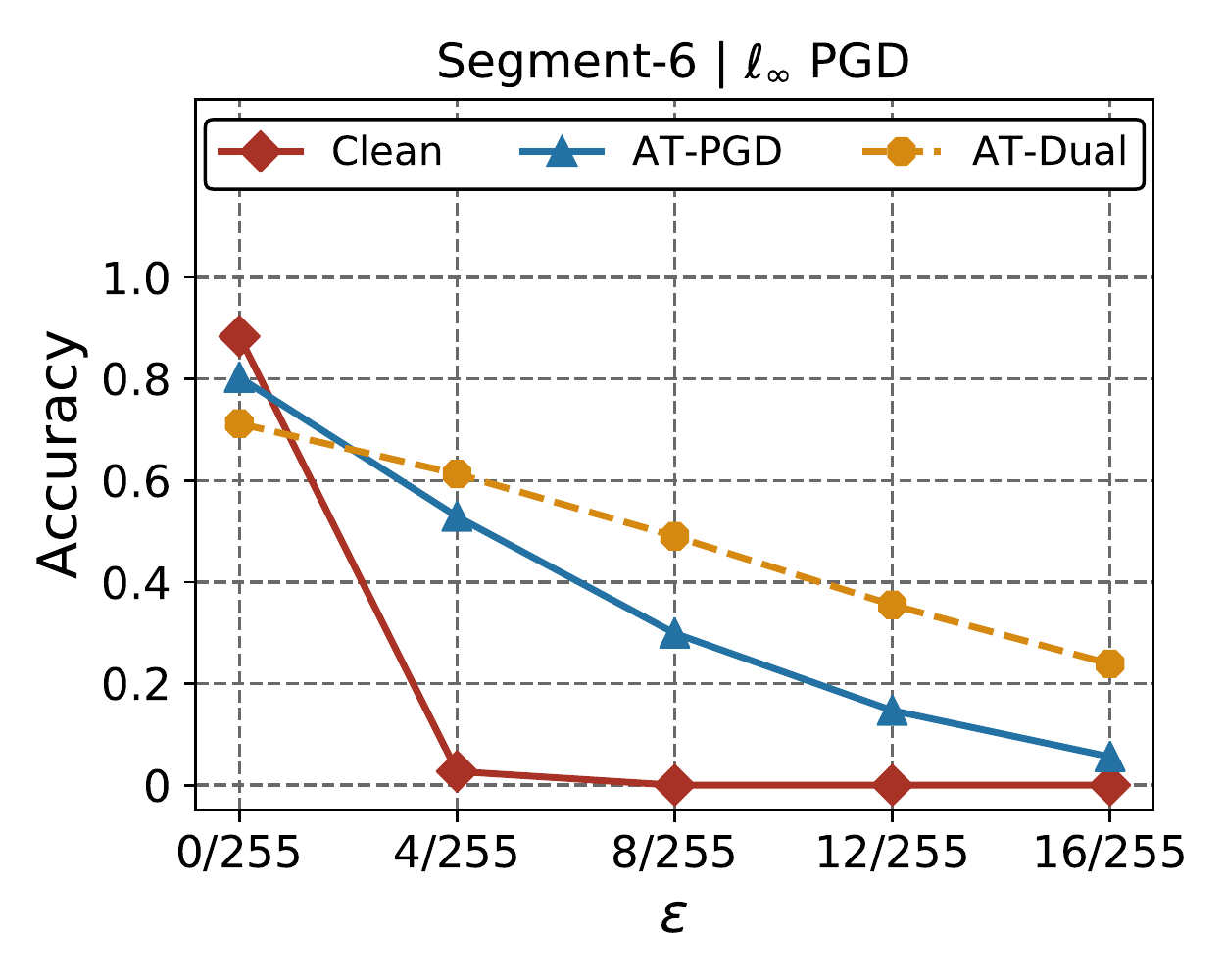} &
 \includegraphics[width=0.26\textwidth]{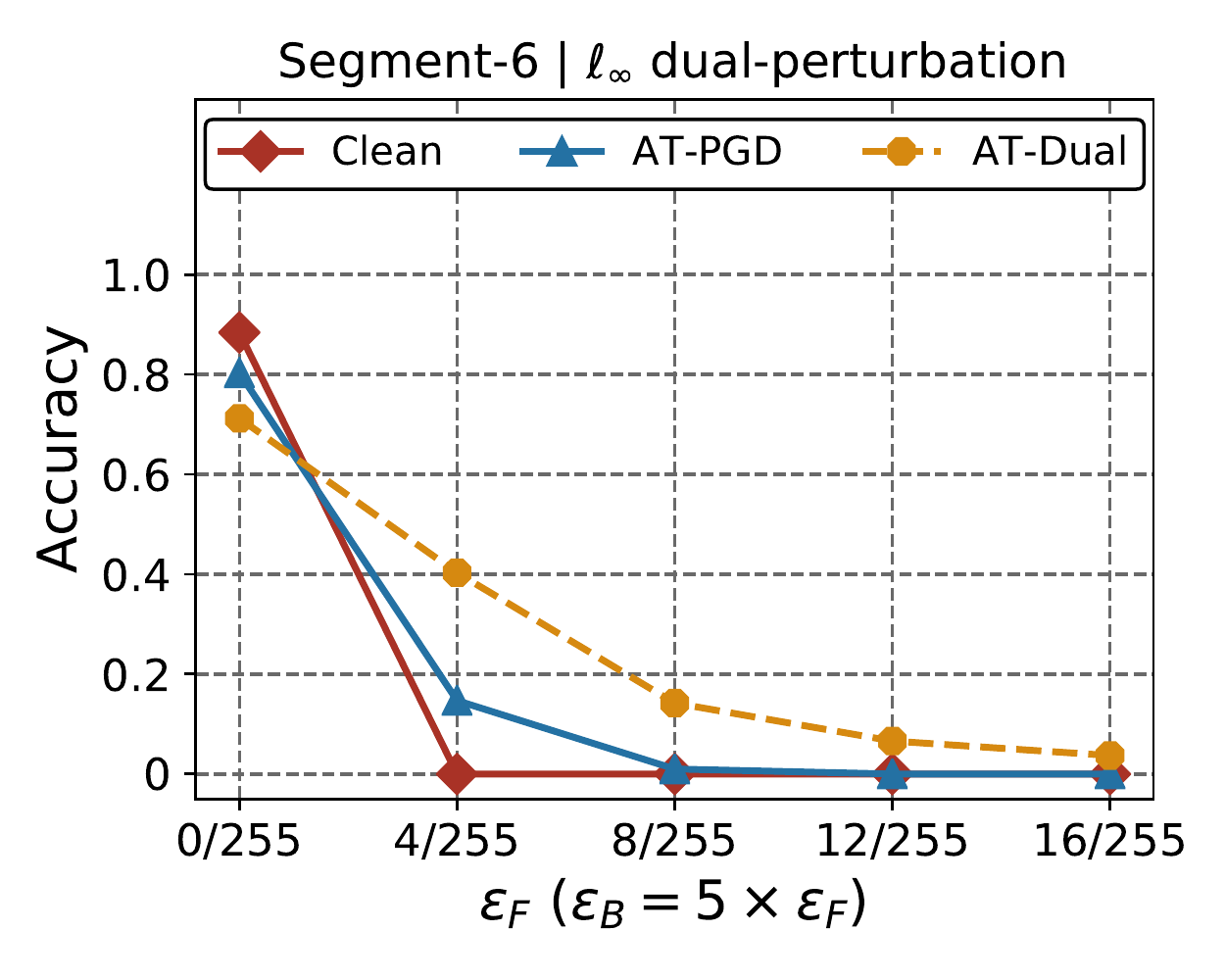} &
 \includegraphics[width=0.26\textwidth]{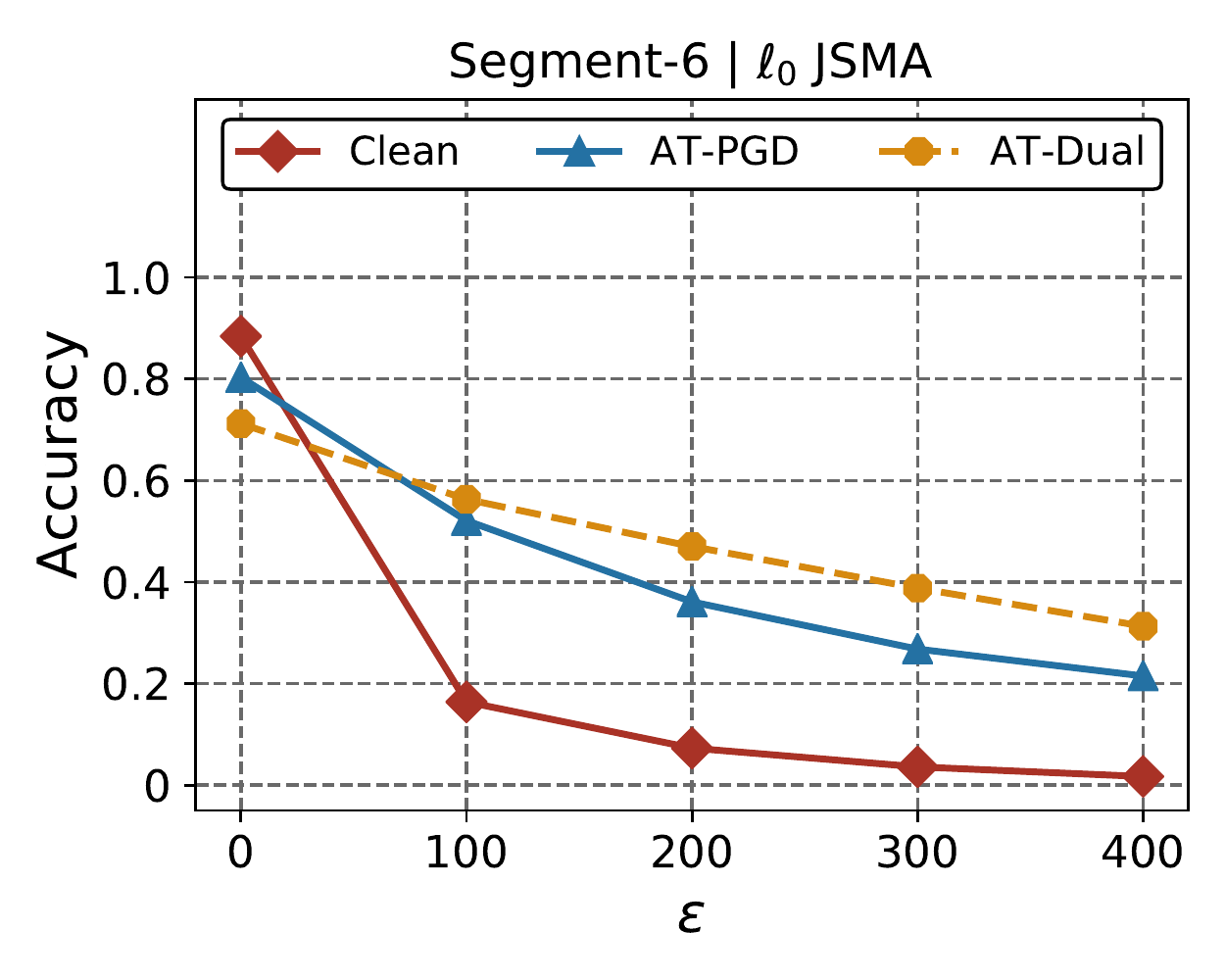} \\
\end{tabular}
\caption{
Robustness to additional white-box attacks on Segment-6. 
Left: 20 steps of $\ell_\infty$ PGD attacks. 
Middle: 20 steps of $\ell_\infty$ dual-perturbation attacks with different foreground and background distortions.
Right: $\ell_0$ JSMA attacks.
}
\label{fig:general_segment6_l2}
\end{figure}

Now, we present experimental results of the robustness of classifiers that use adversarial training with $\ell_2$ norm attacks on Segment-6.
Since DeepGaze II only work on images with more than $35 \times 35$ pixels, we are unable to use DeepGaze II to compute the \emph{foreground score (FS)} for Segment-6.
Hence, in the following experiment on this dataset, we omit the salience term in the optimization problem of Equation 3 and 4 in the main body of the paper.
Specifically, we trained AT-PGD using $\ell_2$ PGD attack with $\epsilon=0.5$, and AT-Dual by using $\ell_2$ dual-perturbation attack with $\{\epsilon_F, \epsilon_B\}=\{0.5, 2.5\}$.
The results are shown in Figure~\ref{fig:white_segment_l2}, ~\ref{fig:black_segment6_l2}, and ~\ref{fig:general_segment6_l2}. 

\newpage
\section{Adversarial Training Using $\ell_\infty$ Norm Attacks on ImageNet-10}

\begin{figure}[h]
\centering
\begin{tabular}{cc}
  \includegraphics[width=0.48\textwidth]{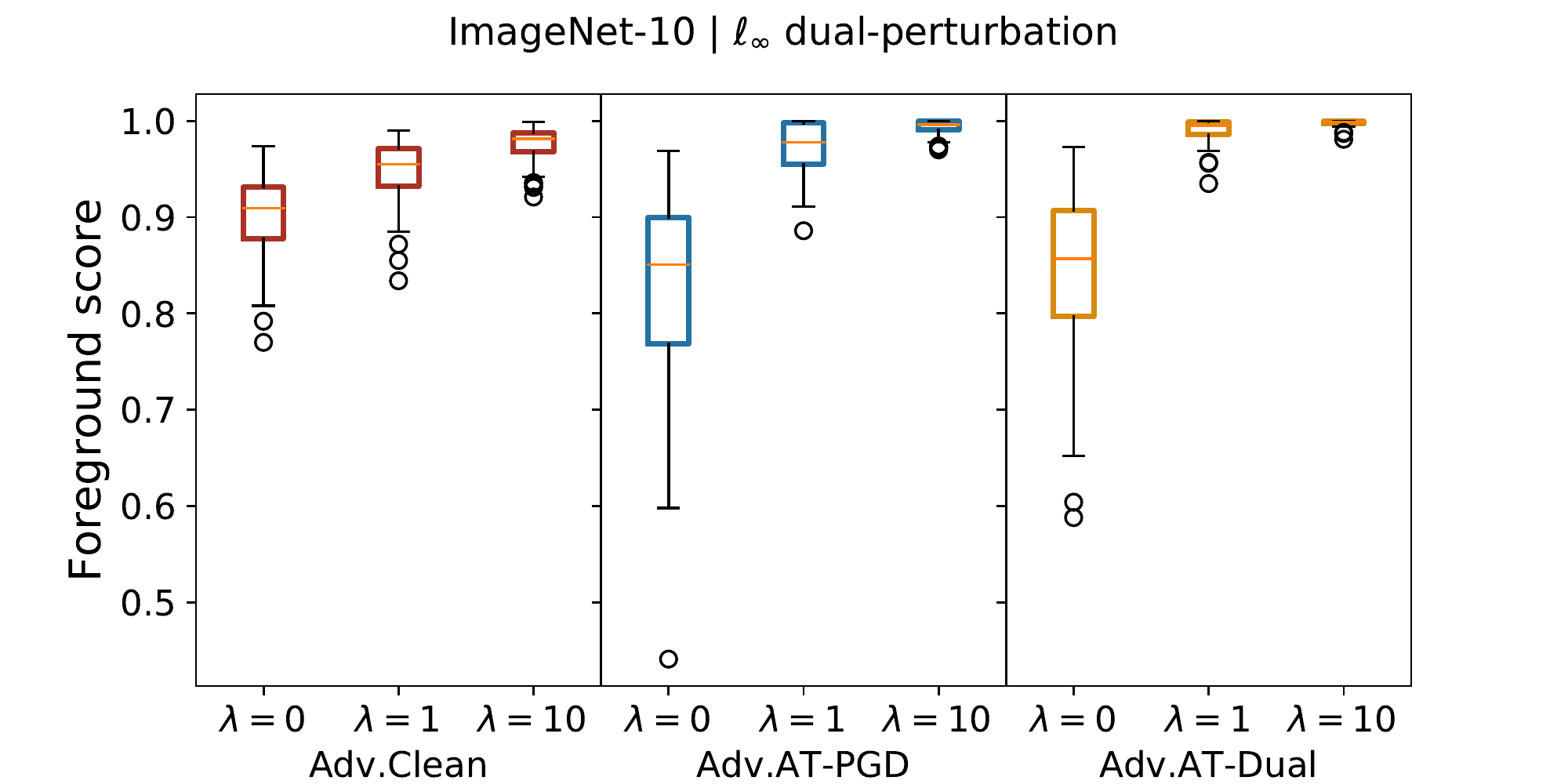} &
 \includegraphics[width=0.24\textwidth]{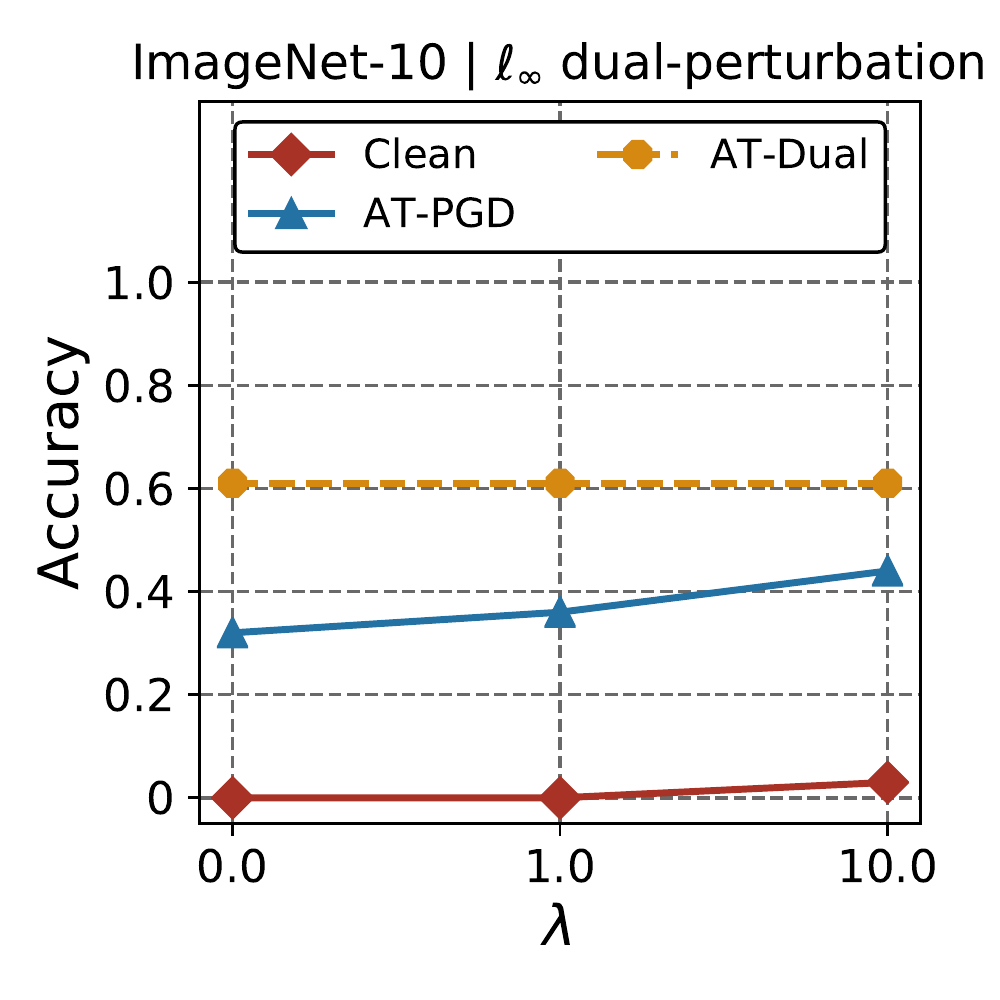}\\
\end{tabular}
	\caption{
	Saliency analysis. 
	The $\ell_\infty$ dual-perturbation attacks are performed by using $\{\epsilon_F, \epsilon_B\}=\{4/255, 20/255\}$, and a variety of $\lambda$ displayed in the figure. 
	Left: foreground scores of dual-perturbation examples in response to different classifiers.
	Right: accuracy of classifiers on dual-perturbation examples with salience control.  	
	}
	\label{fig:saliency_analysis_imagenet_linf}
\end{figure}

\begin{figure}[h]
\centering
\begin{tabular}{ccc}
  \includegraphics[width=0.26\textwidth]{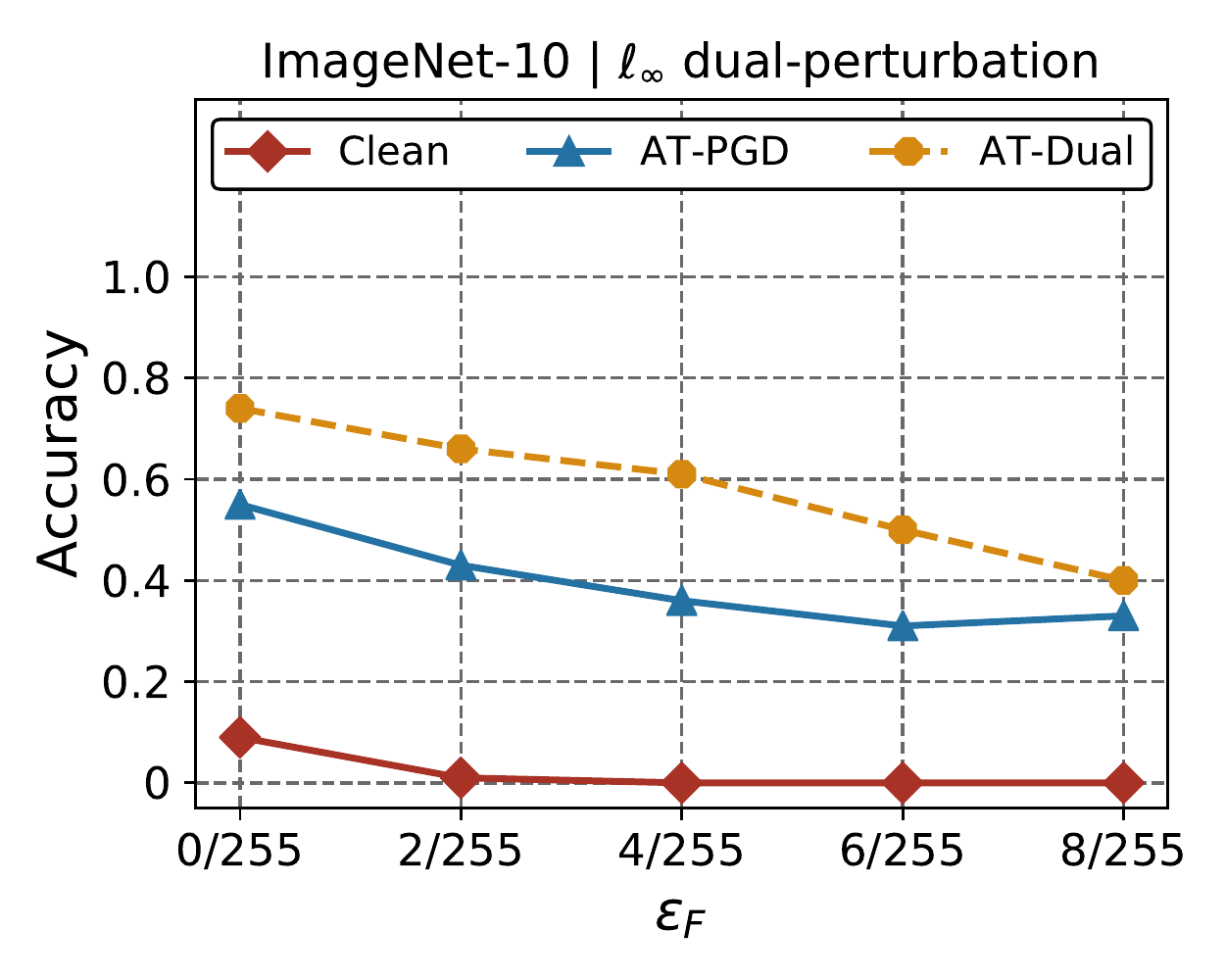} &
  \includegraphics[width=0.26\textwidth]{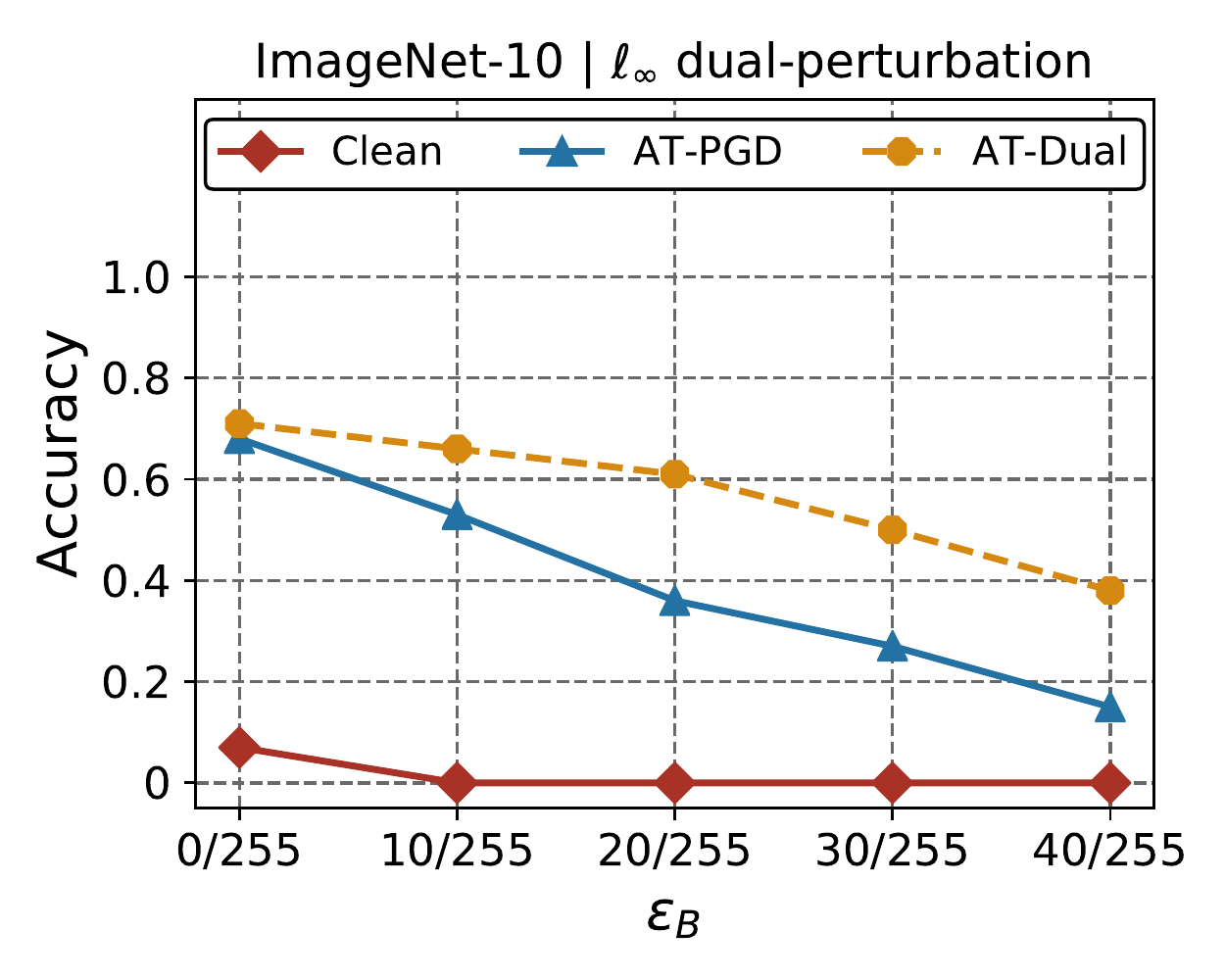} &
  \includegraphics[width=0.26\textwidth]{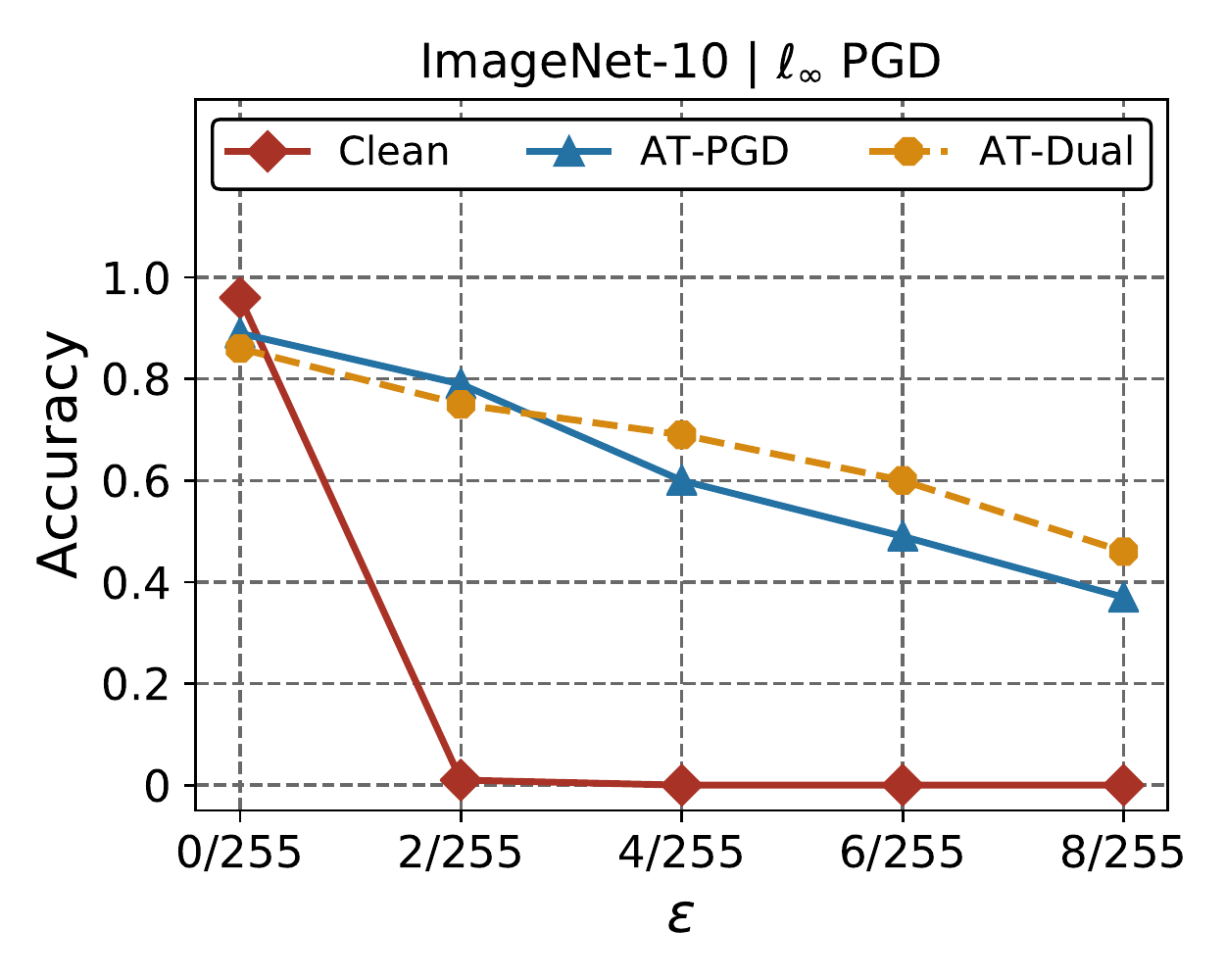}\\
\end{tabular}
\caption{
Robustness to white-box $\ell_\infty$ attacks on ImageNet-10.
Left: $\ell_\infty$ dual-perturbation attacks with different foreground distortions. $\epsilon_B$ is fixed to be 20/255 and $\lambda=1.0$.
Middle: $\ell_\infty$ dual-perturbation attacks with different background distortions. $\epsilon_F$ is fixed to be 4/255 and $\lambda=1.0$.
Right: $\ell_\infty$ PGD attacks. 
}
\label{fig:white_imagenet_linf}
\end{figure}

\begin{figure}[h]
\centering
\begin{tabular}{cc}
  \includegraphics[width=0.35\textwidth]{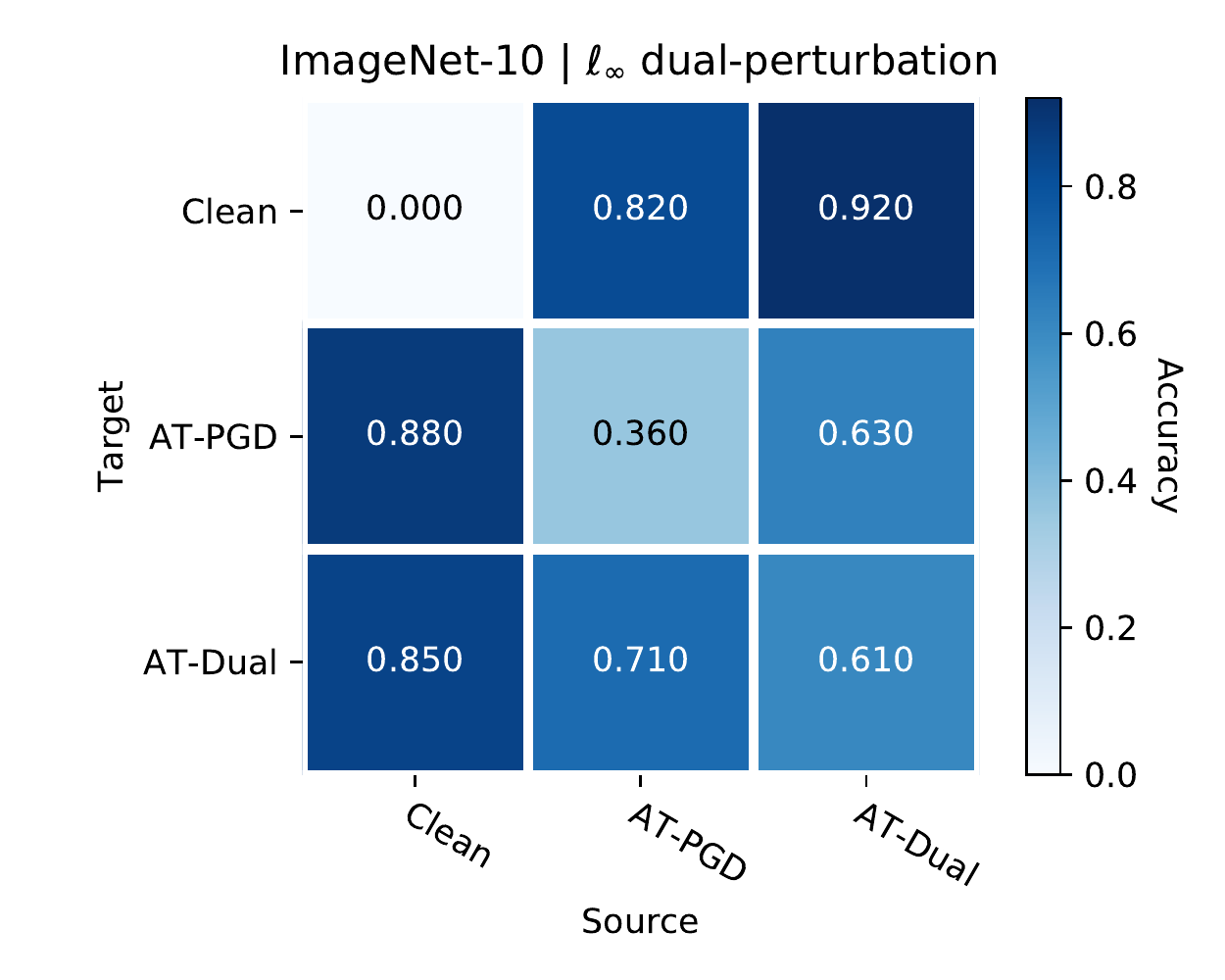} &
  \includegraphics[width=0.35\textwidth]{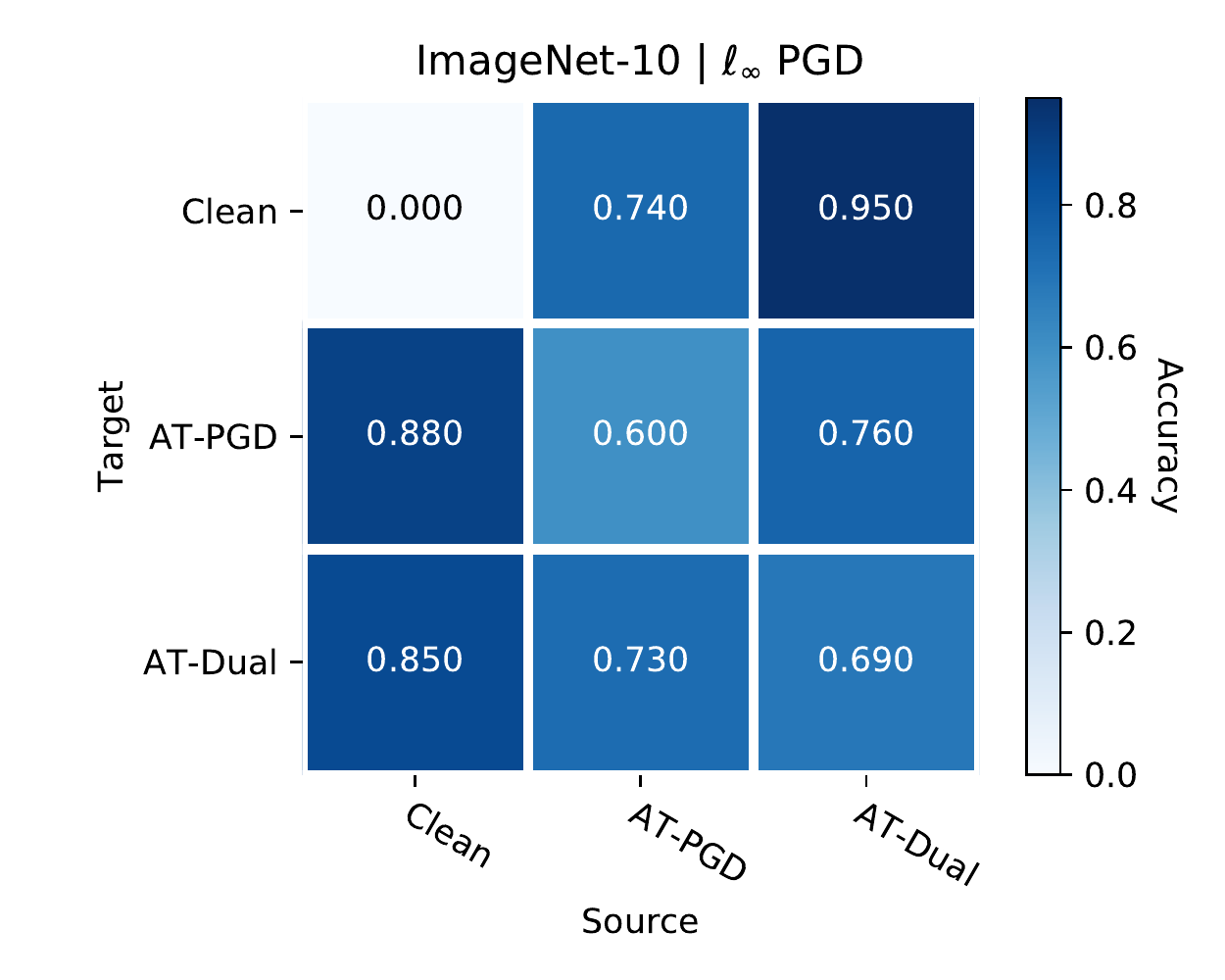}\\
\end{tabular}
\caption{
Robustness against adversarial examples transferred from other models on ImageNet-10.  
Left: $\ell_\infty$ dual-perturbation attacks performed by using $\{\epsilon_F, \epsilon_B, \lambda\}=\{4/255, 20/255, 1.0\}$ on different source models.
Right: $\ell_\infty$ PGD attacks with $\epsilon=4/255$ on different source models.
}
\label{fig:black_imagenet_linf}
\end{figure}

\begin{figure}[h!]
\centering
\begin{tabular}{cccc}
  \includegraphics[width=0.22\textwidth]{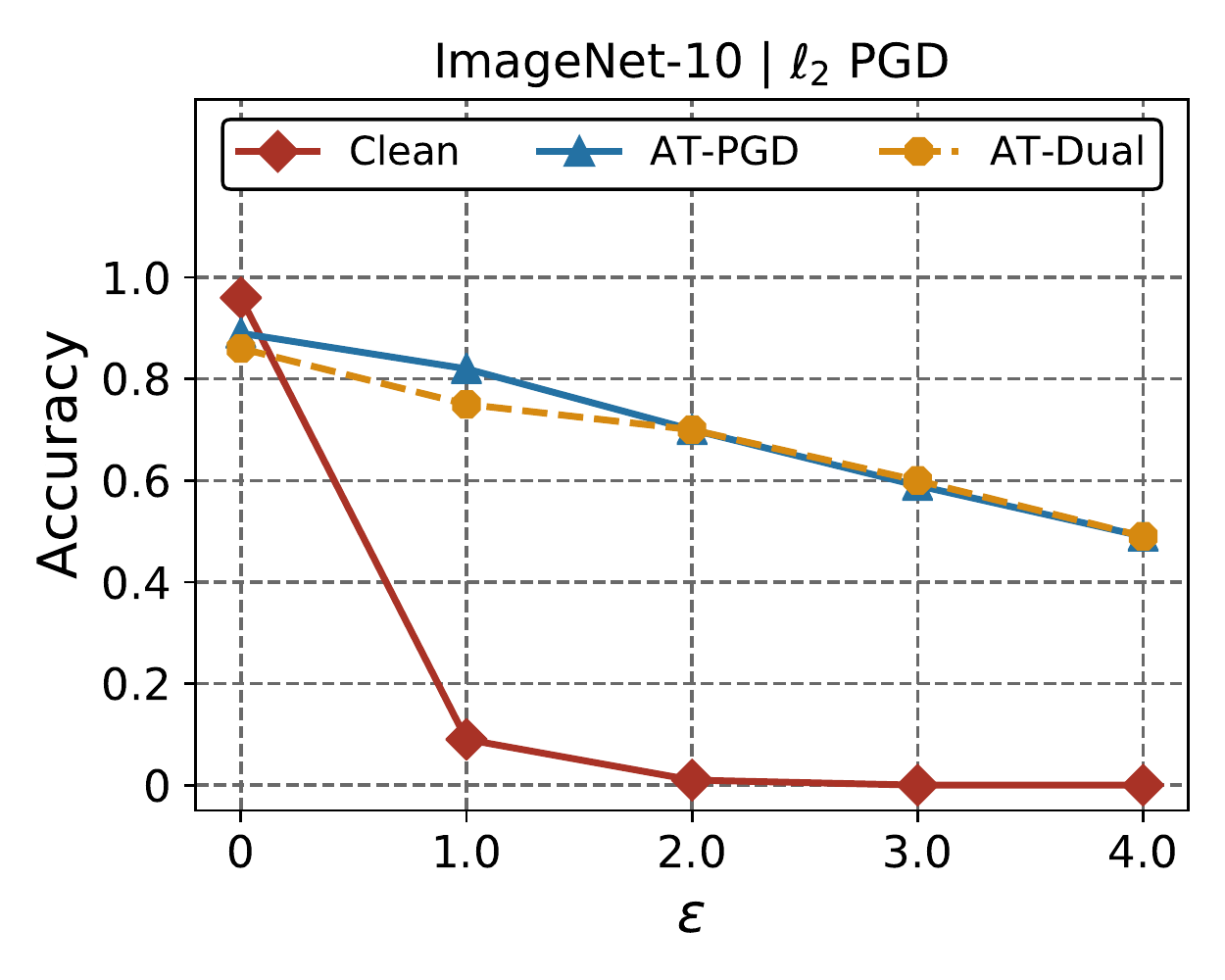} &
 \includegraphics[width=0.22\textwidth]{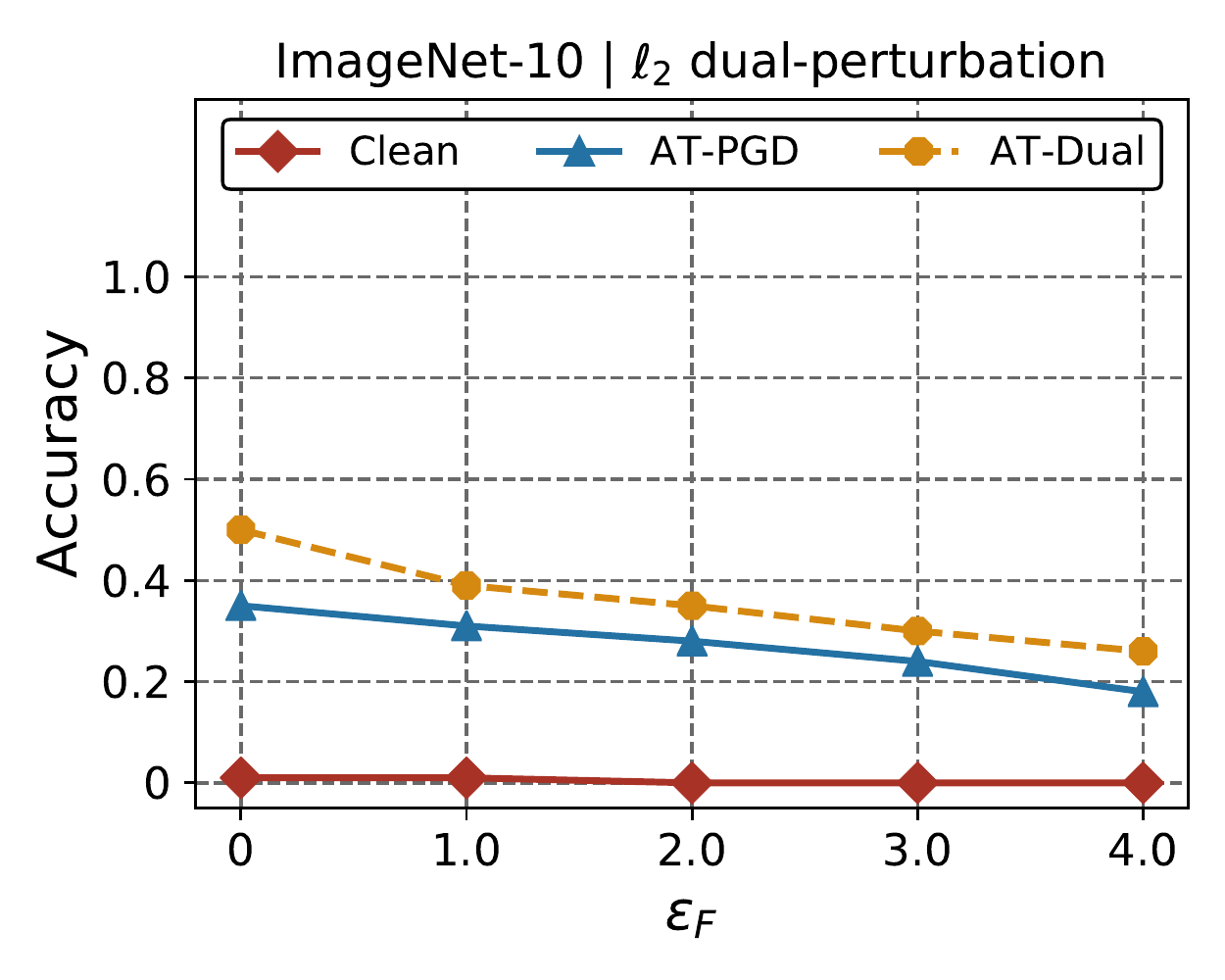} &
  \includegraphics[width=0.22\textwidth]{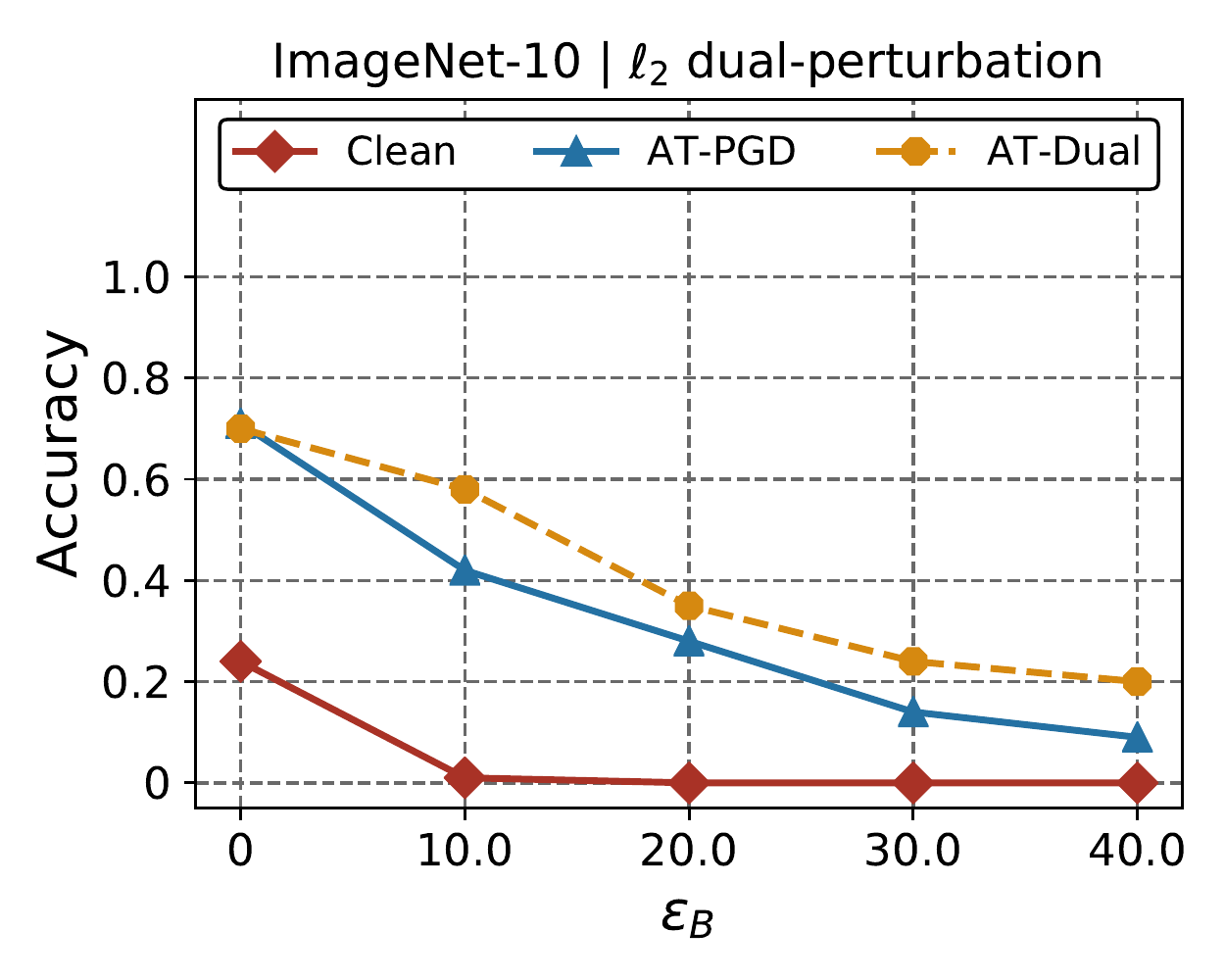} &
 \includegraphics[width=0.22\textwidth]{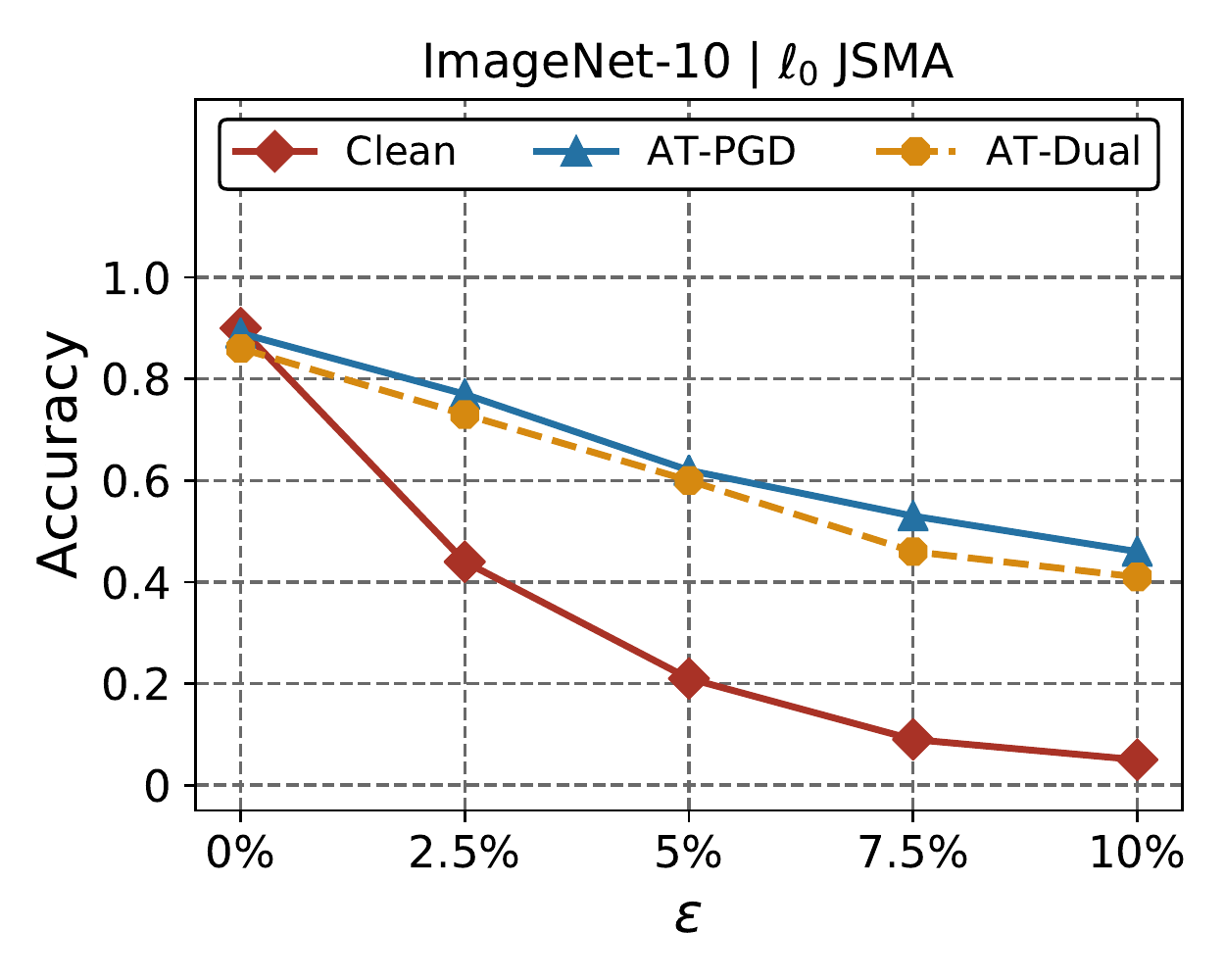} \\
\end{tabular}
\caption{
Robustness to additional white-box attacks on ImageNet-10. 
Left: 100 steps of $\ell_2$ PGD attacks. 
Middle left: 100 steps of $\ell_2$ dual-perturbation attacks with different foreground distortions. $\epsilon_B$ is fixed to be 2.0 and $\lambda=1.0$.
Middle right: 100 steps of $\ell_2$ dual-perturbation attacks with different background distortions. $\epsilon_F$ is fixed to be 20.0 and $\lambda=1.0$.
Right: $\ell_0$ JSMA attacks.
}
\label{fig:general_imagenet_linf}
\end{figure}

Next, we present experimental results of the robustness of classifiers that use adversarial training with $\ell_\infty$ norm attacks on ImageNet-10.
Specifically, we trained AT-PGD using $\ell_\infty$ PGD attack with $\epsilon=4/255$, and AT-Dual by using $\ell_\infty$ dual-perturbation attack with $\{\epsilon_F, \epsilon_B, \lambda\}=\{4/255, 20/255, 0.0\}$.
The results are shown in Figure~\ref{fig:saliency_analysis_imagenet_linf}, ~\ref{fig:white_imagenet_linf}, ~\ref{fig:black_imagenet_linf}, and ~\ref{fig:general_imagenet_linf}. 

\newpage
\section{Adversarial Training Using $\ell_\infty$ Norm Attacks on STL-10}

\begin{figure}[h]
\centering
\begin{tabular}{cc}
  \includegraphics[width=0.48\textwidth]{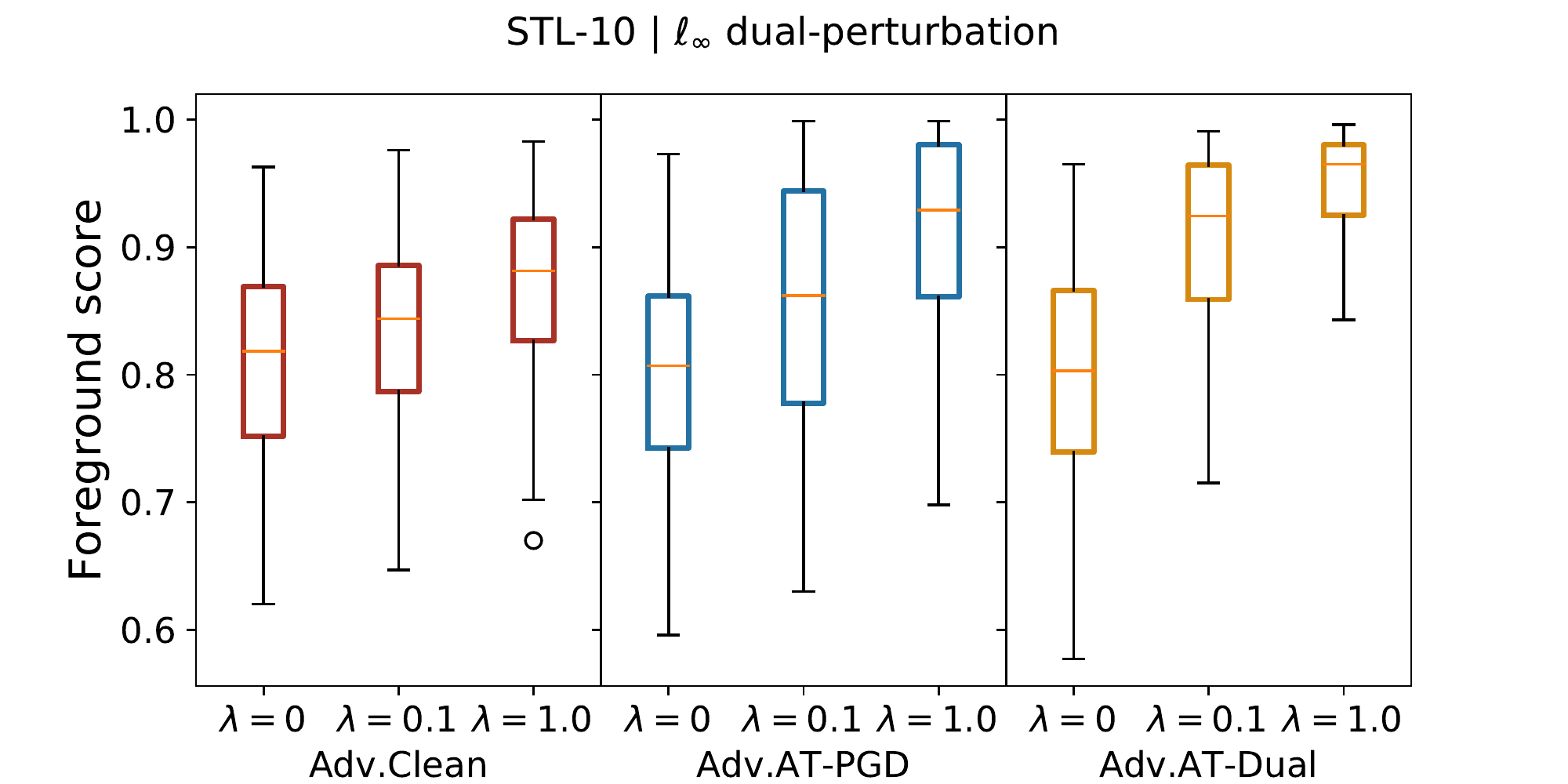} &
 \includegraphics[width=0.24\textwidth]{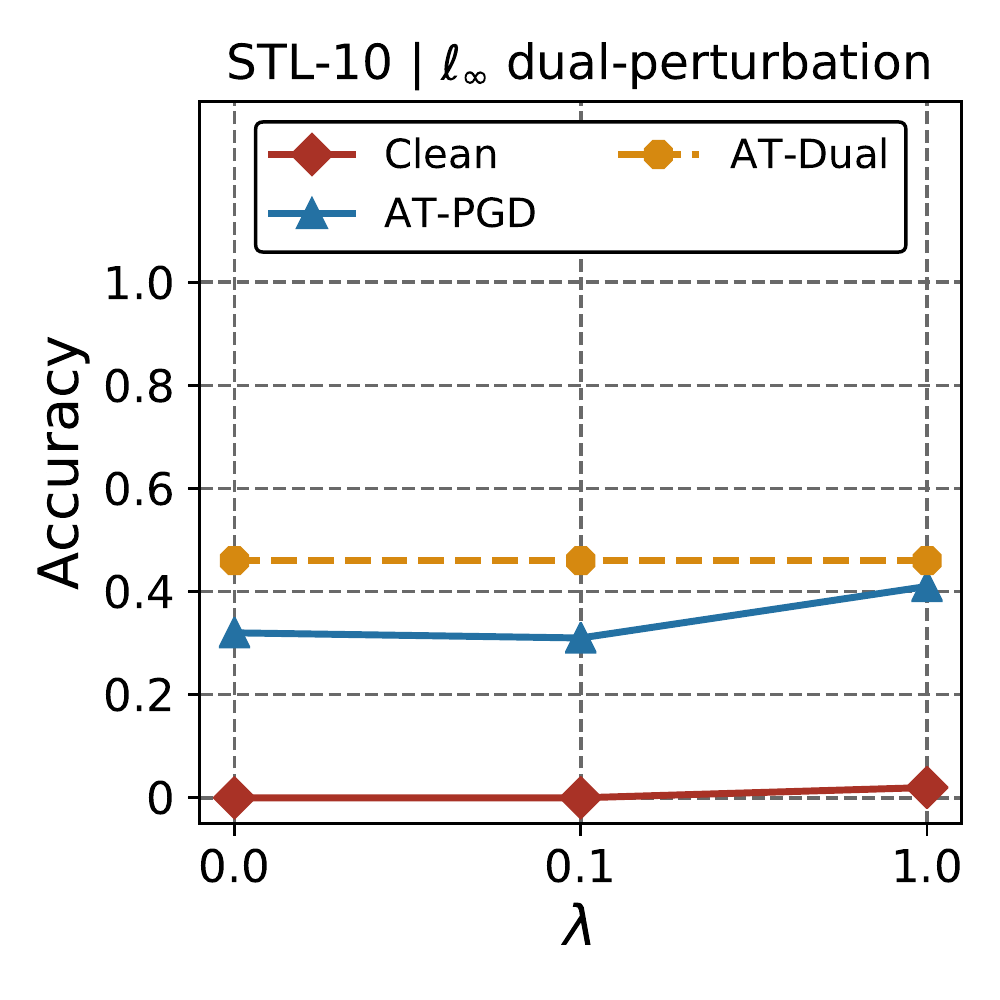}\\
\end{tabular}
	\caption{
	Saliency analysis. 
	The $\ell_\infty$ dual-perturbation attacks are performed by using $\{\epsilon_F, \epsilon_B\}=\{4/255, 20/255\}$, and a variety of $\lambda$ displayed in the figure. 
	Left: foreground scores of dual-perturbation examples in response to different classifiers.
	Right: accuracy of classifiers on dual-perturbation examples with salience control.  	
	}
	\label{fig:saliency_analysis_stl_linf}
\end{figure}

\begin{figure}[h]
\centering
\begin{tabular}{ccc}
  \includegraphics[width=0.26\textwidth]{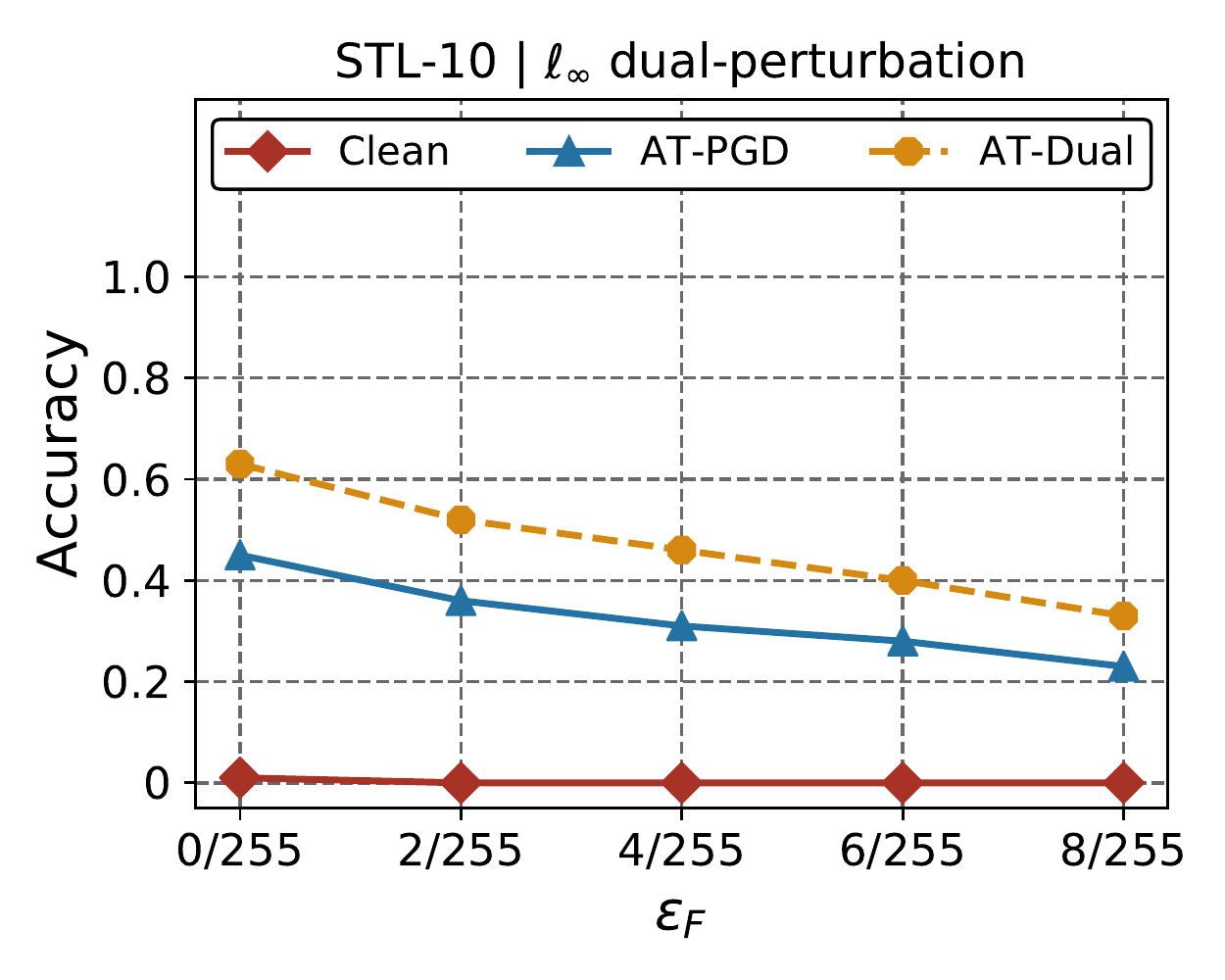} &
  \includegraphics[width=0.26\textwidth]{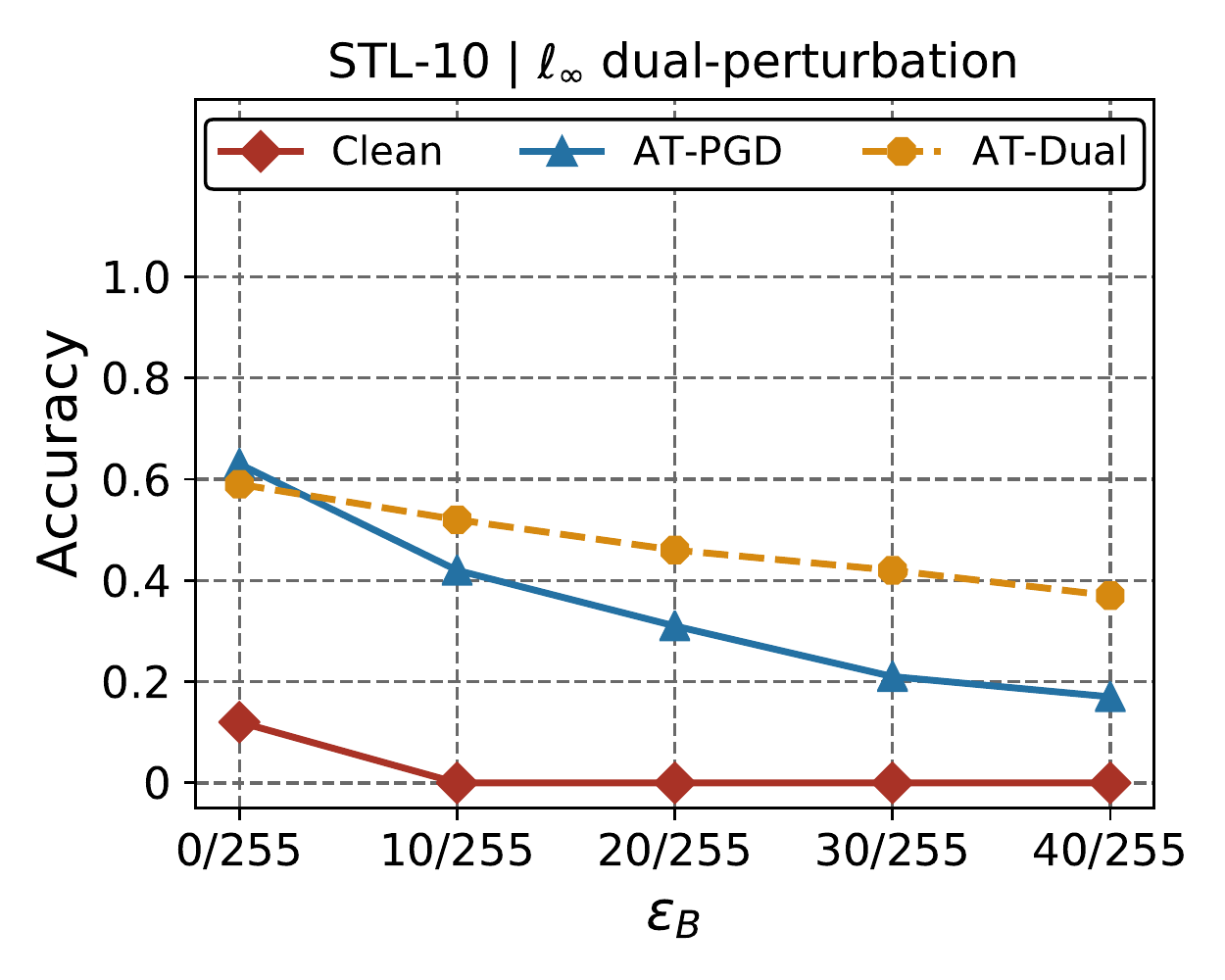} &
  \includegraphics[width=0.26\textwidth]{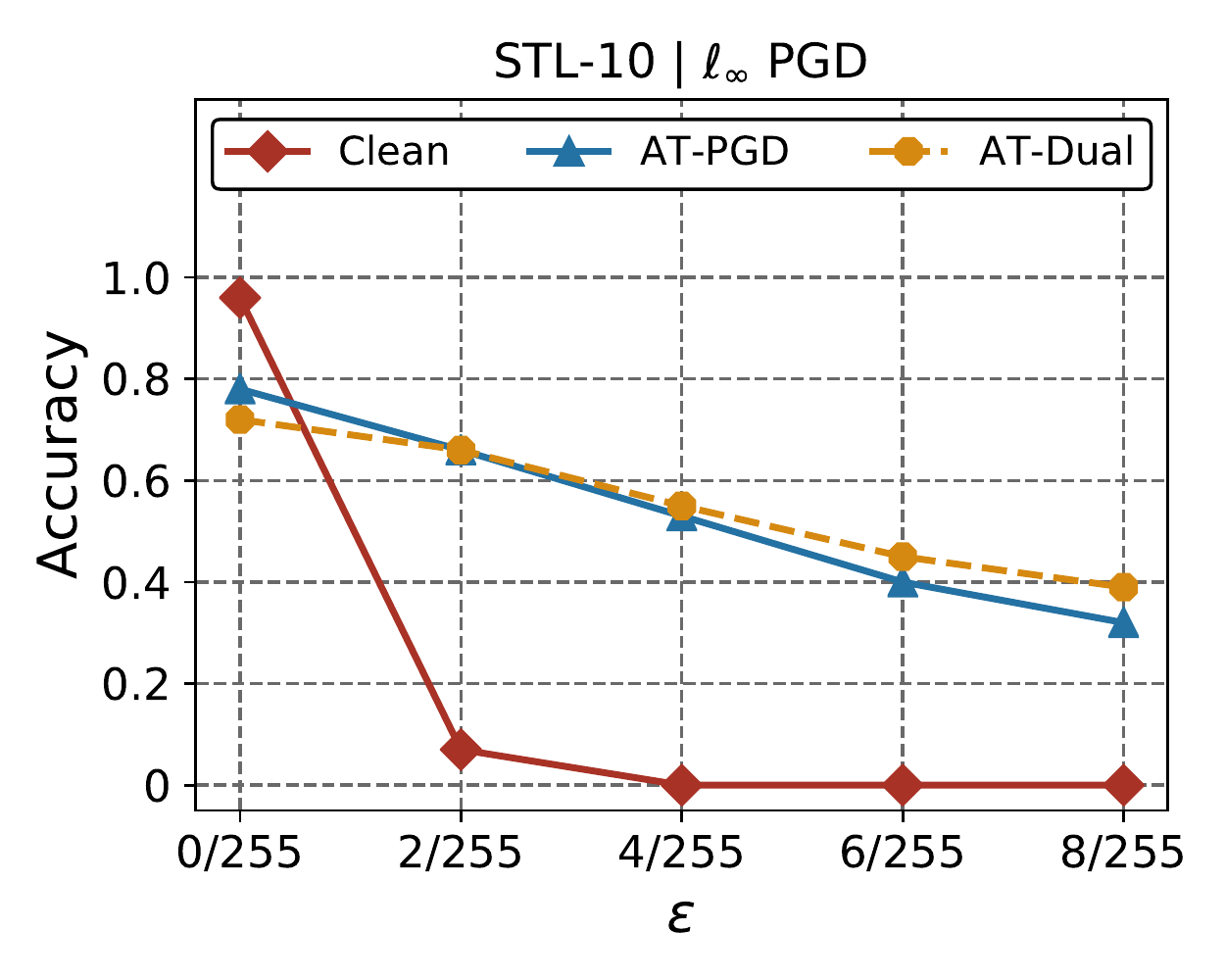}\\
\end{tabular}
\caption{
Robustness to white-box $\ell_\infty$ attacks on STL-10.
Left: $\ell_\infty$ dual-perturbation attacks with different foreground distortions. $\epsilon_B$ is fixed to be $20/255$ and $\lambda=0.1$.
Middle: $\ell_\infty$ dual-perturbation attacks with different background distortions. $\epsilon_F$ is fixed to be $4/255$ and $\lambda=0.1$.
Right: $\ell_\infty$ PGD attacks. 
}
\label{fig:white_stl_linf}
\end{figure}

\begin{figure}[h]
\centering
\begin{tabular}{cc}
  \includegraphics[width=0.35\textwidth]{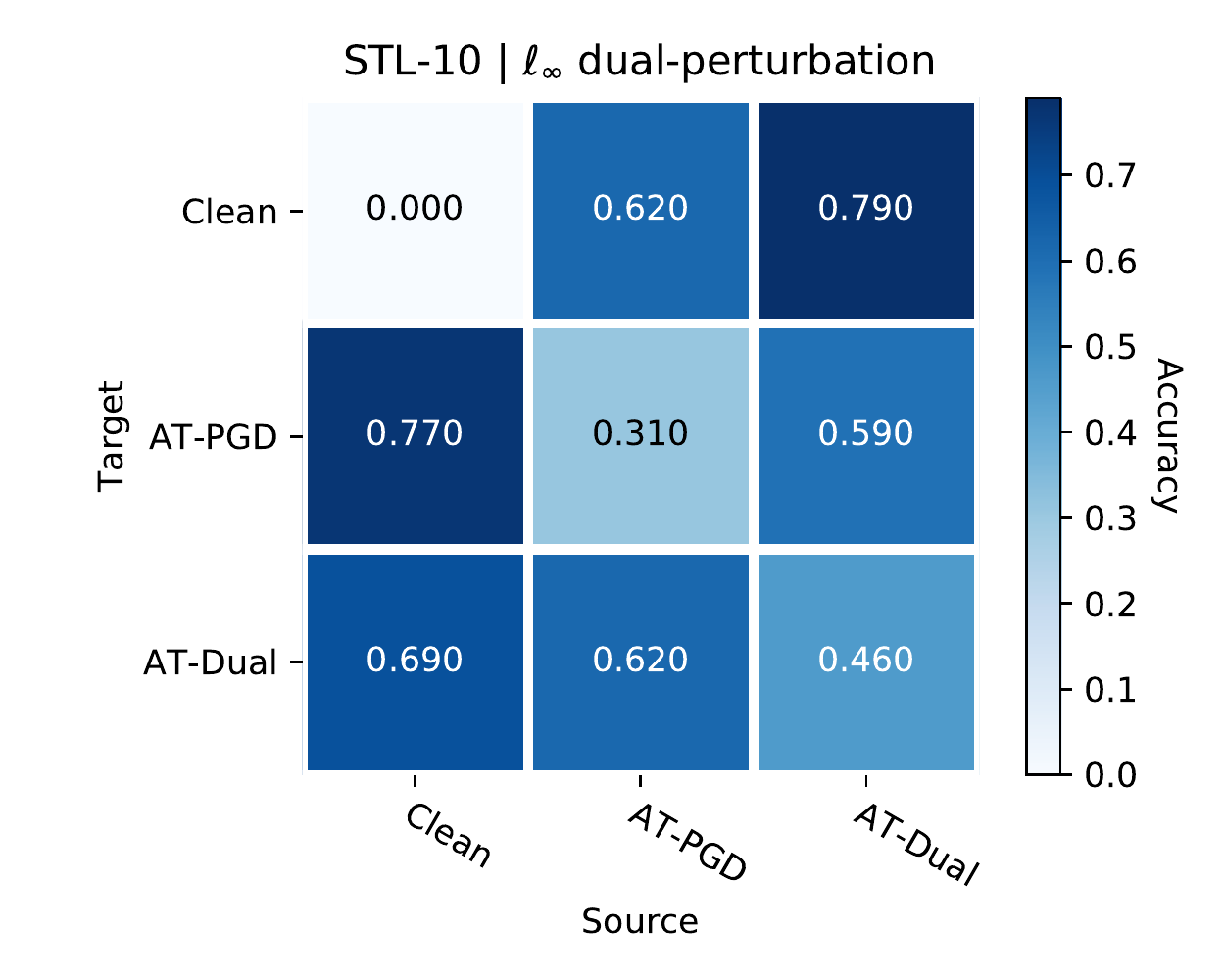} &
  \includegraphics[width=0.35\textwidth]{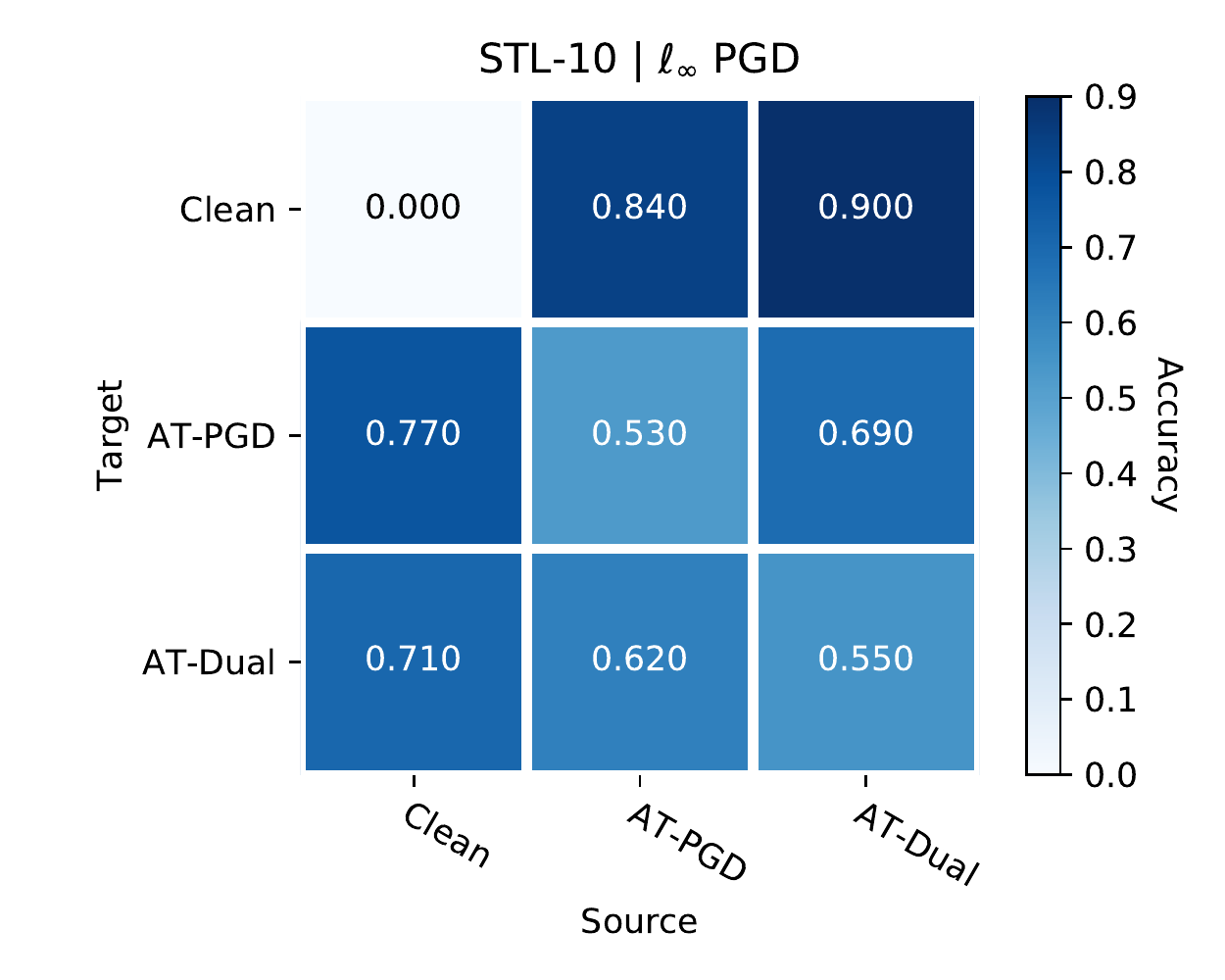}\\
\end{tabular}
\caption{
Robustness against adversarial examples transferred from other models on STL-10.  
Left: $\ell_\infty$ dual-perturbation attacks performed by using $\{\epsilon_F, \epsilon_B, \lambda\}=\{4/255, 20/255, 1.0\}$ on different source models.
Right: $\ell_\infty$ PGD attacks with $\epsilon=4/255$ on different source models.
}
\label{fig:black_stl_linf}
\end{figure}

\begin{figure}[h!]
\centering
\begin{tabular}{cccc}
  \includegraphics[width=0.22\textwidth]{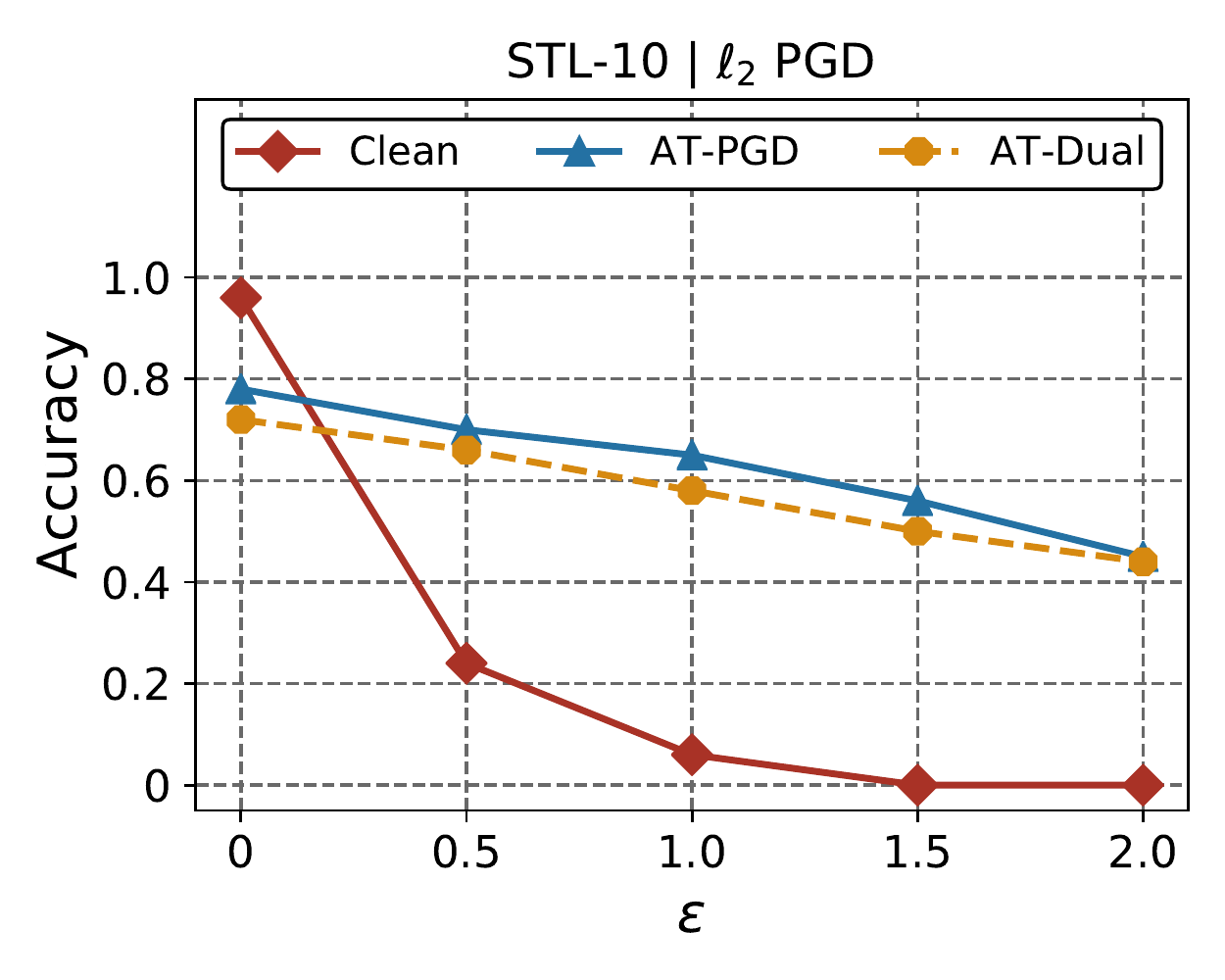} &
 \includegraphics[width=0.22\textwidth]{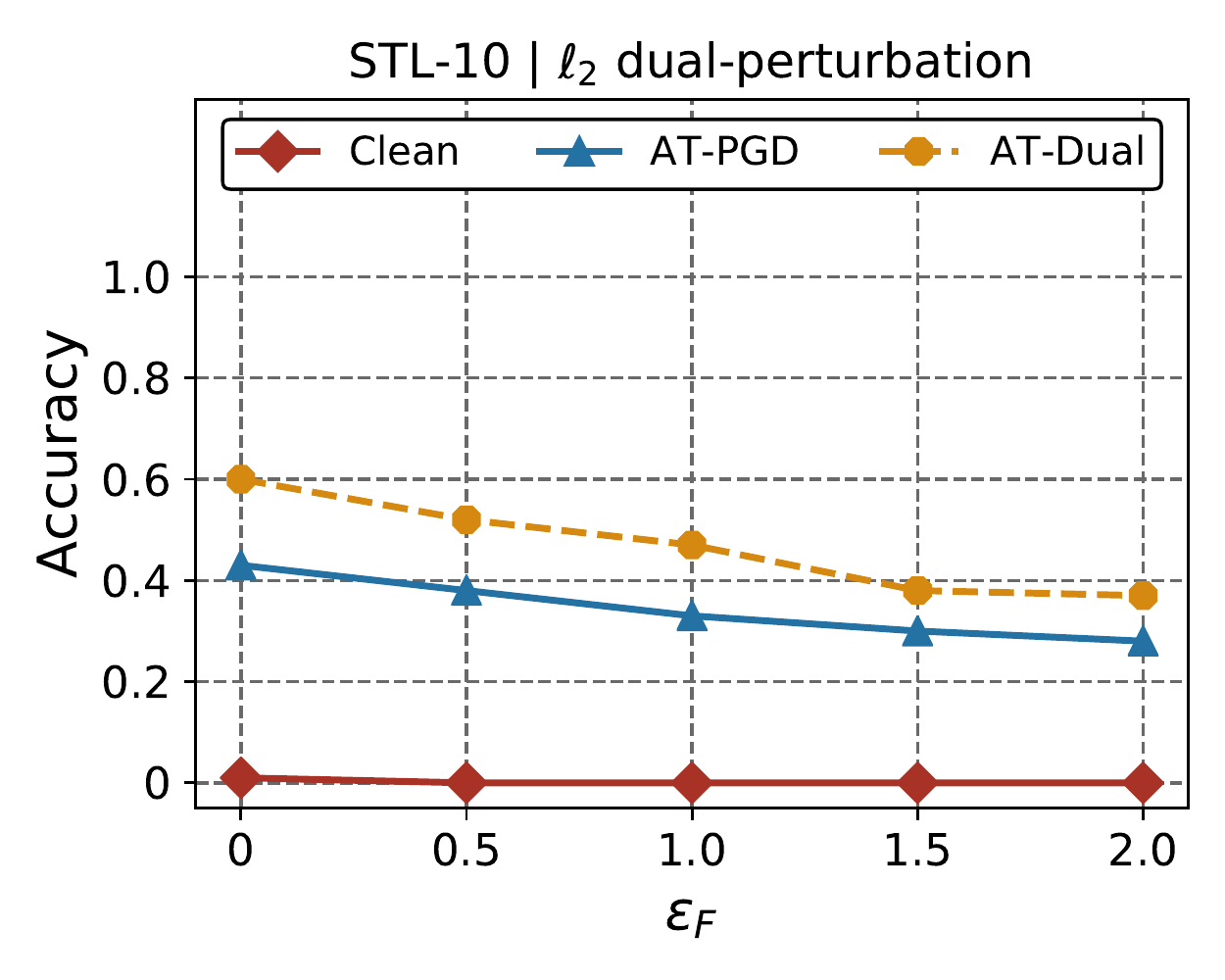} &
  \includegraphics[width=0.22\textwidth]{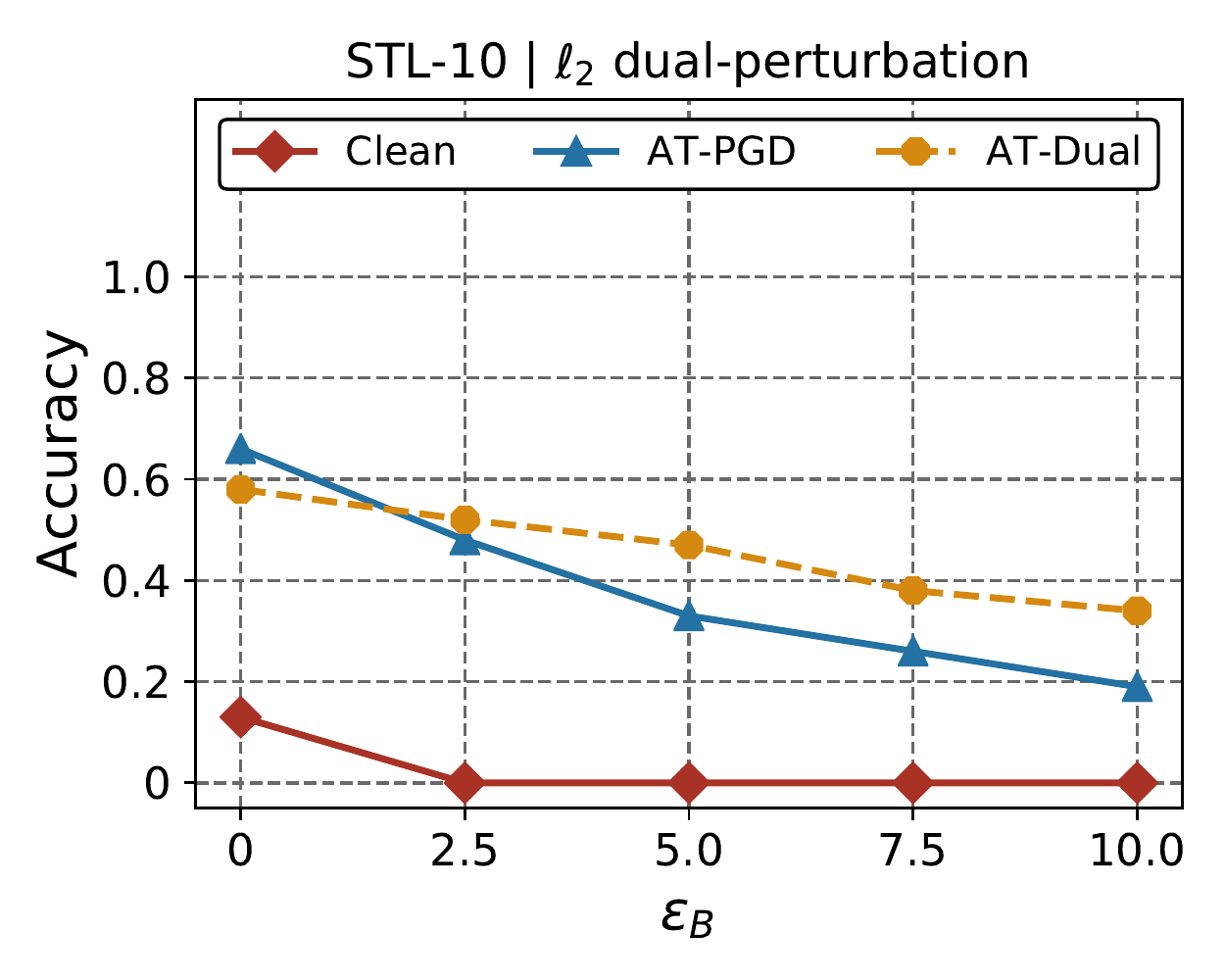} &
 \includegraphics[width=0.22\textwidth]{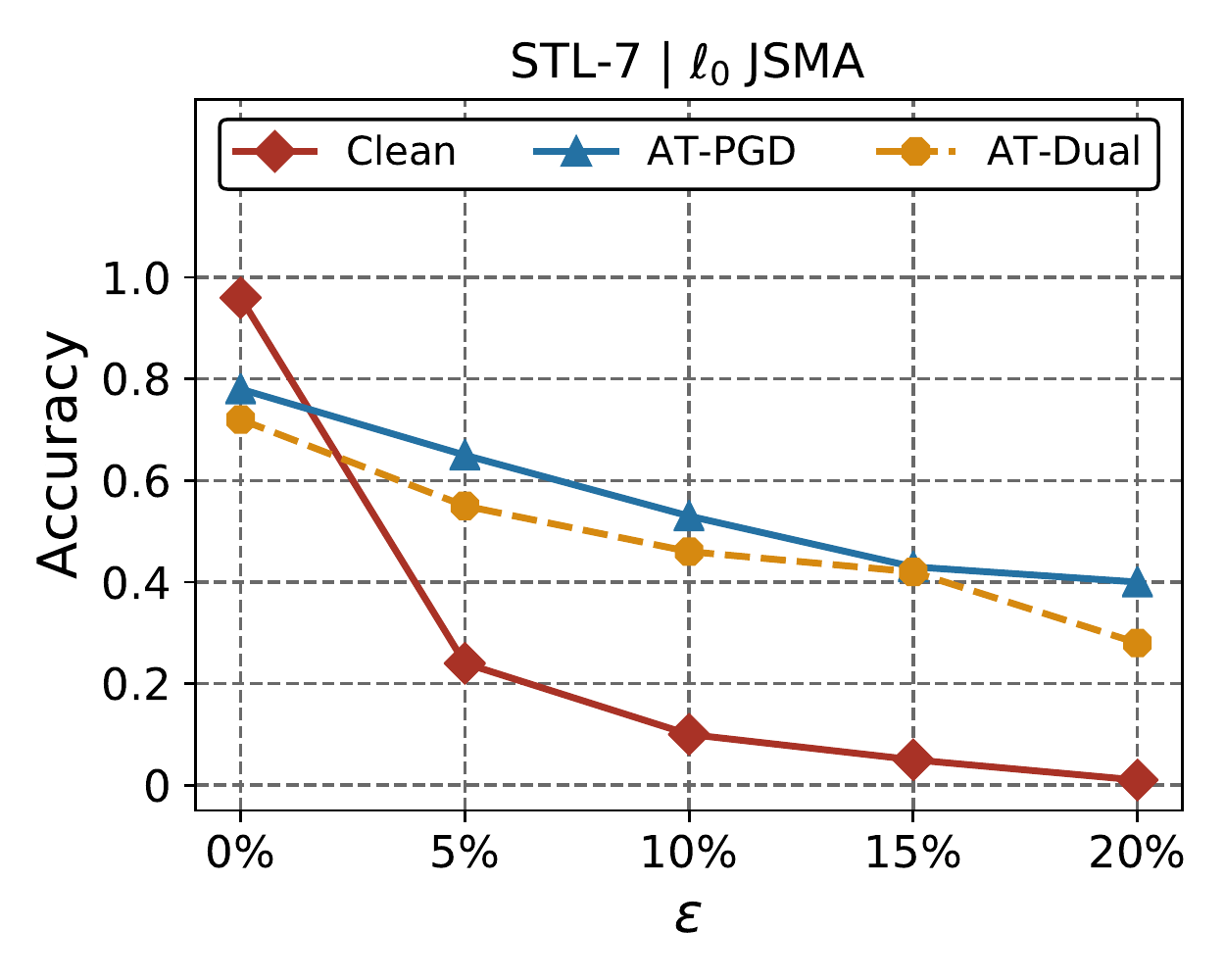} \\
\end{tabular}
\caption{
Robustness to additional white-box attacks on STL-10. 
Left: 100 steps of $\ell_2$ PGD attacks. 
Middle left: 100 steps of $\ell_2$ dual-perturbation attacks with different foreground distortions. $\epsilon_B$ is fixed to be 5.0 and $\lambda=0.1$.
Middle right: 100 steps of $\ell_2$ dual-perturbation attacks with different background distortions. $\epsilon_F$ is fixed to be 1.0 and $\lambda=0.1$.
Right: $\ell_0$ JSMA attacks.
}
\label{fig:general_stl_linf}
\end{figure}

Now, we present experimental results of the robustness of classifiers that use adversarial training with $\ell_\infty$ norm attacks on STL-10.
Specifically, we trained AT-PGD using $\ell_\infty$ PGD attack with $\epsilon=4/255$, and AT-Dual by using $\ell_\infty$ dual-perturbation attack with $\{\epsilon_F, \epsilon_B, \lambda\}=\{4/255, 20/255, 0.0\}$.
The results are shown in Figure~\ref{fig:saliency_analysis_stl_linf}, ~\ref{fig:white_stl_linf}, ~\ref{fig:black_stl_linf}, and ~\ref{fig:general_stl_linf}. 

\newpage
\section{Adversarial Training Using $\ell_\infty$ Norm Attacks on Segment-6}

\begin{figure}[h]
\centering
\begin{tabular}{cc}
  \includegraphics[width=0.35\textwidth]{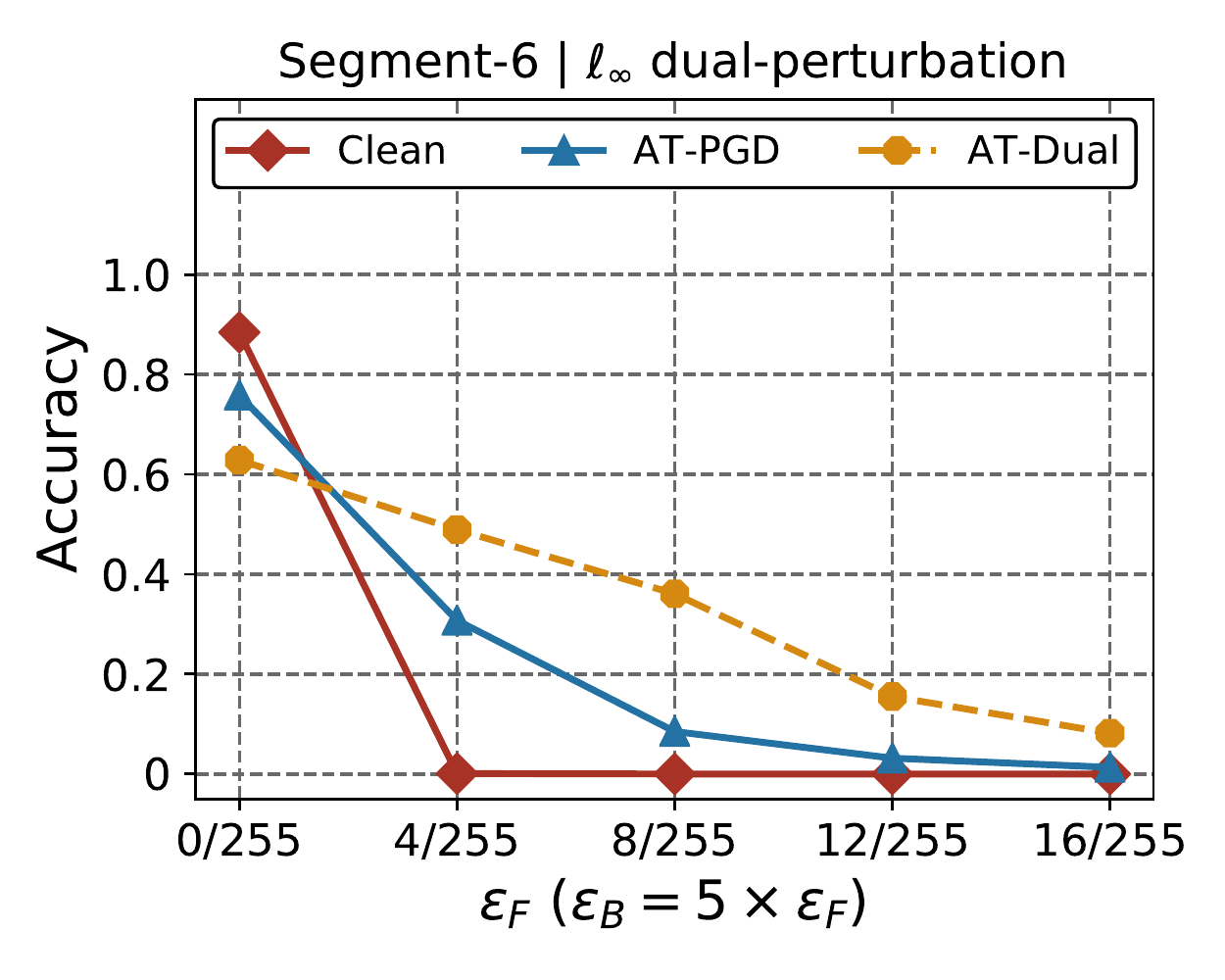} &
  \includegraphics[width=0.35\textwidth]{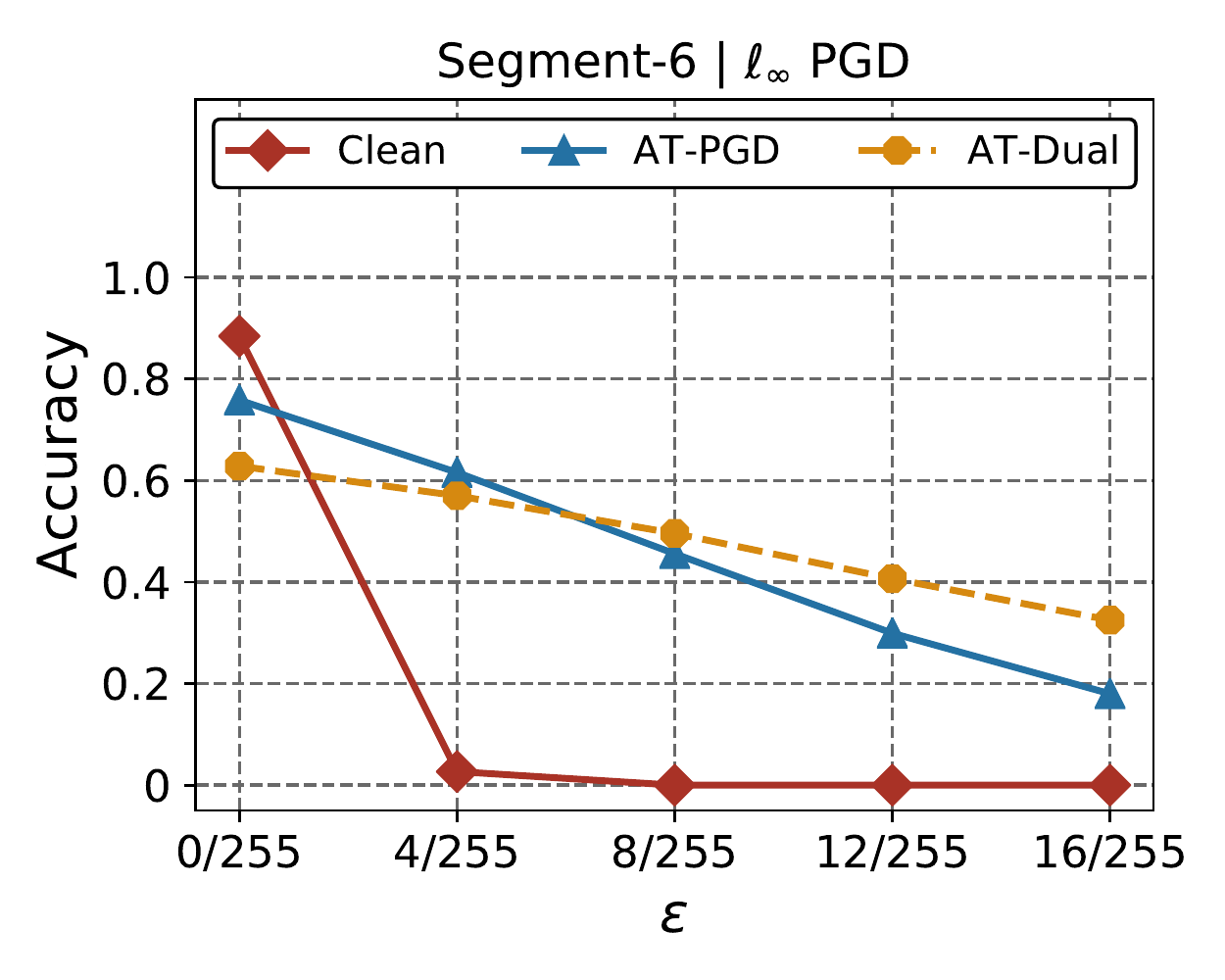}\\
\end{tabular}
\caption{
Robustness to white-box $\ell_\infty$ attacks on Segment-6.
Left: $\ell_\infty$ dual-perturbation attacks with different foreground and background distortions. 
Right: $\ell_\infty$ PGD attacks. 
}
\label{fig:white_segment_linf}
\end{figure}

\begin{figure}[h]
\centering
\begin{tabular}{cc}
  \includegraphics[width=0.35\textwidth]{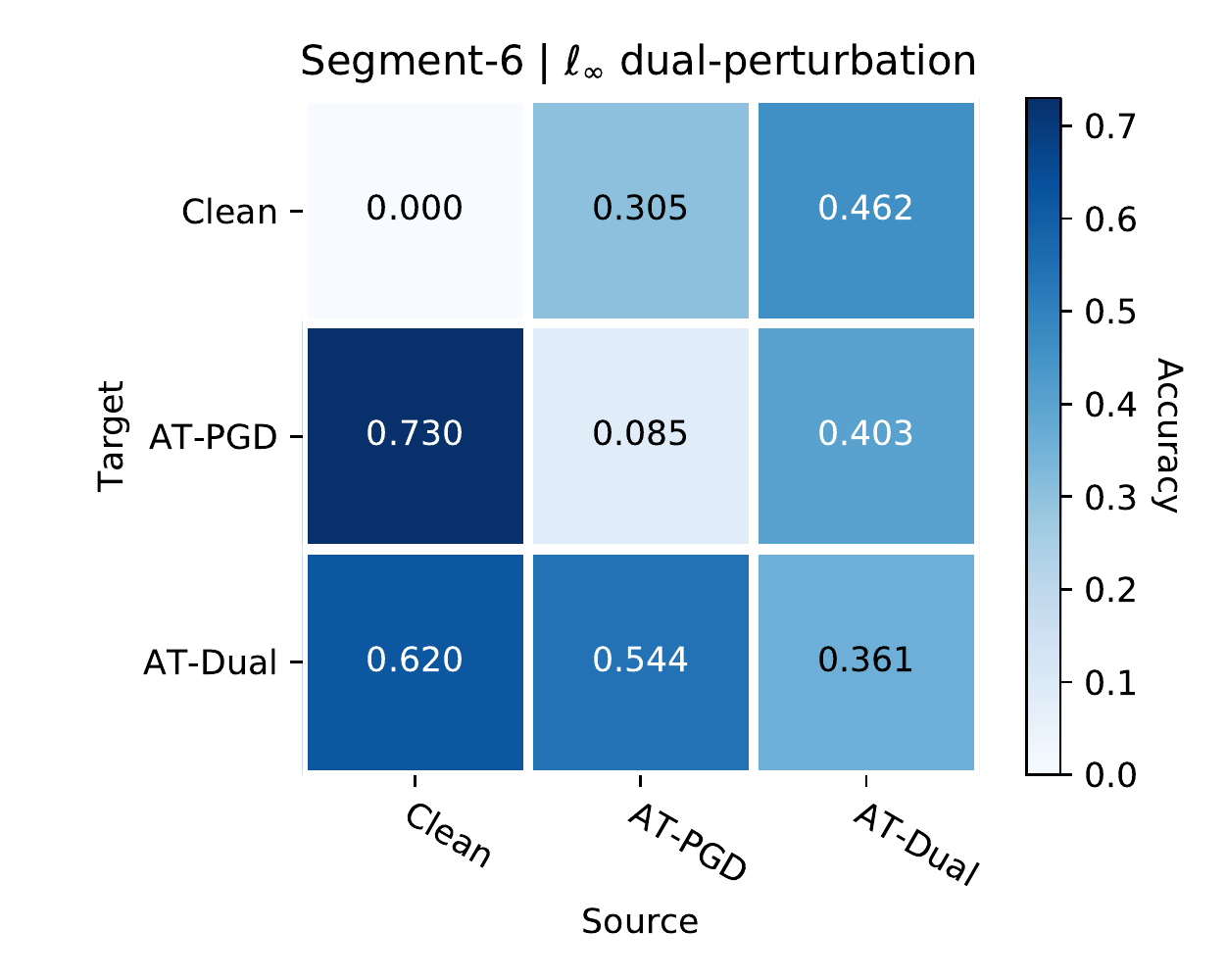} &
  \includegraphics[width=0.35\textwidth]{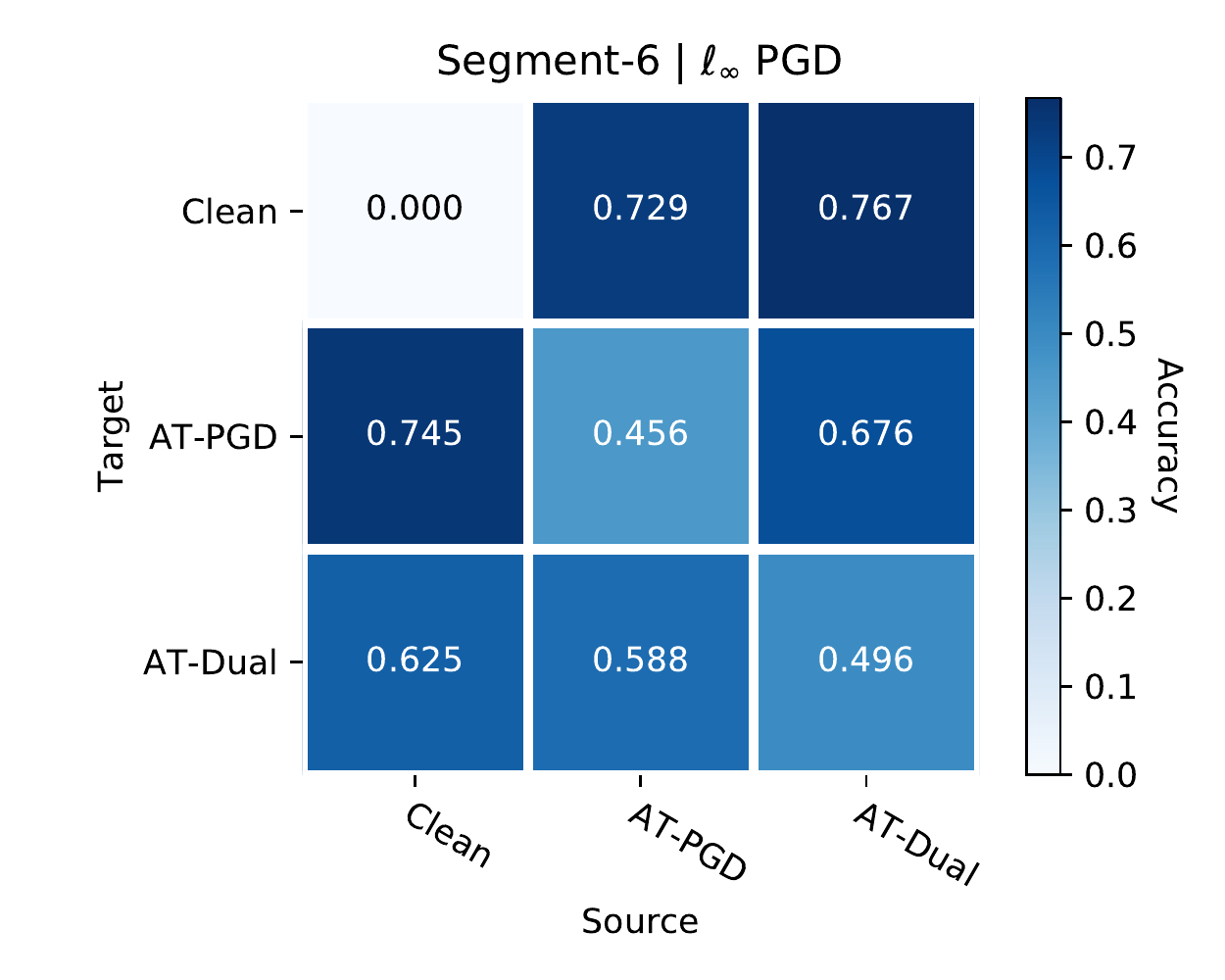}\\
\end{tabular}
\caption{
Robustness against adversarial examples transferred from other models on Segment-6.  
Left: $\ell_\infty$ dual-perturbation attacks performed by using $\{\epsilon_F, \epsilon_B\}=\{8/255, 40/255\}$ on different source models.
Right: $\ell_\infty$ PGD attacks with $\epsilon=8/255$ on different source models.
}
\label{fig:black_segment6_linf}
\end{figure}

\begin{figure}[h!]
\centering
\begin{tabular}{ccc}
  \includegraphics[width=0.26\textwidth]{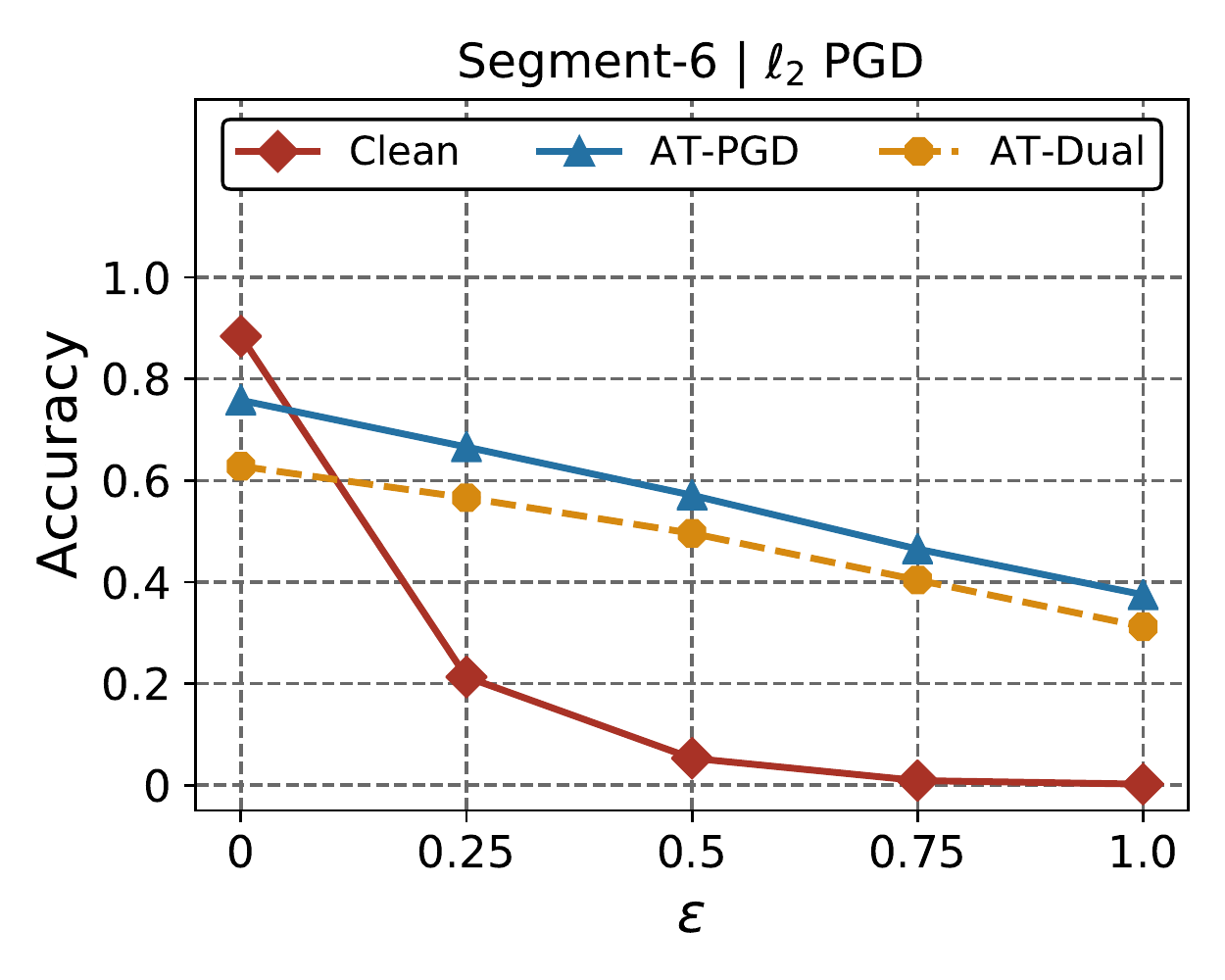} &
 \includegraphics[width=0.26\textwidth]{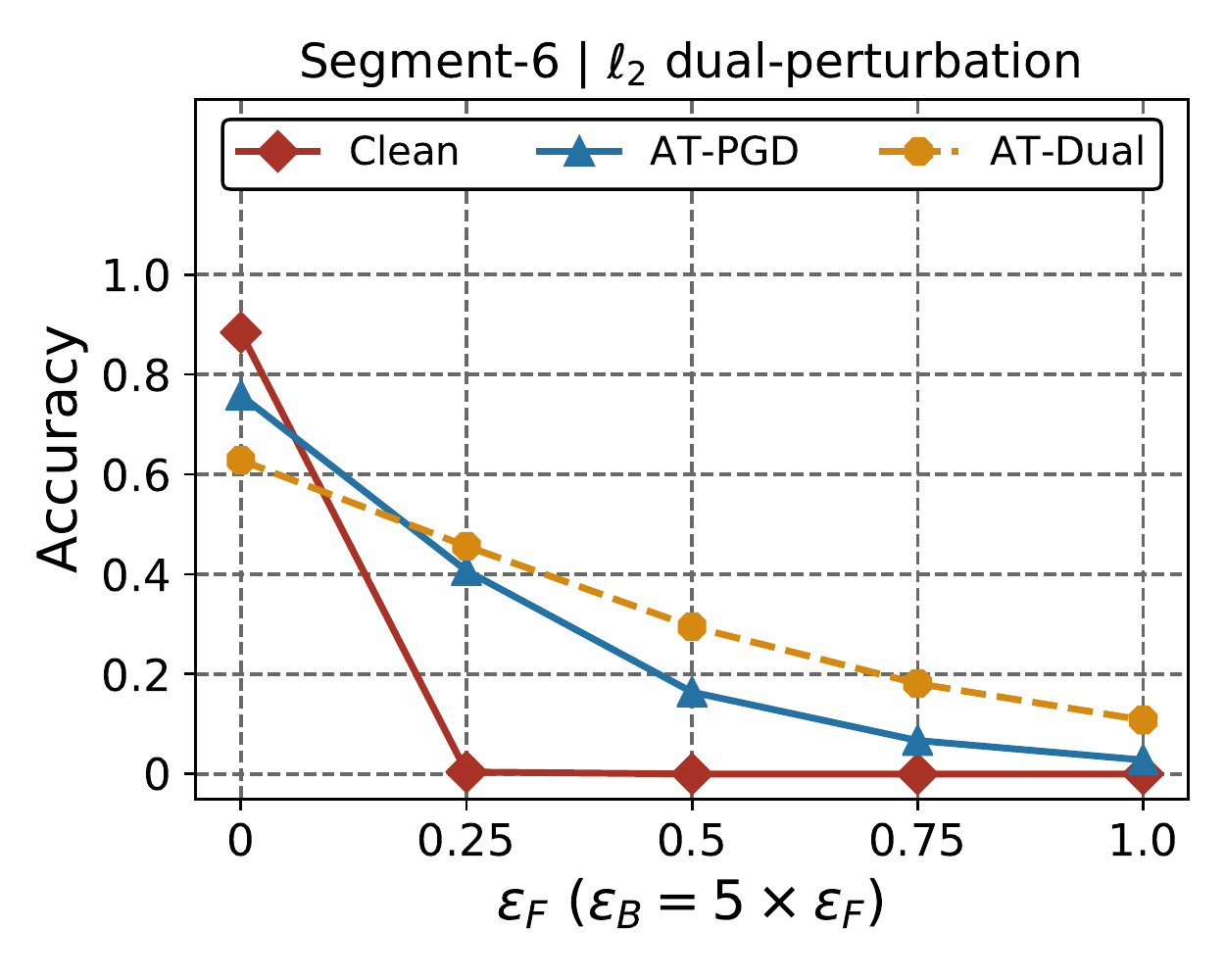} &
 \includegraphics[width=0.26\textwidth]{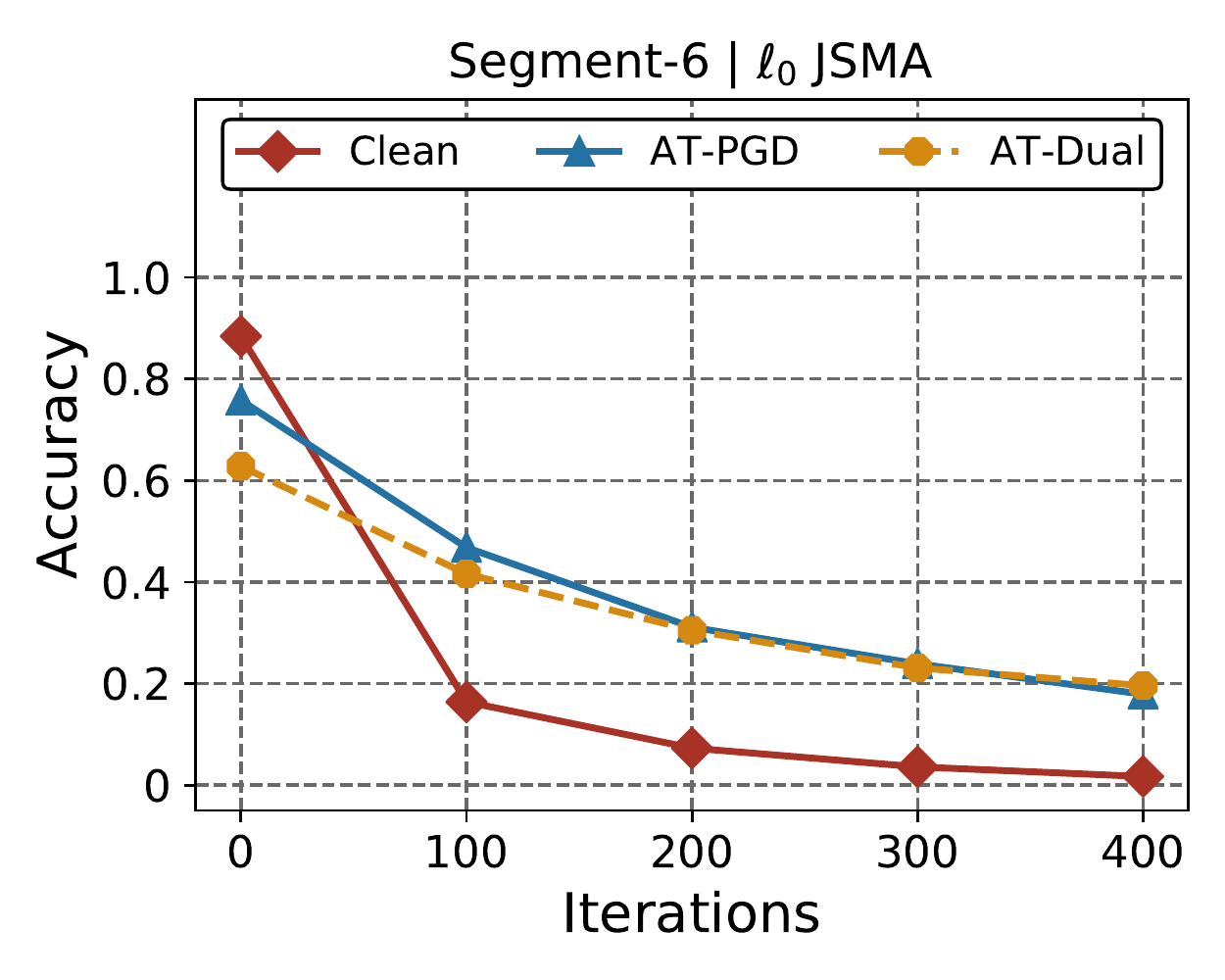} \\
\end{tabular}
\caption{
Robustness to additional white-box attacks on Segment-6. 
Left: 100 steps of $\ell_2$ PGD attacks. 
Middle: 100 steps of $\ell_2$ dual-perturbation attacks with different foreground and background distortions.
Right: $\ell_0$ JSMA attacks.
}
\label{fig:general_segment6_linf}
\end{figure}

Finally, we present experimental results of the robustness of classifiers that use adversarial training with $\ell_\infty$ norm attacks on Segment-6.
We trained AT-PGD using $\ell_\infty$ PGD attack with $\epsilon=8/255$, and AT-Dual by using $\ell_\infty$ dual-perturbation attack with $\{\epsilon_F, \epsilon_B\}=\{8/255, 40/255\}$.
The results are shown in Figure~\ref{fig:white_segment_linf}, ~\ref{fig:black_segment6_linf}, and ~\ref{fig:general_segment6_linf}. 

\newpage
\section{Attacking Randomzied Classifiers}
In addition to \emph{deterministic classifiers} that make a deterministic prediction for a test sample, our proposed attack can be adapted to \emph{stochastic classifiers} that apply randomization at training and prediction time.
For example, for classifiers using \emph{randomized smoothing}, we can refine Equation 3 in the main body of the paper as follows:
\begin{equation}
\max_{\substack{||\bm{\delta} \circ \mathcal{F}(\bm{x})||_p \leq \epsilon_F,\\ ||\bm{\delta}\circ\mathcal{B}(\bm{x})||_p \leq \epsilon_B}} \mathbb{E}_{\bm{\eta}\sim\mathcal{N}(\bm{0}, \sigma^{2} \bm{I})} [ \mathcal{L}\left(h_{\bm{\theta}}(\bm{x}+\bm{\delta}+\bm{\eta}), y\right) + \lambda \cdot \mathcal{S} \left( \bm{x}+\bm{\delta}+\bm{\eta} \right) ],
\label{eq:dual_pgd_rs}
\end{equation}
where $\sigma^{2}$ is the variance of the Gaussian data augmentation in randomized smoothing.~\footnote{Note that the Gaussian perturbations are only used to compute the expection of loss and are not in the resulting adversarial examples.} 
The optimization problem in Equation~\ref{eq:dual_pgd_rs} can be solved by the same approach used for deterministic classifiers, with the following modification on Equation~\ref{eq:gf_gb} at the second step in Section~\ref{sec:solution}:
\begin{equation}
\begin{cases}
g_F = \mathcal{G}(\mathcal{F}(\bm{x}) \circ \nabla_{\bm{\delta}^{(k)}} \mathbb{E}_{\bm{\eta}} [ \mathcal{L}(h_{\bm{\theta}}(\bm{x}+\bm{\delta}^{(k)}+\bm{\eta}), y) + \lambda \cdot \mathcal{S} \left( \bm{x}+\bm{\delta}^{(k)} + \bm{\eta} \right) ]) \\
g_B = \mathcal{G}(\mathcal{B}(\bm{x}) \circ \nabla_{\bm{\delta}^{(k)}} \mathbb{E}_{\bm{\eta}} [ \mathcal{L}(h_{\bm{\theta}}(\bm{x}+\bm{\delta}^{(k)}+\bm{\eta}), y) + \lambda \cdot \mathcal{S} \left( \bm{x}+\bm{\delta}^{(k)} + \bm{\eta} \right) ])
\end{cases}.
\end{equation}

\subsection{Variance in Gaussian Data Augmentation}
Table~\ref{tab:rs_linf} and \ref{tab:rs_segment6_l2} show the effectiveness of \emph{Randomized Smoothing (RS)} against the proposed dual-perturbation attack.
Here, we use different variances in Gaussian data augmentation of \emph{RS}, and fix the number of noise-corrupted copies at prediction time, $n$ to be 100.
It can be seen that \emph{RS} is generally fragile to the dual-perturbation attacks that are adapted to randomized classifiers.
Moreover, increasing $\sigma$, the variance used in Gaussian data augmentation can only marginally improve adversarial robustness to dual-perturbation attacks while significantly decrease accuracy on non-adversarial data.

\begin{table*}[h]
\centering
\scalebox{0.90}{
\begin{tabular}{|c|c|c|c|c|c|c|}
\hline
\multirow{2}{*}{\textbf{Dataset}} & \multirow{2}{*}{\textbf{Defense approach}} & \multicolumn{5}{c|}{\textbf{Attack Strength ($\epsilon_B = 5 \times \epsilon_F$)}}                    \\ \cline{3-7} 
                                  &                                            & $\epsilon_F = 0/255$ & $\epsilon_F=4/255$ & $\epsilon_F=8/255$ & $\epsilon_F=12/255$ & $\epsilon_F=1$ \\ \hline \hline
\multirow{3}{*}{Segment-6}        & RS, $\sigma=0.25$                          & 71.4\%               & 9.6\%             & 0.4\%              & 0.1\%               & 0.0\%          \\ \cline{2-7} 
                                  & RS, $\sigma=0.5$                           & 61.7\%               & 13.7\%             & 1.9\%             & 0.6\%               & 0.2\%          \\ \cline{2-7} 
                                  & RS, $\sigma=1$                             & 47.7\%               & 15.6\%             & 2.8\%             & 0.4\%                 & 0.2\%          \\ \hline \hline
\end{tabular}
}
\caption{Robustness of \emph{RS} against $\ell_\infty$ dual-perturbation attacks.}
\label{tab:rs_linf}
\end{table*}

\begin{table*}[h]
\centering
\scalebox{0.90}{
\begin{tabular}{|c|c|c|c|c|c|}
\hline
\multirow{2}{*}{\textbf{Defense approach}} & \multicolumn{5}{c|}{\textbf{Attack Strength} ($\epsilon_B = 5 \times \epsilon_F$)}                    \\ \cline{2-6} 
                                  & $\epsilon_F = 0$ & $\epsilon_F=0.25$ & $\epsilon_F=0.5$ & $\epsilon_F=0.75$ & $\epsilon_F=1$ \\ \hline \hline
RS, $\sigma=0.25$                 & 71.4\%           & 29.7\%            & 6.7\%            & 0.9\%             & 0.1\%          \\ \hline
RS, $\sigma=0.5$                  & 61.7\%           & 31.6\%            & 11.8\%           & 3.1\%             & 1.3\%          \\ \hline
RS, $\sigma=1$                    & 47.7\%           & 28.2\%            & 14.4\%           & 6.0\%               & 1.5\%          \\ \hline
\end{tabular}
}
\caption{Robustness of \emph{RS} against $\ell_2$ dual-perturbation attacks on Segment-6.}
\label{tab:rs_segment6_l2}
\end{table*}

\subsection{Number of Samples with Gaussian Noise at Prediction Time}
It has been observed that \emph{Randomized Smoothing (RS)} can be computationally inefficient at prediction time as it uses a large number of noise-corrupted copies for each test sample at prediction time.
It is natural to ask whether the prediction time of \emph{RS} can be reduced without significantly sacrificing adversarial robustness in practice. 
We answer this question by studying the effectiveness of \emph{RS} with different $n$, the numbers of noise-corrupted copies at prediction time.
Specifically, we fix $\sigma=0.5$ and set $n$ to be 1, 25, and 100.
Note that when $n=1$, there is no two-sided hypothesis test for prediction; thus, no abstentions are obtained.

Here we use $\ell_\infty$ dual-perturbation attacks on \emph{RS} for demonstration purposes.
The results are shown in Table~\ref{tab:rs_linf_n}.
It can be seen that when $n=25$, the accuracy on both adversarial and non-adversarial data can drop by up to 10\% compared to \emph{RS} using $n=100$.
The reason is that under a small $n$, the prediction appears more likely to abstain.
Interestingly, when $n=1$, the accuracy can be marginally improved compared to $n=100$, with the prediction time being reduced by 99\%.
This indicates that in practice, we would not lose accuracy without using the two-sided hypothesis test at prediction time.

\begin{table*}[h]
\centering
\scalebox{0.90}{
\begin{tabular}{|c|c|c|c|c|c|c|}
\hline
\multirow{2}{*}{\textbf{Dataset}} & \multirow{2}{*}{\textbf{Defense approach}} & \multicolumn{5}{c|}{\textbf{Attack Strength ($\epsilon_B = 5 \times \epsilon_F$)}}                    \\ \cline{3-7} 
                                  &                                            & $\epsilon_F = 0/255$ & $\epsilon_F=4/255$ & $\epsilon_F=8/255$ & $\epsilon_F=12/255$ & $\epsilon_F=1$ \\ \hline \hline
\multirow{3}{*}{Segment-6}        & RS, $n=1$                          & 66.0\%               & 19.8\%   & 3.2\%     & 0.8\%     & 0.3\%          \\ \cline{2-7} 
                                  & RS, $n=25$                         & 49.4\%               & 9.1\%    & 1.3\%     & 0.5\%     & 0.0\%          \\ \cline{2-7} 
                                  & RS, $n=100$                        & 61.7\%               & 13.7\%   & 1.9\%      & 0.6\%     & 0.2\%          \\ \hline 
\end{tabular}
}
\caption{Robustness of \emph{RS} against $\ell_\infty$ dual-perturbation attacks under different numbers of noise-corrupted copies at prediction time.}
\label{tab:rs_linf_n}
\end{table*}

\newpage
\section{Visualization of Loss Gradient}

\begin{figure}[h]
\centering
 \includegraphics[width=0.95\textwidth]{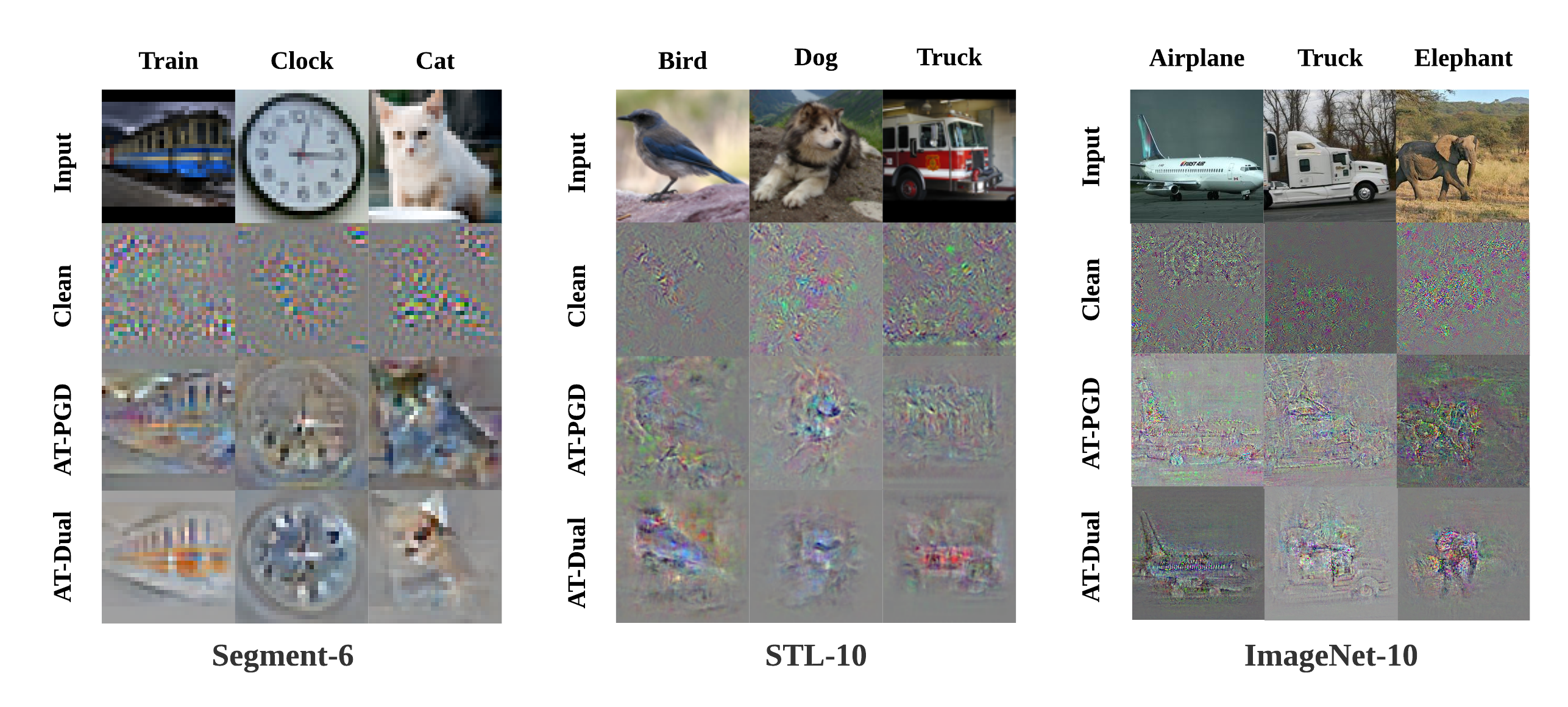} 
\caption{
Visualization of loss gradient of different classifiers with respect to pixels of \emph{non-adversarial} inputs.
AT-PGD and AT-Dual were obtained using adversarial training with corresponding $\ell_2$ norm attacks.
}
\label{fig:visualization_gradient}
\end{figure}

\section{Examples of Dual-Perturbation Attacks}

\begin{figure}[h!]
\centering
 \includegraphics[width=0.95\textwidth]{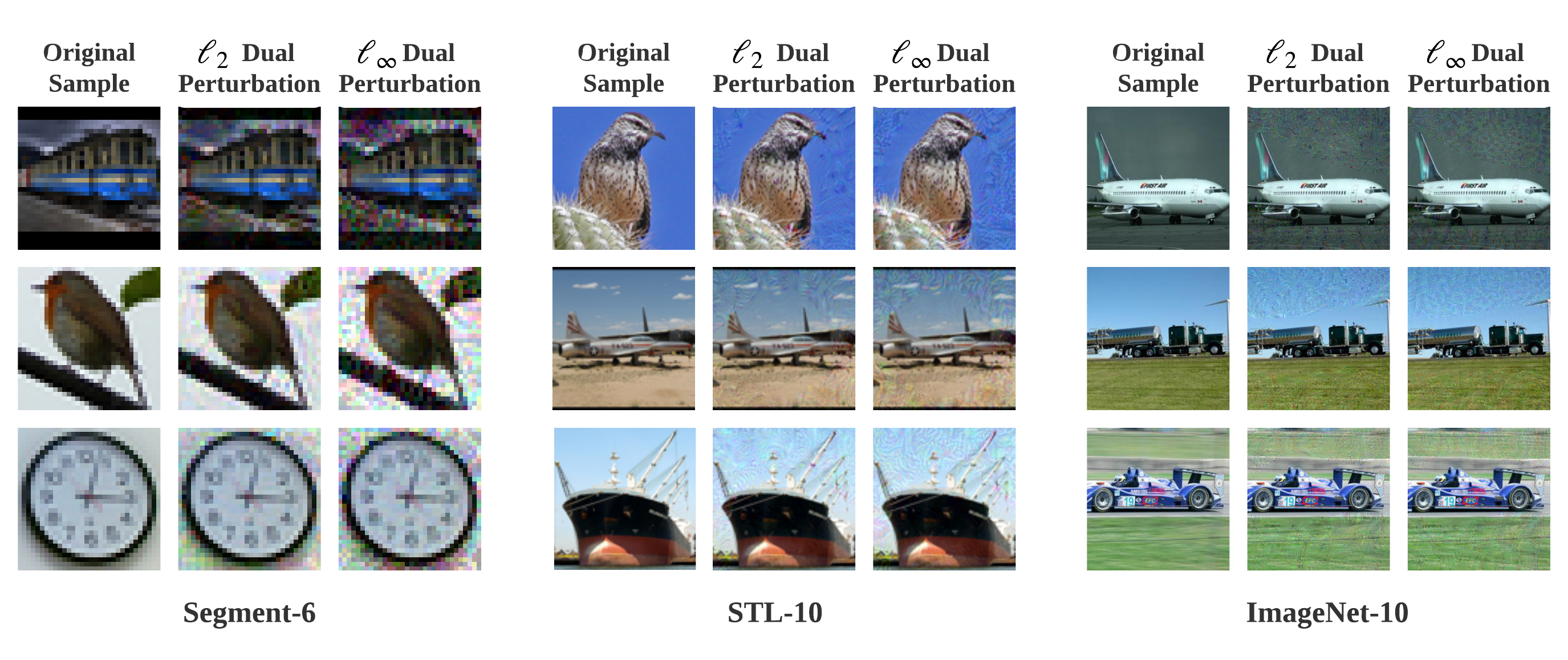} 
\caption{
Dual-perturbation attacks.
Adversarial examples are produced in response to the $\emph{Clean}$ model for each dataset.
}
\label{fig:adv_example}
\end{figure}

\end{document}